%% file: arxiv_main.tex
\documentclass[10pt,twocolumn,letterpaper]{article}

\usepackage[pagenumbers]{cvpr} 

\definecolor{cvprblue}{rgb}{0.21,0.49,0.74}

\input{preamble}

\usepackage[pagebackref=true,breaklinks=true,colorlinks,bookmarks=false]{hyperref}

\def\paperID{****} 
\def\confName{CVPR}
\def\confYear{2025}

\title{OSDFace: One-Step Diffusion Model for Face Restoration}

\author{Jingkai Wang$^{1}$\thanks{Equal contribution.}~,\enspace Jue Gong$^{1}$\footnotemark[1],\enspace Lin Zhang$^{1}$,\enspace Zheng Chen$^{1}$\\
\enspace Xing Liu$^{2}$,\enspace Hong Gu$^{2}$,\enspace Yutong Liu$^{1}$\thanks{Corresponding authors.}~,\enspace Yulun Zhang$^{1}$\footnotemark[2],\enspace Xiaokang Yang$^{1}$\\
\textsuperscript{1}Shanghai Jiao Tong University, \enspace
\textsuperscript{2}vivo Mobile Communication Co., Ltd \\
}

\begin{document}
\maketitle

\vspace{-3mm}
\begin{abstract}
Diffusion models have demonstrated impressive performance in face restoration. Yet, their multi-step inference process remains computationally intensive, limiting their applicability in real-world scenarios. Moreover, existing methods often struggle to generate face images that are harmonious, realistic, and consistent with the subject’s identity. In this work, we propose OSDFace, a novel one-step diffusion model for face restoration. Specifically, we propose a visual representation embedder (VRE) to better capture prior information and understand the input face. In VRE, low-quality faces are processed by a visual tokenizer and subsequently embedded with a vector-quantized dictionary to generate visual prompts. Additionally, we incorporate a facial identity loss derived from face recognition to further ensure identity consistency. We further employ a generative adversarial network (GAN) as a guidance model to encourage distribution alignment between the restored face and the ground truth. Experimental results demonstrate that OSDFace surpasses current state-of-the-art (SOTA) methods in both visual quality and quantitative metrics, generating high-fidelity, natural face images with high identity consistency. The code and model will be released at \url{https://github.com/jkwang28/OSDFace}.
\vspace{-3mm}
\end{abstract}

\setlength{\abovedisplayskip}{2pt}
\setlength{\belowdisplayskip}{2pt}

\vspace{-2mm}
\section{Introduction}
\label{sec:intro}
\vspace{-1.5mm}

Face restoration aims to restore high-quality (HQ) face images from degraded low-quality (LQ) inputs caused by complex degradation processes. Blur, noise, downsampling, and JPEG compression are common degradation types that can significantly affect the quality of face images, leading to an ill-posed problem. Recent advancements focus on leveraging CNN-based methods and Transformer-based methods~\cite{zhou2022codeformer,wang2023restoreformer++,xie2024pltrans,tsai2024daefr} to restore HQ face images, and have shown remarkable performance. However, due to the severe degradation of face images, generative models with strong prior knowledge exhibit irreplaceable advantages.

Generative adversarial networks (GANs)~\cite{goodfellow2014gan} and diffusion models~\cite{ho2020denoising,song2021denoising,song2021scorebased} have gained considerable attention for generating high-fidelity images with perceptual details. Many efforts, such as DifFace~\cite{yue2024difface} and DiffBIR~\cite{lin2024diffbir}, have also been applied to restoring HQ face images~\cite{ChenPSFRGAN,wang2021gfpgan,Yang2021GPEN,chan2021glean,miao2024waveface,yang2023pgdiff,chen2023BFRffusion,qiu2023diffbfr,Suin2024CLRFace}. However, GANs are challenging to train and may encounter issues such as mode collapse and training instability. Conversely, diffusion models produce promising results but require multiple forward passes during inference, leading to increased computational cost and longer inference time.

\begin{figure}[t]
\begin{center}
\scriptsize
\scalebox{1}{
    \hspace{-0.4cm}
    \begin{adjustbox}{valign=t}
    \begin{tabular}{cccc}
    \includegraphics[width=0.24\columnwidth]{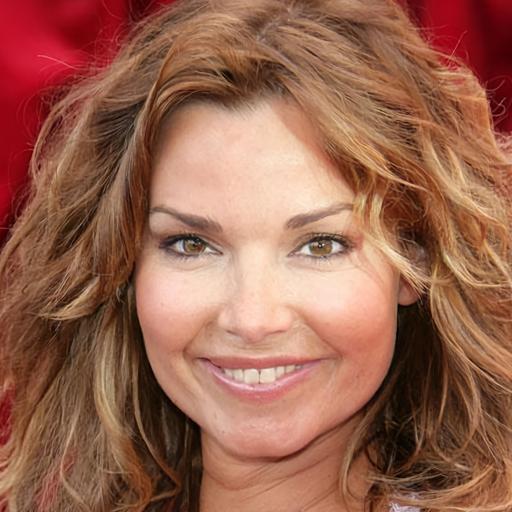} \hspace{-4mm} &
    \includegraphics[width=0.24\columnwidth]{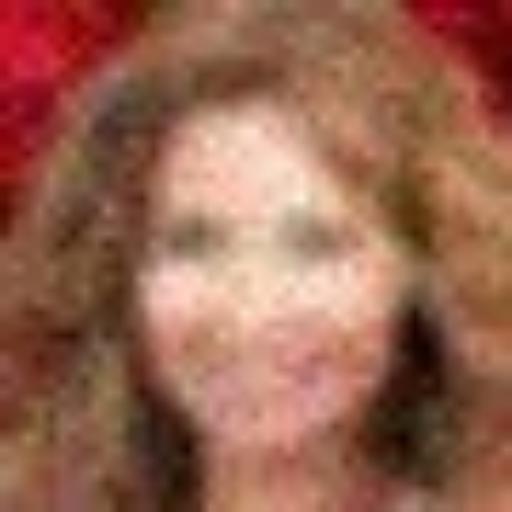} \hspace{-4mm} &
    \includegraphics[width=0.24\columnwidth]{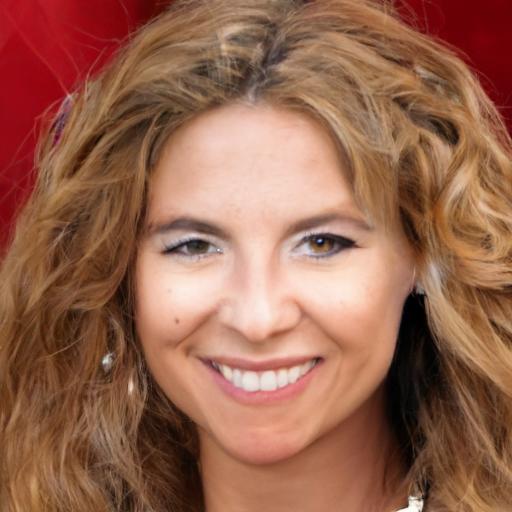} \hspace{-4mm} &
    \includegraphics[width=0.24\columnwidth]{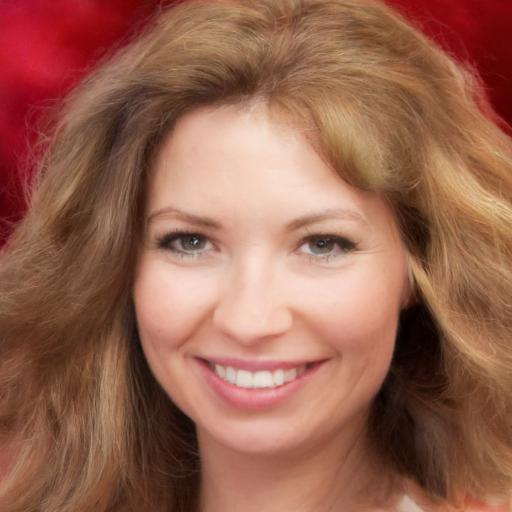} \hspace{-4mm} 
    \vspace{-0.5mm} \\
    HQ (512$\times$512) \hspace{-4mm} &
    LQ (512$\times$512) \hspace{-4mm} &
    PGDiff~\cite{yang2023pgdiff} \hspace{-4mm} &
    DifFace~\cite{yue2024difface} \hspace{-4mm} \\
    \multicolumn{2}{c}{MACs (T) / Time (s) / \# Steps} \hspace{-4mm} & 
    481.0 / 85.8 / 1,000 \hspace{-4mm} & 
    18.68 / 7.05 / 250 \hspace{-4mm} 
    \vspace{-0.2mm} \\
    \includegraphics[width=0.24\columnwidth]{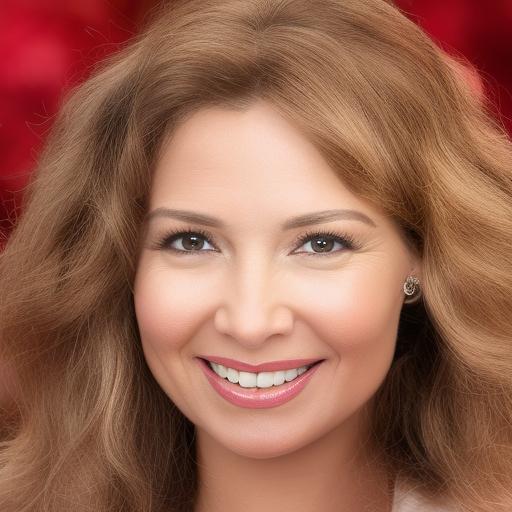} \hspace{-4mm} &
    \includegraphics[width=0.24\columnwidth]{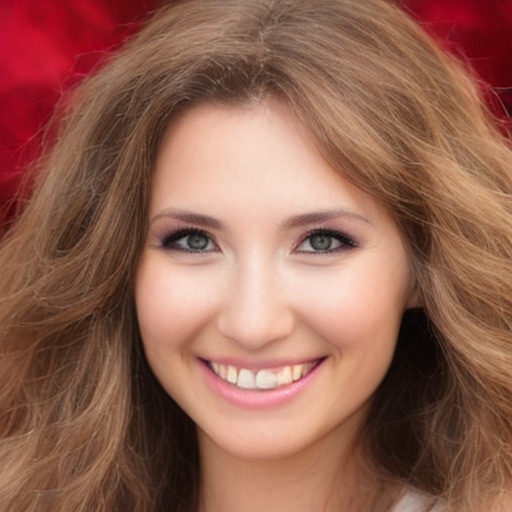} \hspace{-4mm} &
    \includegraphics[width=0.24\columnwidth]{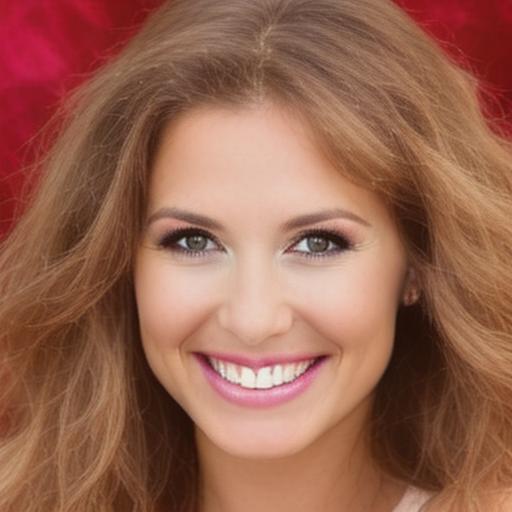} \hspace{-4mm} &
    \includegraphics[width=0.24\columnwidth]{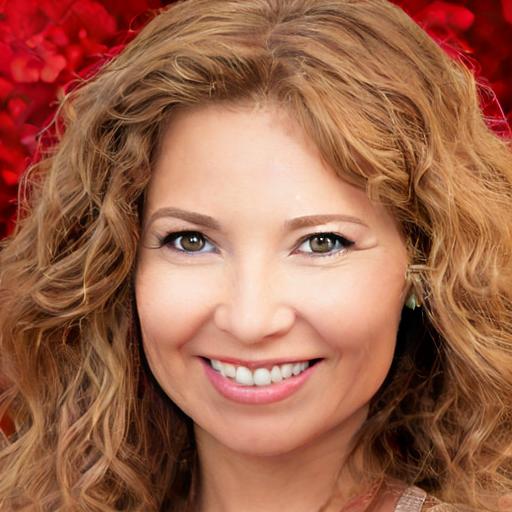} \hspace{-4mm}
    \vspace{-0.5mm} \\
    DiffBIR~\cite{lin2024diffbir} \hspace{-4mm} &
    OSEDiff~\cite{wu2024osediff} \hspace{-4mm} &
    OSEDiff*~\cite{wu2024osediff} \hspace{-4mm} &
    OSDFace (ours) \hspace{-4mm} \\
    189.2 / 9.03 / 50 \hspace{-4mm} & 
    2.269 / 0.13 / 1 \hspace{-4mm} & 
    2.269 / 0.13 / 1 \hspace{-4mm} & 
    2.132 / 0.10 / 1 \hspace{-4mm} 
    \\ 
    \end{tabular}
    \end{adjustbox}
}
\end{center}
\vspace{-7mm}
\caption{Visual samples of diffusion-based face restoration methods. We provide multiply-accumulate operations (MACs), time, and number of timesteps during inference. Our OSDFace achieves a more natural and faithful visual result than other ones.}
\label{fig:page1_compare}
\vspace{-7mm}
\end{figure}

To speed up the diffusion inference, one-step diffusion (OSD) models~\cite{wu2024osediff,li2024dfosd} have emerged as a promising research topic. Benefiting from the recently developed generative diffusion models, especially large-scale pretrained text-to-image (T2I) models~\cite{rombach2022ldm,saharia2022photo}, OSD models~\cite{wu2024osediff,li2024dfosd} enjoy the powerful restoration capability and fast inference speed. It is anticipated that OSD models could be highly competitive in face restoration practical applications. 

As shown in Fig.~\ref{fig:page1_compare}, most diffusion-based face restoration methods can effectively recover basic facial features such as eyes, mouth, and contour. However, details, such as realistic-looking hair and complex backgrounds, often remain unharmonious. The primary issue lies in the insufficient incorporation of face priors~\cite{Yan2024}. Some diffusion-based methods~\cite{yue2024difface, lin2024diffbir, chen2023BFRffusion} rely solely on the diffusion model without considering face priors, resulting in unrealistic restorations. Other methods~\cite{yang2023pgdiff, Suin2024CLRFace} attempt to use face priors, but either limit the generative capabilities or cause information reduction. On the other hand, existing general OSD image restoration models, such as OSEDiff~\cite{wu2024osediff}, are not specifically designed for facial features. OSEDiff employs image-to-tag prior and works well for natural images. However, it meets challenges when applied to faces, as humans are highly sensitive to facial features, and even minor inconsistencies or unrealistic details are easily noticeable.

\begin{figure}[t]
\begin{center}
\includegraphics[width=1.0\columnwidth]{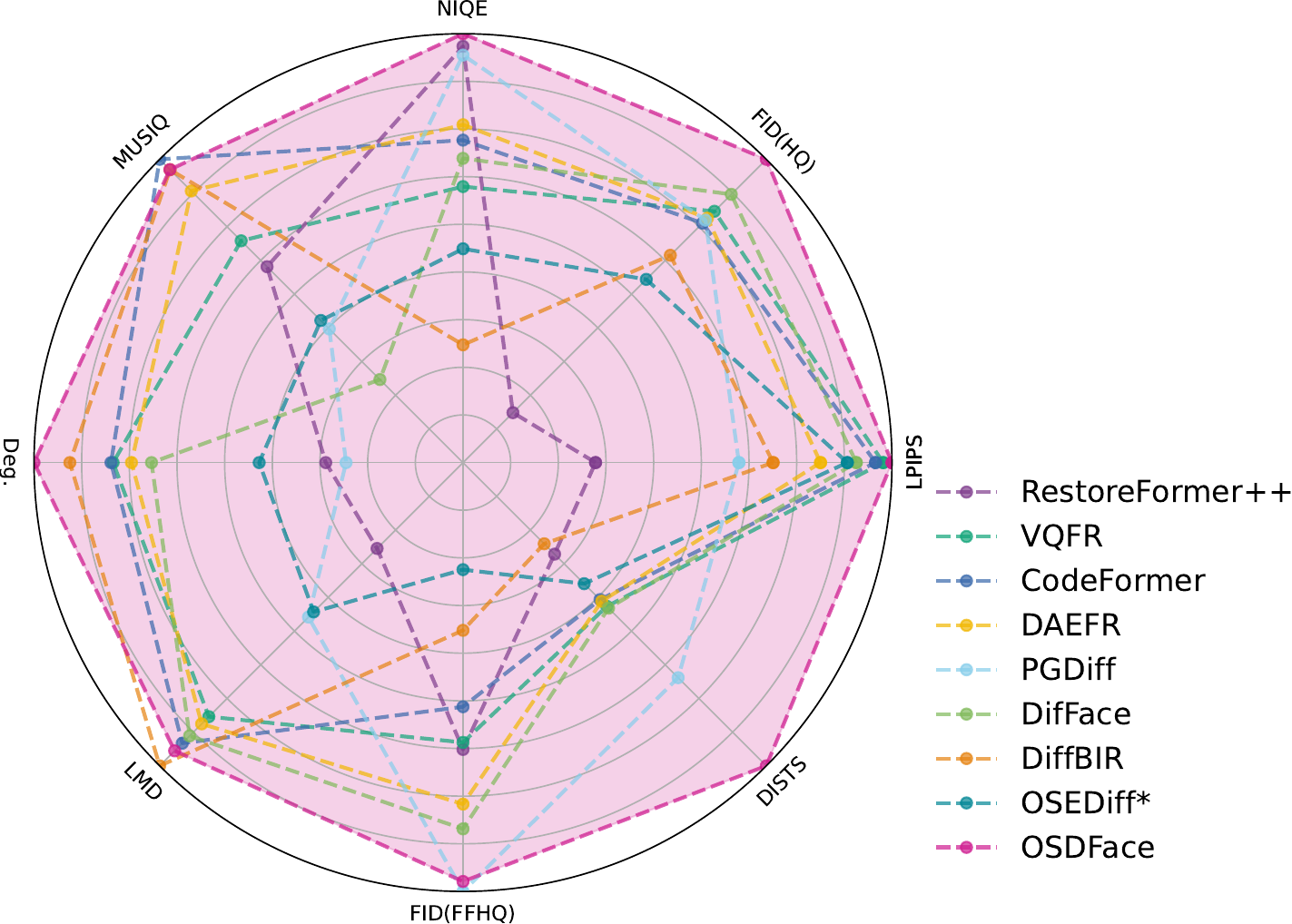}
\end{center}
\vspace{-7mm}
\caption{\small  Performance comparison on the CelebA-Test. Those metrics which smaller scores indicate better image quality, are inverted and normalized for display. OSDFace achieves leading scores on most metrics with only one diffusion step.}
\label{fig:rador}
\vspace{-6mm}
\end{figure}

Thus, it is worthwhile to investigate how to design a prior capturer and apply it to OSD model.
\textbf{One key issue is that} designing priors to capture rich facial features is challenging. Vector-quantized (VQ) prior-based methods, such as CodeFormer~\cite{zhou2022codeformer} and DAEFR~\cite{tsai2024daefr}, effectively leverage codebooks trained in HQ data as priors for face restoration. These models perform well in generating facial features. However, the face images often lack fine details and display unreal backgrounds because they are generated by the codebook items directly.
\textbf{Furthermore}, incorporating priors into diffusion models for face restoration is also a critical focus. Numerous efforts have been undertaken to incorporate face priors. For example, PGDiff~\cite{yang2023pgdiff} employs a face prior by pretraining a VQ-based restorer and using it as a target to guide the diffusion model. Due to limitations imposed by the target model, such a configuration inhibits the generative capabilities. In addition, CLRFace~\cite{Suin2024CLRFace} passes denoised latent vectors to a pretrained codebook and a subsequent decoder. Relying on direct image generation with a fixed-size codebook limits the diversity of results.

To alleviate the limitations, we propose a novel method, OSDFace, for face restoration. \textbf{Firstly}, OSDFace is an OSD model that leverages the powerful image restoration capabilities of diffusion models, as shown in Fig.~\ref{fig:rador}, while offering a fast inference speed (about 0.1s for a 512$\times$512 image). \textbf{Secondly}, our OSDFace integrates a visual representation embedder (VRE) to extract rich facial information directly from LQ inputs. Our proposed VRE consists of two components: a visual tokenizer and a VQ embedder. The visual tokenizer contains a VAE encoder and a VQ dictionary matching function. Trained on LQ data, it could efficiently capture information from LQ faces. In the same LQ semantic space, the VQ embedder performs a dictionary lookup of feature categories to obtain the visual prompt in $\mathcal{O}(1)$ time. \textbf{Thirdly}, to better keep a consistent identity, we incorporate a facial identity loss derived from face recognition into our training process. This approach significantly reduces the distance between the generated faces and the ground truth in the compact deep feature space. Besides, we employ a GAN discriminator as a guidance model, which encourages the distribution alignment between the generated faces and the ground truth. The alignment enables the model suitable for faces in complex environments.  

Our contributions can be summarized as follows.
\begin{itemize}
\item We propose OSDFace, a novel and effective one-step diffusion model for face restoration. This is the first attempt to utilize one-step diffusion for restoring faces. 
\item We design the visual representation embedder (VRE). Using LQ dictionary, VRE captures rich prior from LQ images for a deeper understanding of visual content. 
\item We customize a comprehensive method for realistic face alignment, incorporating facial identity loss for identity consistency and GAN loss for distribution alignment.
\item Our OSDFace achieves significant SOTA face restoration performance, excelling in visual quality and quantitative metrics with reduced computational costs. 
\end{itemize}

\vspace{-1mm}
\section{Related Work}
\label{sec:related}
\vspace{-1mm}
\subsection{Face Restoration}
\vspace{-1mm}
Face restoration aims to recover high-quality (HQ) faces from low-quality (LQ) inputs affected by complex degradations. A key challenge for researchers is how to grasp face prior efficiently and effectively. Statistical priors are widely used in traditional image processing techniques. Recently, with the development of deep learning, researchers have begun to explore more advanced learning-based methods.

\begin{figure*}[t]
\begin{center}
\includegraphics[width=1.0\textwidth]{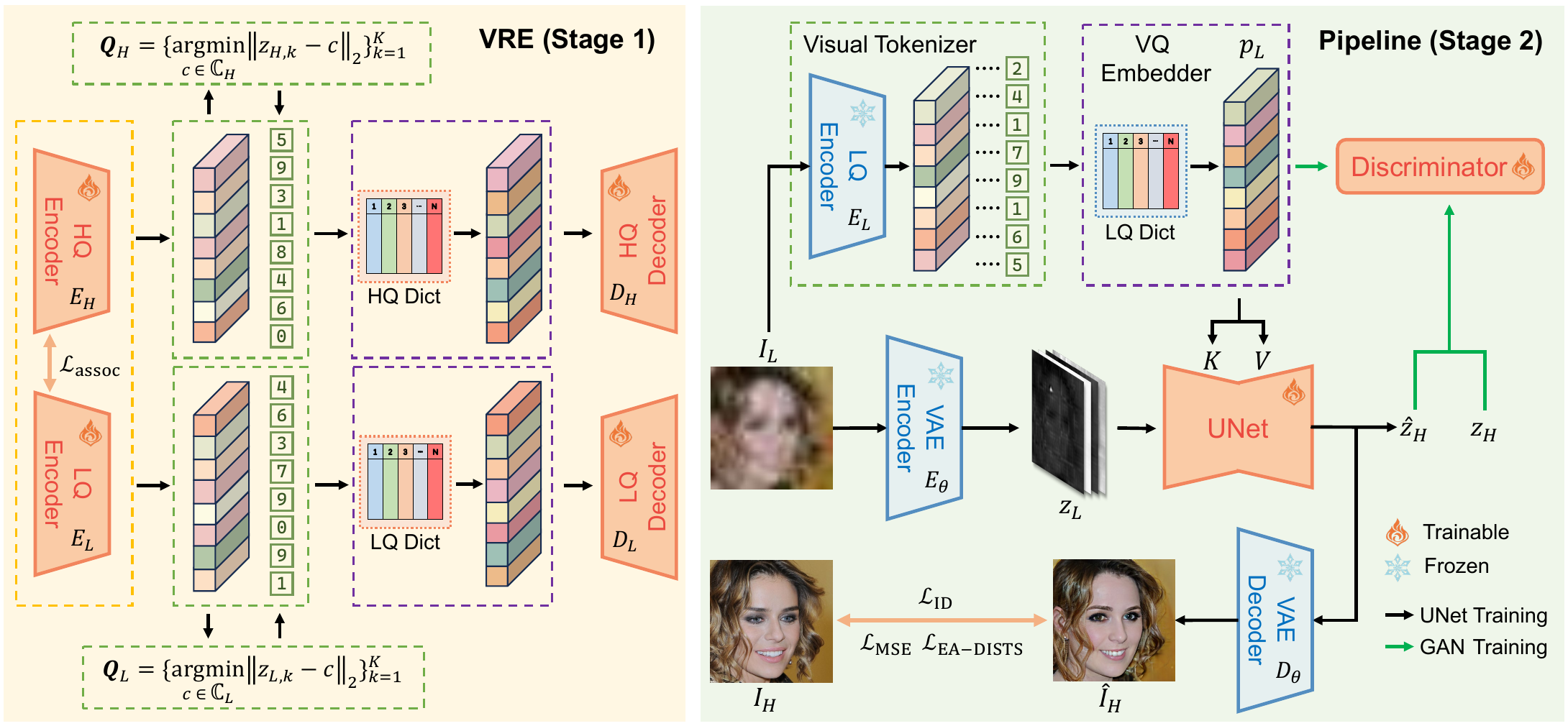}
\end{center}
\vspace{-6.5mm}
\caption{\small Training framework of OSDFace. \textbf{First}, to establish a visual representation embedder (VRE), we train the autoencoder and VQ dictionary for HQ and LQ face domains using self-reconstruction and feature association loss $\mathcal{L}_{\text{assoc}}$. \textbf{Then}, we use the VRE containing LQ encoder and dictionary to embed the LQ face $I_L$, producing the visual prompt embedding $p_L$. \textbf{Next}, the LQ image $I_L$ along with $p_L$ are inputed into the generator $\mathcal{G}_\theta$ to yield the predicted HQ face $\hat{I}_H$: $\hat{I}_H=\mathcal{G}_\theta(I_L; \operatorname{VRE}(I_L))$. The generator $\mathcal{G}_\theta$ incorporates the pretrained VAE and UNet from Stable Diffusion, with only the UNet fine-tuned via LoRA.  \textbf{Additionally}, a series of feature alignment losses are applied to ensure the generation of harmonious and coherent face images. The generator and discriminator are trained alternately.}
\label{fig:overall}
\vspace{-5.5mm}
\end{figure*}

Geometric priors~\cite{yu2018super,chen2018fsrnet,kim2019progressive,shen2018deep}, while useful, are limited by the degree of degradation in low-quality images, restricting their ability to recover high-quality details. Reference priors~\cite{dogan2019exemplar,li2020enhanced,li2018learning,li2020blind} aim to supplement missing information using the corresponding high-quality reference images. However, obtaining references that match the target data is often impractical.
Generative priors~\cite{menon2020pulse,gu2020image,wan2020bringing,yang2020hifacegan,wang2021gfpgan} from pretrained models are widely used for face restoration. These priors are typically incorporated into encoder-decoder networks through iterative latent optimization or direct latent encoding to enhance fidelity.
This approach has demonstrated stronger restoration potential compared to earlier methods. 
Recently, diffusion priors~\cite{kawar2022ddrm,wang2022ddnm,fei2022gdp,yang2023pgdiff} have gained attention with advancements in diffusion models. The degraded images are generated from intermediate outputs at each diffusion iteration. In parallel, VQ prior-based methods~\cite{gu2022vqfr,zhou2022codeformer,wang2023restoreformer++,tsai2024daefr} have shown great promise, achieving impressive results in face restoration. They typically reconstruct images by matching features extracted from LQ inputs to items in a codebook, and then directly output the restored images. However, the limited capacity of VQ priors often leads to blurry backgrounds and unprecise details. Therefore, how to leverage face prior information more effectively remains a crucial problem.

\vspace{-1mm}
\subsection{Diffusion Models}
\vspace{-1mm}

Diffusion models, known for powerful generative capacity, transform random noise into structured data through iterative denoising processes. Moreover, diffusion models have achieved promising results in image restoration recently, showing strong performance in image-to-image tasks using efficient guidance strategies~\cite{lin2024diffbir,miao2024waveface,yue2024difface,chen2023BFRffusion,Suin2024CLRFace,qiu2023diffbfr}. However, some models possess highly complex structures~\cite{Suin2024CLRFace,chen2023BFRffusion}. Most of them require numerous iterative steps, limiting their popularization for real-world applications.

Reducing the inference steps of diffusion models is critical to accelerating generation speeds and reducing computational costs. Yet, further reduction often results in a significant performance drop, making it crucial to balance inference speed with model capability. Most one-step diffusion (OSD) methods employ distillation to learn from a teacher model, ensuring the quality of the generated images~\cite{yin2024dmd,yin2024dmd2,wang2024sinsr,wu2024osediff}. D$^3$SR~\cite{li2024dfosd} achieves promising results with OSD, free from the limitations of model distillation. However, all models are designed primarily for natural image restoration and lack strong generalization for face-specific tasks. It is essential to incorporate prior knowledge specific to human faces to enhance the applicability of OSD models in face restoration. Consequently, the models could generate more realistic and high-fidelity face images.

\vspace{-1mm}
\section{Method}
\label{sec:method}
\vspace{-1mm}

We aim to extract as many features as possible from low-quality (LQ) source images. These features are used as prompts to guide the diffusion model in generating realistic faces that closely resemble the individuals in the original images. To accomplish this, we propose an innovative one-step diffusion model (see Fig.~\ref{fig:overall}) focused on face restoration. This model enhances the realism of the generated faces while preserving the characteristics of LQ images.

\vspace{-1mm}
\subsection{One-Step Diffusion (OSD) Model}
\label{sec:diffusion_models}
\vspace{-1mm}
Latent diffusion models~\cite{rombach2022ldm} are structured around both forward and reverse operations. During the forward diffusion phase, Gaussian noise with a variance $\beta_t \in (0, 1)$ is incrementally added to the latent vector $z$ at each timestep, resulting in
$z_t = \sqrt{\bar{\alpha}_t} \, z + \sqrt{1 - \bar{\alpha}_t} \, \varepsilon,$
where $\varepsilon \sim \mathcal{N}(0, I)$. Here, $\alpha_t$ is defined as $1 - \beta_t$, and $\bar{\alpha}_t$ denotes the cumulative product of ${\alpha}_s$ up to timestep $t$: $\bar{\alpha}_t= \prod_{s=1}^{t} \alpha_s$. 

In the reverse phase, the clean latent vector $\hat{z}_0$ can be estimated directly using the predicted noise $\hat{\varepsilon}$:
\begin{equation}
\hat{z}_0 = \frac{z_t - \sqrt{1 - \bar{\alpha}_t} \, \hat{\varepsilon}}{\sqrt{\bar{\alpha}_t}}. 
\end{equation}
The predicted noise $\hat{\varepsilon}$ could be formally expressed as $ \varepsilon_{\theta}(z_t; p, t)$, where $p$ denotes prompt embedding.

As illustrated in Fig.~\ref{fig:overall}, we first employ the encoder $ E_\theta $ to map the low-quality (LQ) image $ I_L $ 
into the latent space, yielding $ z_L = E_\theta(I_L)$.
Next, we perform one denoising step to obtain the predicted noise $\hat{\varepsilon} $, allowing us to compute the predicted high-quality (HQ) latent vector $ \hat{z}_H $:
\begin{equation}
\hat{z}_H = \frac{z_L - \sqrt{1 - \bar{\alpha}_{T_L}} \varepsilon_{\theta} (z_L; p, T_L)}{\sqrt{\bar{\alpha}_{T_L}}},
\label{eq:generate_zh}
\end{equation}
where $ \varepsilon_\theta $ denotes the denoising network parameterized by $ \theta $, and $ T_L $ is the diffusion timestep.

Unlike one-step T2I diffusion models~\cite{song2023consistency,yin2024dmd}, the UNet input in OSDFace is not entirely Gaussian noise. Therefore, we predefine a parameter $ T_L \in [0, T]$ and feed it into the UNet, where $ T $ represents the total diffusion timesteps (in Stable Diffusion, $ T $ is set to 1,000). Finally, we use the decoder $ D_\theta $ to reconstruct the HQ face $ \hat{I}_H $ from $ \hat{z}_H $:
\begin{equation}
\hat{I}_H = D_\theta(\hat{z}_H).
\end{equation}

Denoting the entire generator as $ \mathcal{G} $, the overall computation procedure can be represented as:
\begin{equation}
\hat{I}_H = \mathcal{G}_\theta(I_L; p).
\end{equation}

\vspace{-1mm}
\subsection{Stage 1. Visual Representation Embedder}
\label{sec:Visual_Representation_Embedder}
\vspace{-1mm}
Previous studies have emphasized the critical importance of understanding input images and integrating face prior into models. To incorporate priors, we first train a feature extraction module, \ie, visual representation embedder (VRE), in stage 1 for LQ inputs. In stage 2, we utilize the pretrained VRE to guide the diffusion model, allowing it to leverage prior knowledge from the input image more effectively.

We design a variational autoencoder (VAE) with a vector-quantized (VQ) dictionary~\cite{oord2017vqvae, esser2021vqgan} for effective feature extraction. The architecture consists of the VRE and a VAE decoder utilized exclusively during the first training stage. Through self-reconstruction training, the VRE builds a dictionary of image categories as priors, enabling it to act as a multi-class embedder using the VQ dictionary.

Given a low-quality (LQ) face $ I_L \in \mathbb{R}^{H \times W \times 3} $, the encoder $ E_L $ processes the image, obtaining a set of feature vector $\{ z_{L,k}\in\mathbb{R}^{d} \}_{k=1}^K$. Then, each feature $ z_L $ is transformed through a matching mechanism, generating a token $q \in \mathbb{N}$ that corresponds to an item in the learnable low-quality VQ dictionary $ \mathbb{C}_L = \{ c_q \in \mathbb{R}^d \}_{q=1}^N $.

\noindent\textbf{Visual Tokenizer.} Tokenization is the process of dividing the input face into smaller units, \ie, tokens. In our framework, the visual tokenizer maps the input face to categories in the low-quality VQ dictionary:
\begin{equation}
\mathcal{M} \circ E_L: \mathbb{R}^{H \times W \times 3} \to \mathbb{N}^K.
\label{eq:visual_encoder}
\end{equation}

For an LQ face denoted as $I_L$, we apply the VAE encoder $ E_L $ followed by the matching function $ \mathcal{M} $ to obtain the token set $ \mathbf{Q}_L = \mathcal{M}(E_L(I_L)) $. The token set $ \mathbf{Q}_L $ represents the predicted categories corresponding to LQ face.

\begin{figure}[t]
\begin{center}
\vspace{1.5mm}
\scalebox{0.68}{
\hspace{-0.4cm}
\begin{adjustbox}{valign=t}
\small 
\begin{tabular}{ccccc}
\includegraphics[width=0.28\columnwidth]{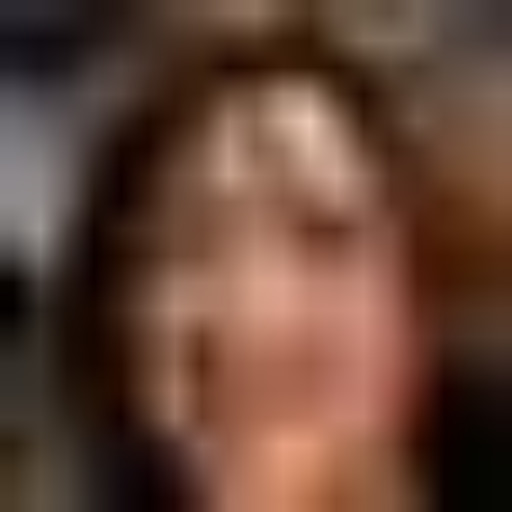} \hspace{-4mm} &
\includegraphics[width=0.28\columnwidth]{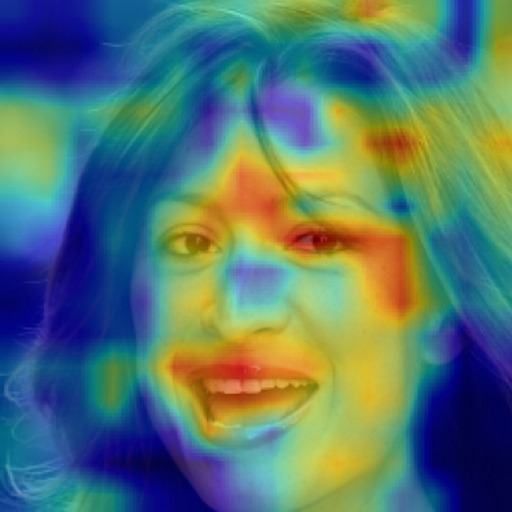} \hspace{-4mm} &
\includegraphics[width=0.28\columnwidth]{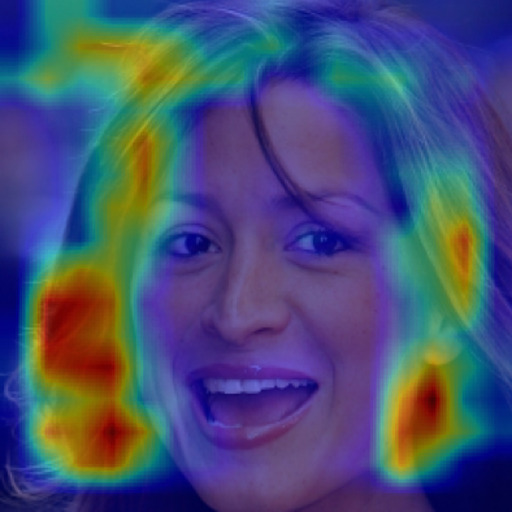} \hspace{-4mm} &
\includegraphics[width=0.28\columnwidth]{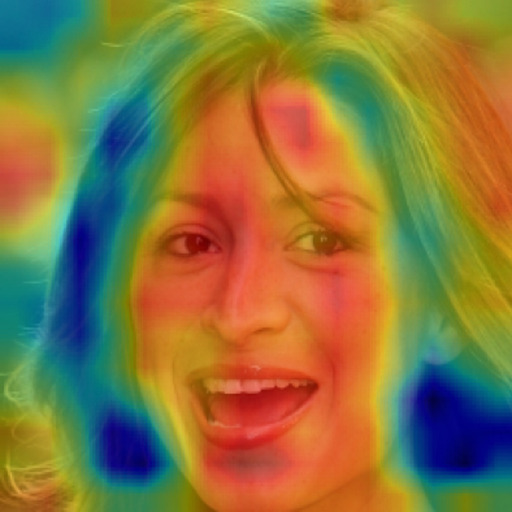} \hspace{-4mm} &
\includegraphics[width=0.28\columnwidth]{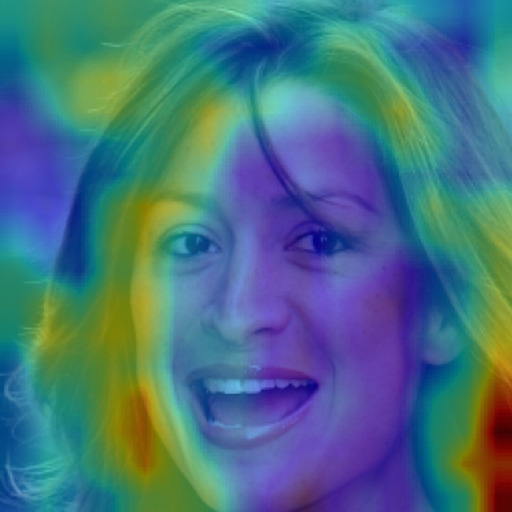} \hspace{-4mm}
\\
   
LQ \hspace{-4mm} &
att\_120 \hspace{-4mm} &
att\_173 \hspace{-4mm} &
enc\_407 \hspace{-4mm} &
enc\_470 \hspace{-4mm}
\\
\includegraphics[width=0.28\columnwidth]{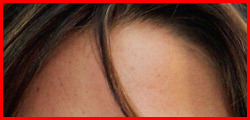} \hspace{-4mm} &
\includegraphics[width=0.28\columnwidth]{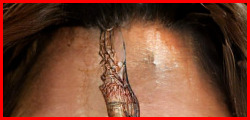} \hspace{-4mm} &
\includegraphics[width=0.28\columnwidth]{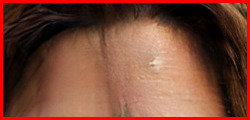} \hspace{-4mm} &
\includegraphics[width=0.28\columnwidth]{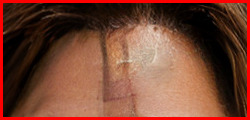} \hspace{-4mm} &
\includegraphics[width=0.28\columnwidth]{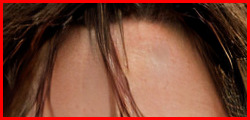} \hspace{-4mm} 
\\
\includegraphics[width=0.28\columnwidth]{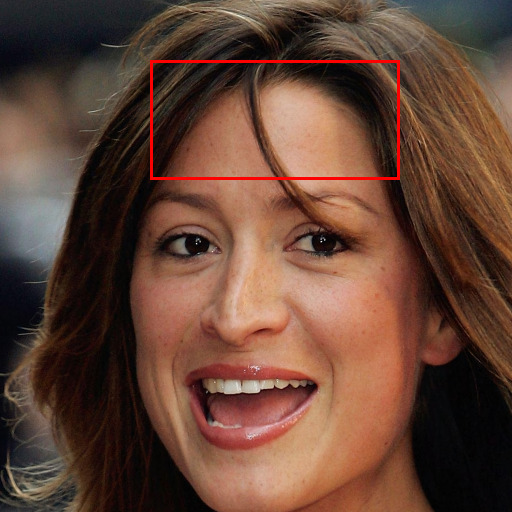} \hspace{-4mm} &
\includegraphics[width=0.28\columnwidth]{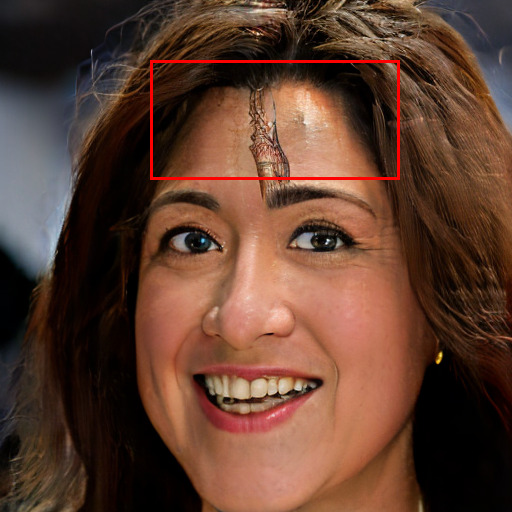} \hspace{-4mm} &
\includegraphics[width=0.28\columnwidth]{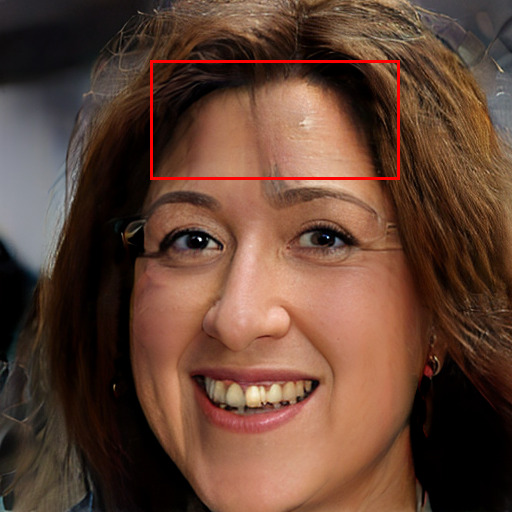} \hspace{-4mm} &
\includegraphics[width=0.28\columnwidth]{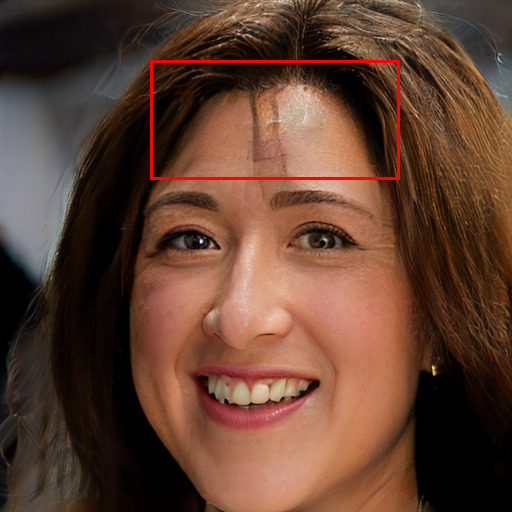} \hspace{-4mm} &
\includegraphics[width=0.28\columnwidth]{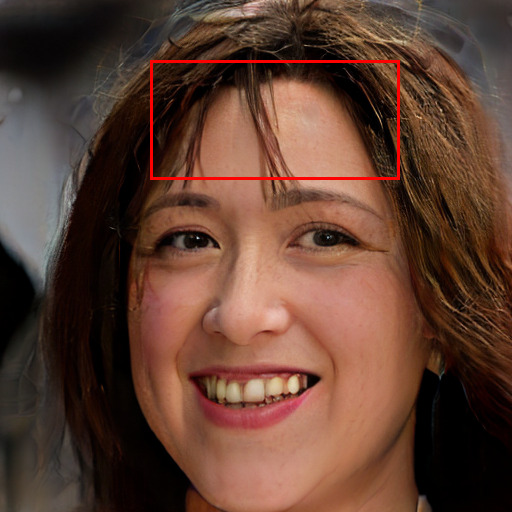} \hspace{-4mm} 
\\

HQ \hspace{-4mm} &
Learnable \hspace{-4mm} &
DAPE~\cite{wu2024seesr} \hspace{-4mm} &
VRE (w/o~assoc)  \hspace{-4mm} &
VRE (w/~assoc) \hspace{-4mm} 
\\
\end{tabular}
\end{adjustbox}
}
\end{center}
\vspace{-6mm}
\caption{Attention maps of VRE and visual comparison of prompt embedding generation. We use CelebA-Test 0262 as an example.}
\label{fig:embedding_ablation}
\vspace{-6.5mm}
\end{figure}

\noindent\textbf{VQ Embedder.} The VQ embedder processes the token set by retrieving the corresponding items from the VQ dictionary. Specifically, for each image token $ q $ at position $ k $, the embedder uses $ q $ as an index to access the dictionary item:
\begin{equation}
z_{k} = \text{dict}(q), \quad q \in \mathbf{Q}_L,
\label{eq:exchange_code_item}
\end{equation}
where $ z_{k} $ represents the $ k $-th element of the visual prompt.

Unlike using image-to-tag models~\cite{radford2021lclip,Zhang2024ram} to generate tags as textual prompts, VRE directly tokenizes each face and converts the tokens into embeddings. This approach eliminates the information loss during the image-tag-embedding process. As illustrated in Fig.~\ref{fig:embedding_ablation}, VRE shows a strong recovery capacity. To study the working mechanism of VRE, we visualize the weight matrix of the final self-attention layer and the latent vector output by the VRE encoder. They are labeled as “att” and “enc” in the figure captions, respectively. The visualizations clearly indicate that the VRE focuses on both facial and nonfacial features. In summary, we can embed the LQ face $I_L$ using VRE to obtain the visual prompt $p_L = \operatorname{VRE}(I_L)$.

We construct two VQ dictionaries corresponding to the HQ and LQ image categories and train VQVAE using vector quantization~\cite{oord2017vqvae} with self-reconstruction. Following VQGAN~\cite{esser2021vqgan}, we use GAN loss to promote extracting real-world features over noise. However, the LQ encoder sometimes focuses on meaningless categories due to strong degradation. Thus, we employ an alignment training strategy inspired by the CLIP model~\cite{radford2021lclip} to align the categories between LQ and HQ faces. By enhancing the diagonal correlation within VQ dictionaries, we could guide the LQ encoder to align its attention with the HQ encoder. 

\noindent\textbf{Training Objectives.} We divide the loss functions into two parts. The first is a series of losses designed specifically for training the VQ dictionary, and the second is for association. Following previous works~\cite{gu2022vqfr,zhou2022codeformer,wang2022restoreformer}, we incorporate four types of losses during training. Three of them are image restoration losses, \ie, absolute differences $\mathcal{L}_1$, perceptual function $\mathcal{L}_{\text{per}}$~\cite{zhang2018lpips, johnson2016perceptualloss}, and discriminative loss $\mathcal{L}_{\text{dis}}$~\cite{isola2017cGan}). Since the VQ is non-differentiable, we adopt a classic quantization loss $\mathcal{L}_{\text{VQ}}$~\cite{bengio2013estimating,oord2017vqvae,esser2021vqgan} additionally. 
\begin{equation}
\begin{aligned}
&\mathcal{L}_1 = \| I_h - I_h^{\text{rec}} \|_1, \mathcal{L}_{\text{per}} = \| \Phi(I_h) - \Phi(I_h^{\text{rec}}) \|_2^2, \\
&\mathcal{L}_{\text{dis}} = \left[ \log D(I_h) + \log(1 - D(I_h^{\text{rec}})) \right], \\
&\mathcal{L}_{\text{VQ}} = \| \text{sg}(Z_h) - Z_h^c \|_2^2 + \beta \| Z_h - \text{sg}(Z_h^c) \|_2^2.
\end{aligned}
\end{equation}

Here, $\Phi(\cdot)$ represents the feature extractor of VGG19 \cite{simonyan2015vgg}, $D$ is a patch-based discriminator \cite{isola2017cGan}, and $\text{sg}(\cdot)$ denotes the stop-gradient operation. $\beta$ is set to 0.25. 

The combined loss $\mathcal{L}_{\text{VQ}}$ is then defined as
\begin{equation}
\mathcal{L}_{\text{VQ}} = \mathcal{L}_1 + \lambda_{\text{per}} \cdot \mathcal{L}_{\text{per}} + \lambda_{\text{dis}} \cdot \mathcal{L}_{\text{dis}} + \mathcal{L}_{\text{VQ}},
\end{equation}
where we set $\lambda_{\text{per}} = 1.0$ and $\lambda_{\text{dis}} = 0.8$ in this setup.

Furthermore, we employ the cross-entropy loss introduced by DAEFR~\cite{tsai2024daefr} to enhance the correlation between the HQ and LQ features. The similarity matrix $M_{\text{assoc}} \in \mathbb{R}^{K\times K}$ is constructed from the encoded feature $z_H$ and $z_L$. We define the similarity score along the HQ axis of $M_{\text{assoc}}$ as $p_{i,j}^h$, and along the LQ axis as $p_{i,j}^l$. The cross-entropy losses are then defined as
\begin{equation}
\begin{aligned}
\mathcal{L}_{\text{CE}}^{H} &= -\frac{1}{K} \sum_{i=1}^K \sum_{j=1}^K y_{i,j} \log(p_{i,j}^h), \\
\mathcal{L}_{\text{CE}}^{L} &= -\frac{1}{K} \sum_{i=1}^K \sum_{j=1}^K y_{i,j} \log(p_{i,j}^l),
\end{aligned}
\end{equation}
where $y_{i,j}$ represents the ground truth label. The final target for the feature association component is then defined as
\begin{equation}
\mathcal{L}_{\text{assoc}} =  \left( \mathcal{L}_{\text{CE}}^{H} + \mathcal{L}_{\text{CE}}^{L} \right) / 2.
\end{equation}

The overall loss function for training the VRE at this stage can be expressed as
\begin{equation}
\mathcal{L}_{\text{total}} = \mathcal{L}_{\text{VQ}} + \lambda_{\text{assoc}} \cdot \mathcal{L}_{\text{assoc}},
\end{equation}
where $\lambda_{\text{assoc}}$ is set to 0 for the initial epochs, and subsequently adjusted to 1 for the remaining epochs.

\vspace{-1mm}
\subsection{Stage 2. Realistic Face Alignment}
\label{sec:loss}
\vspace{-0.75mm}

Face restoration presents unique challenges and higher demands than general image restoration, as humans are exceptionally familiar with and sensitive to faces. Even minor inconsistencies or unnatural details in a generated face image are immediately noticeable (see Fig.~\ref{fig:page1_compare}), often resulting in an unrealistic appearance overall.

To ensure highly realistic face restoration, we design a series of losses aimed at guiding the model to generate globally harmonious and locally coherent face images. These carefully crafted losses play a key role in achieving high fidelity and visual authenticity. The overall loss function for the generator is defined as
\begin{equation}
\begin{aligned}
    \mathcal{L}_\text{gen} =  \lambda_\text{dis}\cdot\mathcal{L}_\mathcal{G}(z_H, \hat{z}_H) +\lambda_\text{ID}\cdot\mathcal{L}_\text{ID}(I_H, \hat{I}_H) \\ 
     + \lambda_\text{per}\cdot\mathcal{L}_{\text{EA-DISTS}}(I_H, \hat{I}_H)+ \operatorname{MSE}(I_H, \hat{I}_H). 
\end{aligned}
\end{equation}
\vspace{0.5mm}

\noindent\textbf{Facial Identity Loss.} A key challenge in face restoration is how to match each LQ face to a real, unique face portrait. However, during reconstruction, the model often struggles to render ambiguous features, leading to disharmonious facial attributes in the generated images. Inspired by advancements in face recognition tasks~\cite{taigman2014deepface,schroff2015facenet,liu2017sphereface}, we introduce a facial identity loss, specifically designed for face restoration. This loss leverages a pretrained face recognition model to measure feature similarity between the generated face and the target HQ image. The facial identity loss helps to enhance both the overall harmony of the generated face and the precise alignment of facial features.

We utilize a pretrained ArcFace model~\cite{deng2019arcface} to encode both the generated and HQ faces into identity embeddings (IDs). The cosine similarity between these IDs is then computed and used as a loss to guide model training. Denote the facial feature extraction model as $\mathcal{F}$. The facial identity loss $\mathcal{L}_\text{ID}$ can be defined as follows
\begin{equation}
\mathcal{L}_\text{ID} = 1 - \cos \left( \mathcal{F}(I_H), \mathcal{F}(\hat{I}_H) \right).
\end{equation}

\noindent\textbf{Perceptual Loss.} Although LPIPS~\cite{zhang2018lpips} is widely used for perceptual evaluation, it can sometimes introduce visible artifacts, especially in diffusion models. To alleviate this issue, we use DISTS~\cite{ding2020dists} to better capture texture details. During the reconstruction of LQ faces, DISTS preserves texture details more effectively while maintaining perceptual similarity, particularly in areas like hair and skin.

In addition, the accurate generation of edge details is essential for achieving perceived sharpness in face restoration. Thus, following the practice in D$^3$SR~\cite{li2024dfosd}, we introduce an edge-aware DISTS (EA-DISTS) as our perceptual loss
\begin{equation}
\begin{aligned}
\mathcal{L}_{\text{EA-DISTS}}(\hat{I}_H, I_H) = \mathcal{L}_{\text{DISTS}}(\hat{I}_H, I_H)& \\
+ ~\mathcal{L}_{\text{DISTS}}(\mathcal{S}(\hat{I}_H), \mathcal{S}(I_H))&,
\end{aligned}
\end{equation}
where $\mathcal{S}(\cdot)$ is the Sobel operator.

\noindent\textbf{GAN Loss.} OSD models face challenges in generating stable images, due to their limited computational capacity. Previous studies~\cite{wang2024sinsr,wu2024osediff} typically employ distillation techniques to transfer knowledge from multi-step diffusion models. However, these methods are restricted by the performance limitations of the teacher models.

As an alternative, we use a discriminative network to enhance the realism of generated faces. This approach provides greater flexibility and improves computational efficiency. The adversarial loss, used to update both the generator $\mathcal{G}_{\theta}$ and discriminator $\mathcal{D}_{\theta}$, is defined as
\begin{equation}
    \mathcal{L}_\mathcal{G} = -\mathbb{E}_t \left[\log \mathcal{D}_{\theta} \left( F(\hat{z}_H, t) \right)\right],
\end{equation}
\begin{equation}
\begin{aligned}
\mathcal{L}_\mathcal{D} = &-\mathbb{E}_t \left[\log \left( 1 - \mathcal{D}_{\theta} \left( F(\hat{z}_H, t) \right) \right)\right] \\
&- \mathbb{E}_t \left[\log \mathcal{D}_{\theta} \left( F({z}_H, t) \right)\right], 
\end{aligned}
\vspace{-1.5mm}
\end{equation}
where $z_H$ is the latent vector of HQ face, and $F(\cdot, t)$ denotes the forward diffusion process at timestep $t \in [0, T]$.

\begin{figure*}[t]

\scriptsize
\centering
\newcommand{\imgid}{00000051}
\newcommand{\imgnote}{0051}
\scalebox{0.98}{
    \hspace{-0.4cm}
    \begin{adjustbox}{valign=t}
\begin{tabular}{ccccccccccc}
    \includegraphics[width=0.088\textwidth]{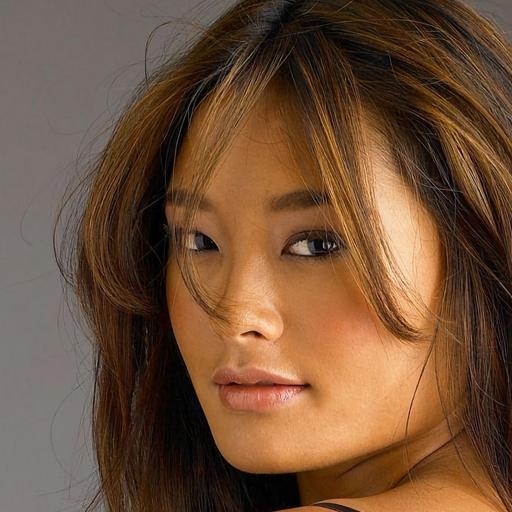} \hspace{-4mm} &
    \includegraphics[width=0.088\textwidth]{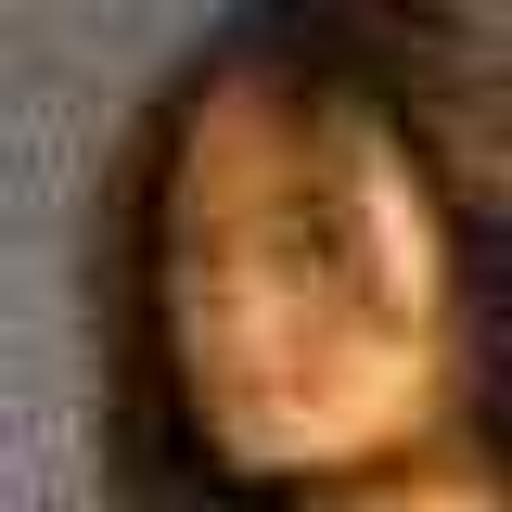} \hspace{-4mm} &
    \includegraphics[width=0.088\textwidth]{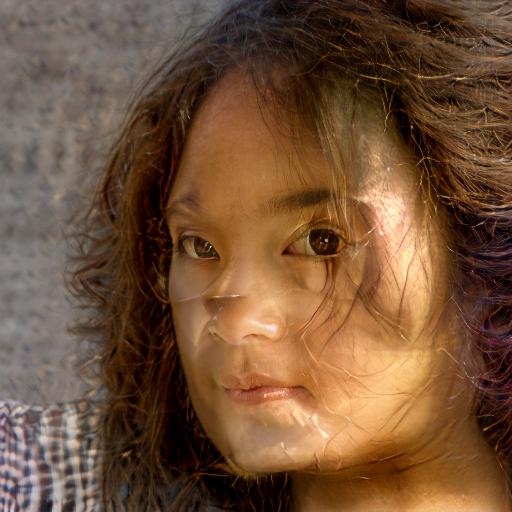} \hspace{-4mm} &
    \includegraphics[width=0.088\textwidth]{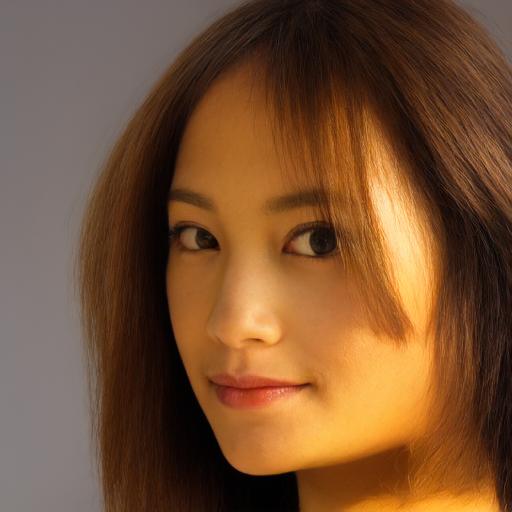} \hspace{-4mm} &
    \includegraphics[width=0.088\textwidth]{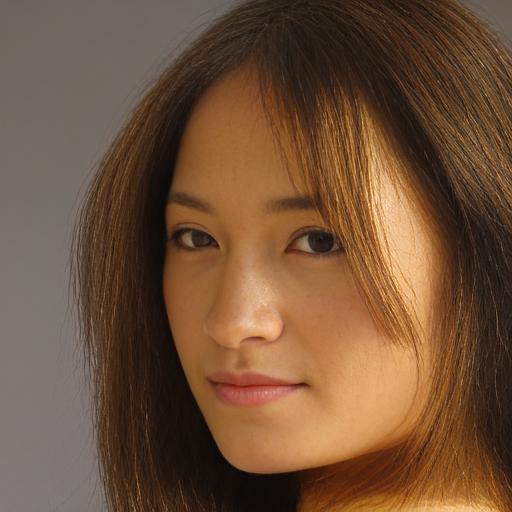} \hspace{-4mm} &
    \includegraphics[width=0.088\textwidth]{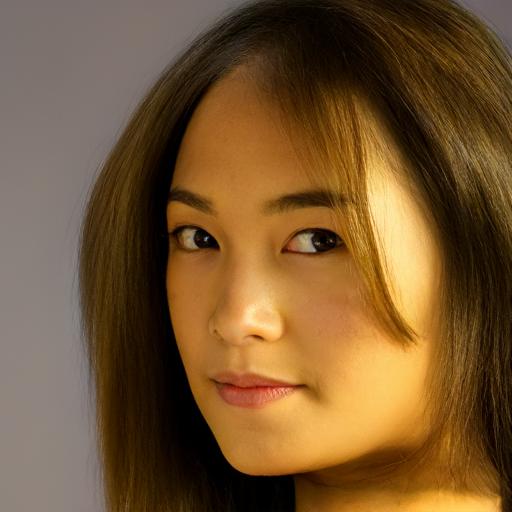} \hspace{-4mm} &
    \includegraphics[width=0.088\textwidth]{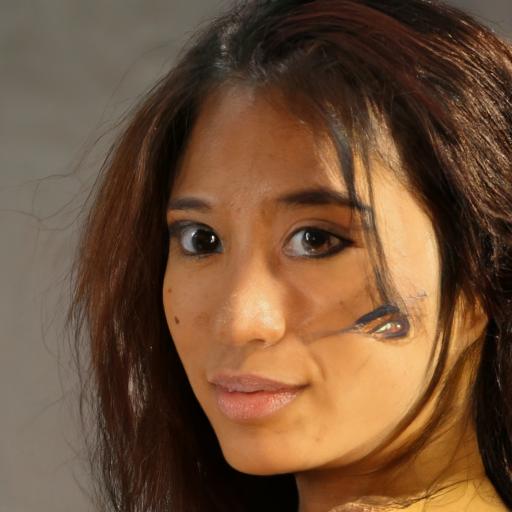} \hspace{-4mm} &
    \includegraphics[width=0.088\textwidth]{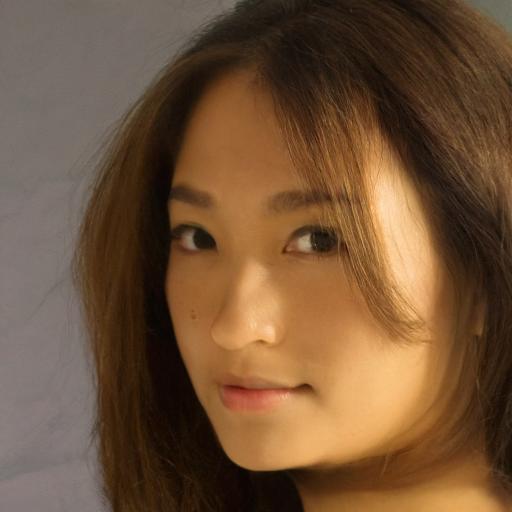} \hspace{-4mm} &
    \includegraphics[width=0.088\textwidth]{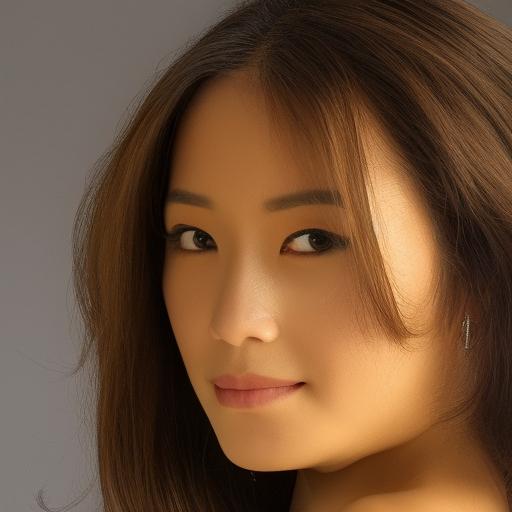} \hspace{-4mm} &
    \includegraphics[width=0.088\textwidth]{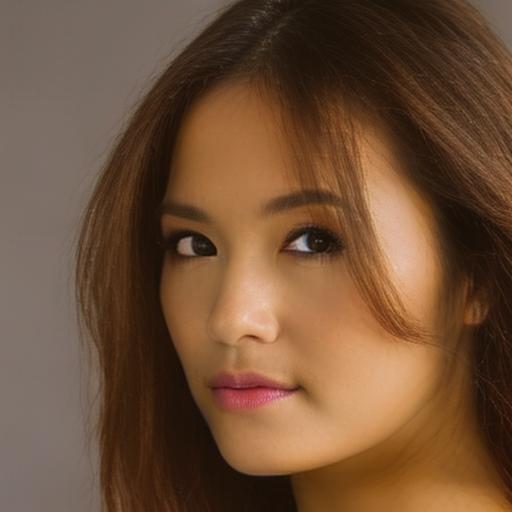} \hspace{-4mm} &
    \includegraphics[width=0.088\textwidth]{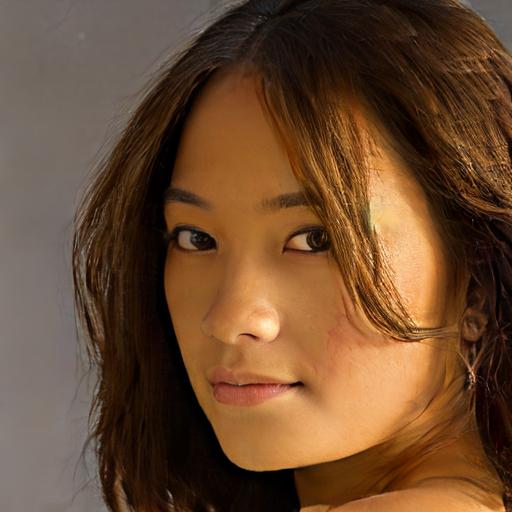} \hspace{-4mm}
    \\
    HQ - \imgnote \hspace{-4mm} &
    LQ - \imgnote \hspace{-4mm} &
    RF++~\cite{wang2023restoreformer++} \hspace{-4mm} &
    VQFR~\cite{gu2022vqfr} \hspace{-4mm} &
    CodeFormer~\cite{zhou2022codeformer} \hspace{-4mm} &
    DAEFR~\cite{tsai2024daefr} \hspace{-4mm} & 
    PGDiff~\cite{yang2023pgdiff} \hspace{-4mm} &
    DifFace~\cite{yue2024difface} \hspace{-4mm} &
    DiffBIR~\cite{lin2024diffbir} \hspace{-4mm} &
    OSEDiff*~\cite{wu2024osediff} \hspace{-4mm} &
    OSDFace (ours) \hspace{-4mm}
    \\
    \end{tabular}
    \end{adjustbox}
}
\renewcommand{\imgid}{00000054}
\renewcommand{\imgnote}{0054}
\scalebox{0.98}{
    
    \hspace{-0.4cm}
    \begin{adjustbox}{valign=t}
    \begin{tabular}{ccccccccccc}
    \includegraphics[width=0.088\textwidth]{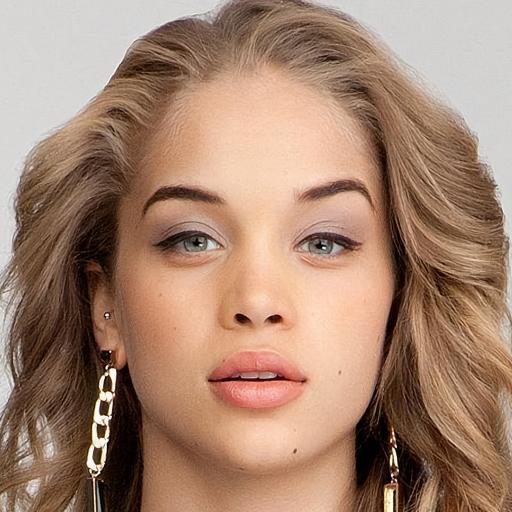} \hspace{-4mm} &
    \includegraphics[width=0.088\textwidth]{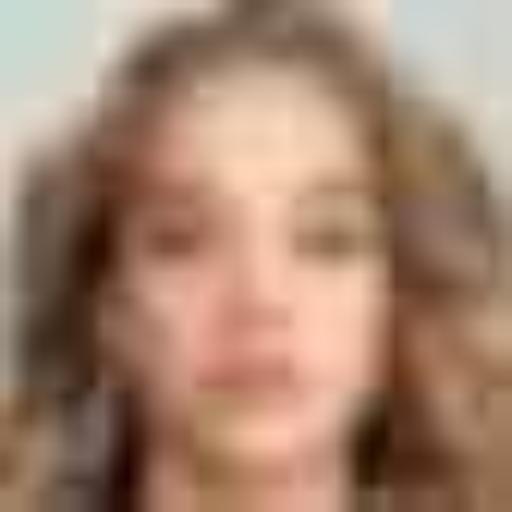} \hspace{-4mm} &
    \includegraphics[width=0.088\textwidth]{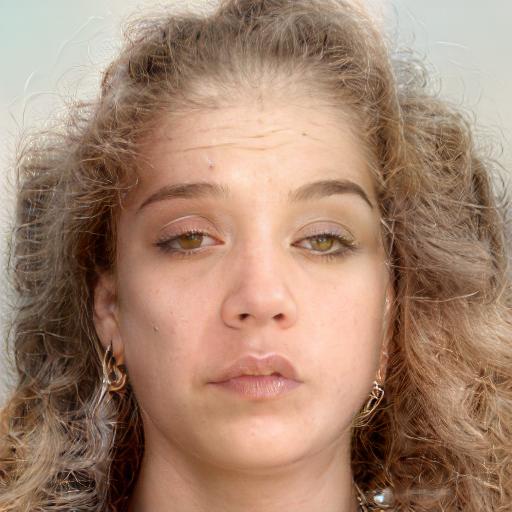} \hspace{-4mm} &
    \includegraphics[width=0.088\textwidth]{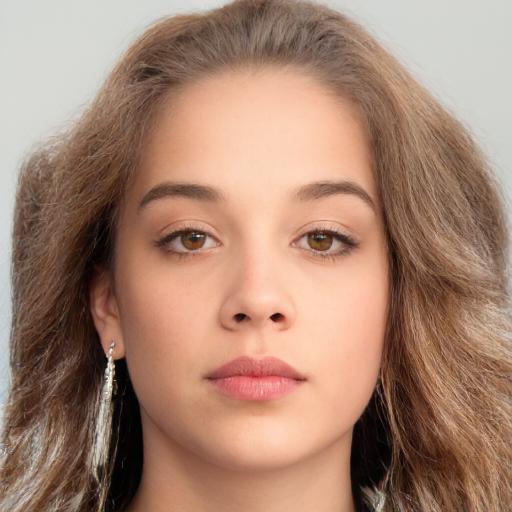} \hspace{-4mm} &
    \includegraphics[width=0.088\textwidth]{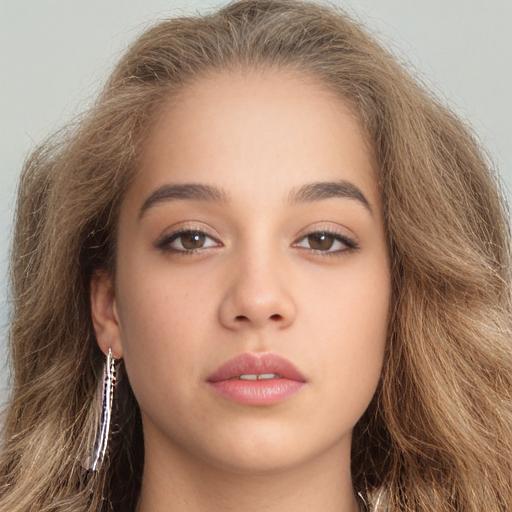} \hspace{-4mm} &
    \includegraphics[width=0.088\textwidth]{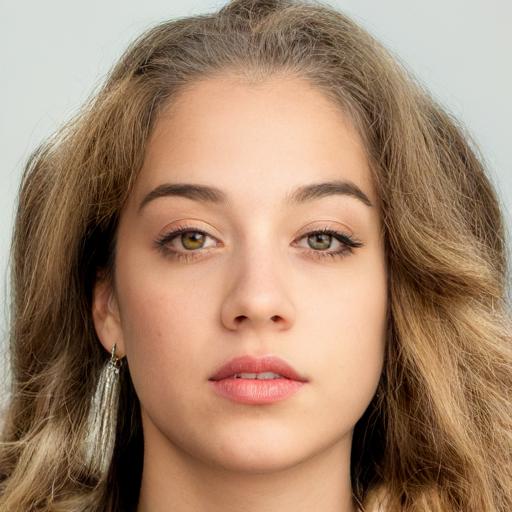} \hspace{-4mm} &
    \includegraphics[width=0.088\textwidth]{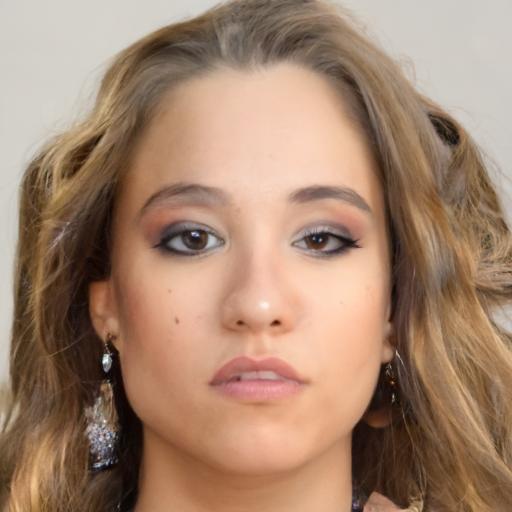} \hspace{-4mm} &
    \includegraphics[width=0.088\textwidth]{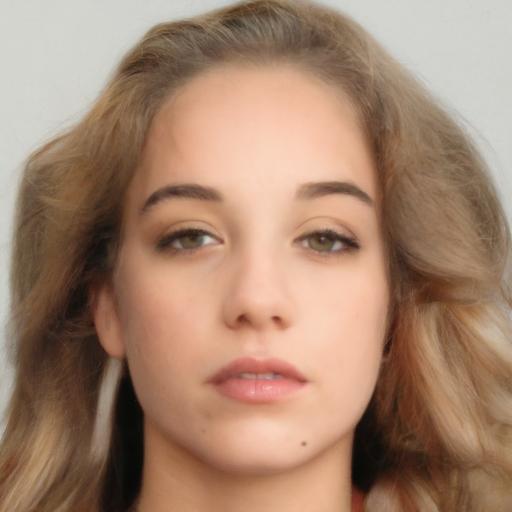} \hspace{-4mm} &
    \includegraphics[width=0.088\textwidth]{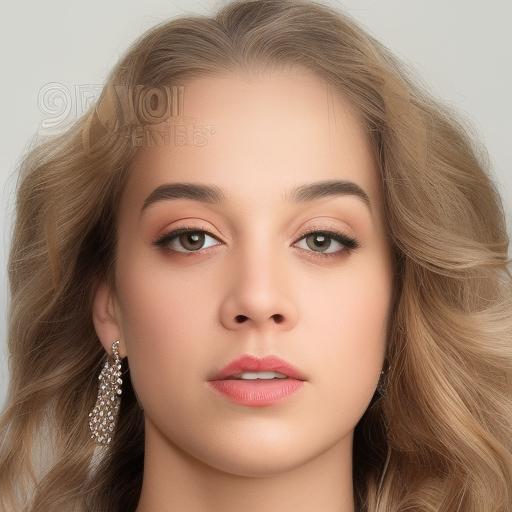} \hspace{-4mm} &
    \includegraphics[width=0.088\textwidth]{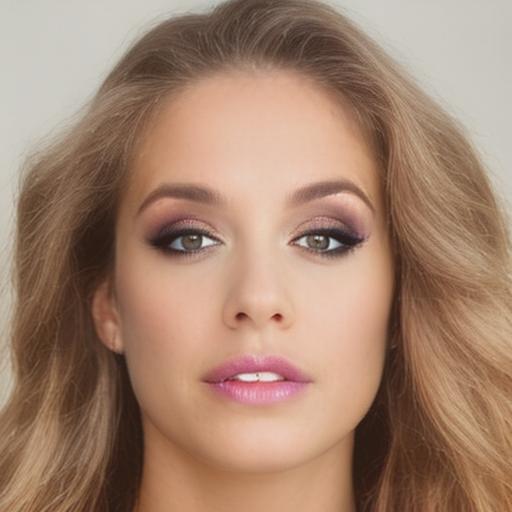} \hspace{-4mm} &
    \includegraphics[width=0.088\textwidth]{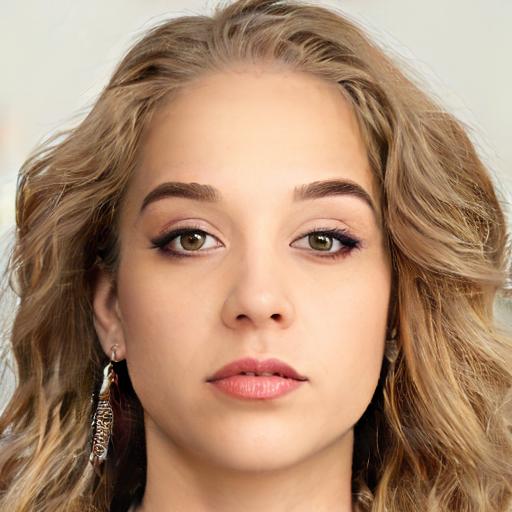} \hspace{-4mm}
    \\
    HQ - \imgnote \hspace{-4mm} &
    LQ - \imgnote \hspace{-4mm} &
    RF++~\cite{wang2023restoreformer++} \hspace{-4mm} &
    VQFR~\cite{gu2022vqfr} \hspace{-4mm} &
    CodeFormer~\cite{zhou2022codeformer} \hspace{-4mm} &
    DAEFR~\cite{tsai2024daefr} \hspace{-4mm} & 
    PGDiff~\cite{yang2023pgdiff} \hspace{-4mm} &
    DifFace~\cite{yue2024difface} \hspace{-4mm} &
    DiffBIR~\cite{lin2024diffbir} \hspace{-4mm} &
    OSEDiff*~\cite{wu2024osediff} \hspace{-4mm} &
    OSDFace (ours) \hspace{-4mm}
    \\
    \end{tabular}
    \end{adjustbox}
}
\renewcommand{\imgid}{00000067}
\renewcommand{\imgnote}{0067}
\scalebox{0.98}{
    
    \hspace{-0.4cm}
    \begin{adjustbox}{valign=t}
    \begin{tabular}{ccccccccccc}
    \includegraphics[width=0.088\textwidth]{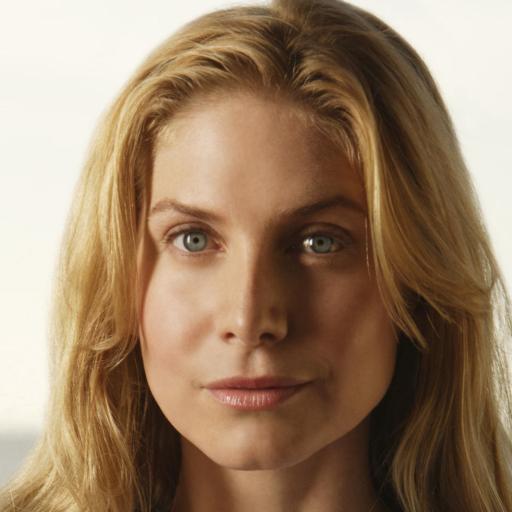} \hspace{-4mm} &
    \includegraphics[width=0.088\textwidth]{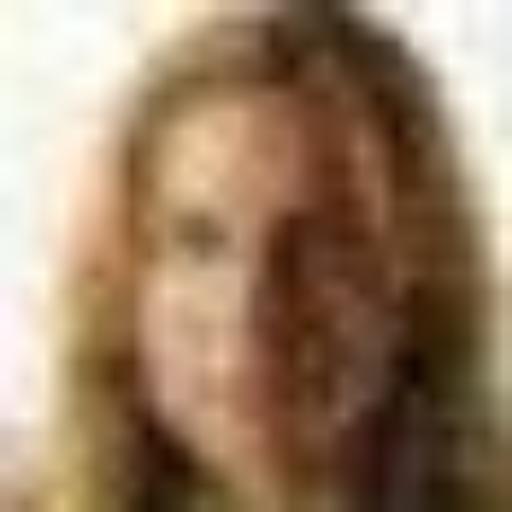} \hspace{-4mm} &
    \includegraphics[width=0.088\textwidth]{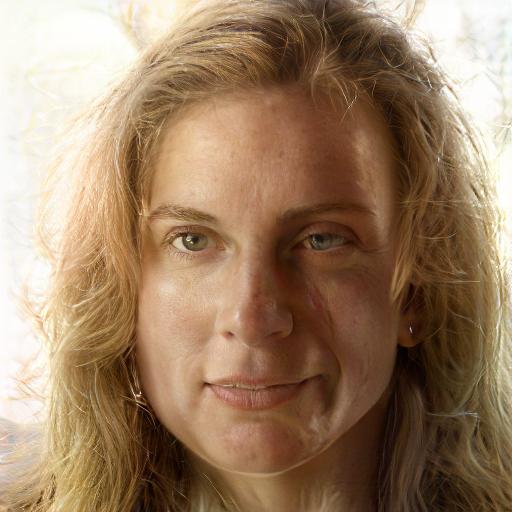} \hspace{-4mm} &
    \includegraphics[width=0.088\textwidth]{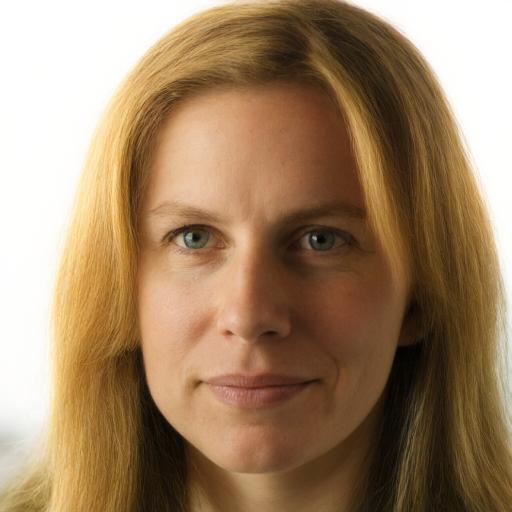} \hspace{-4mm} &
    \includegraphics[width=0.088\textwidth]{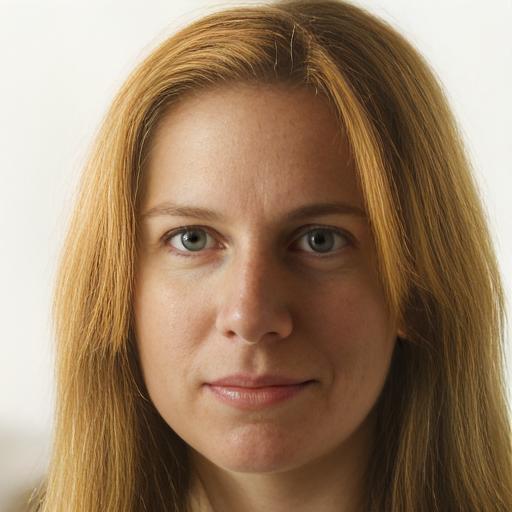} \hspace{-4mm} &
    \includegraphics[width=0.088\textwidth]{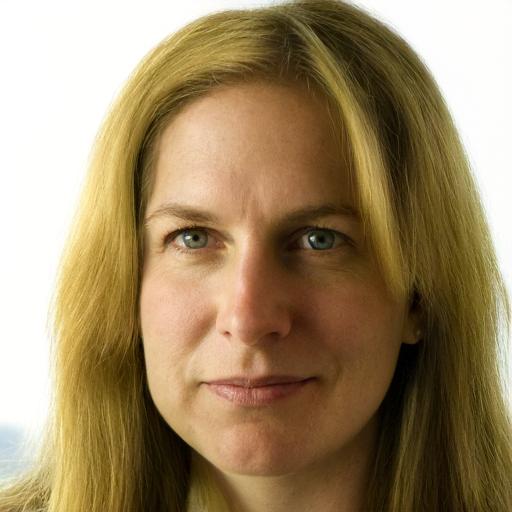} \hspace{-4mm} &
    \includegraphics[width=0.088\textwidth]{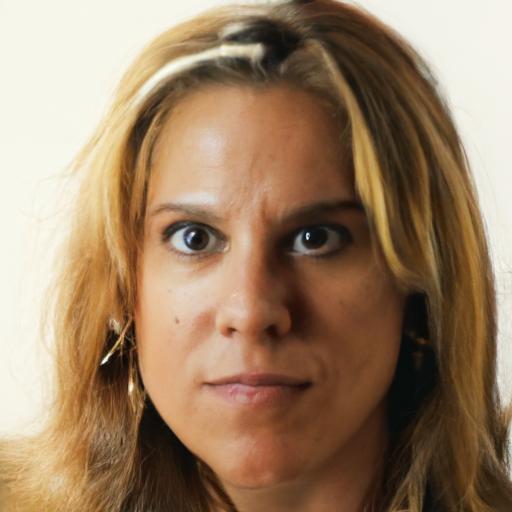} \hspace{-4mm} &
    \includegraphics[width=0.088\textwidth]{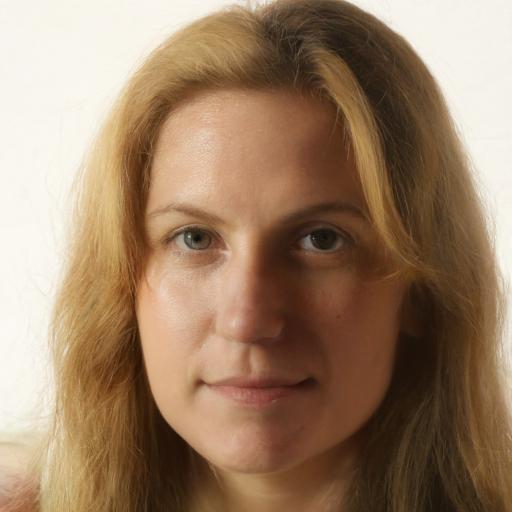} \hspace{-4mm} &
    \includegraphics[width=0.088\textwidth]{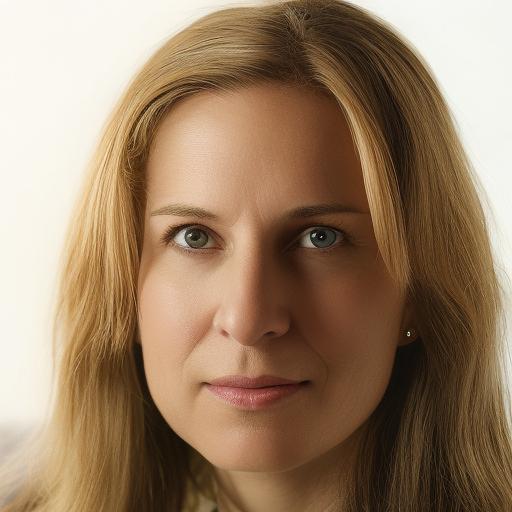} \hspace{-4mm} &
    \includegraphics[width=0.088\textwidth]{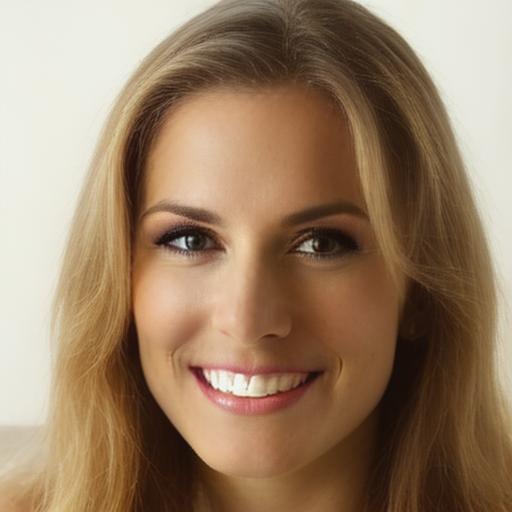} \hspace{-4mm} &
    \includegraphics[width=0.088\textwidth]{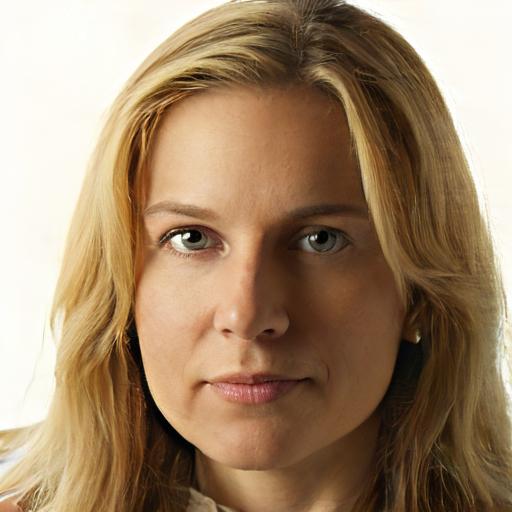} \hspace{-4mm}
    \\
    HQ - \imgnote \hspace{-4mm} &
    LQ - \imgnote \hspace{-4mm} &
    RF++~\cite{wang2023restoreformer++} \hspace{-4mm} &
    VQFR~\cite{gu2022vqfr} \hspace{-4mm} &
    CodeFormer~\cite{zhou2022codeformer} \hspace{-4mm} &
    DAEFR~\cite{tsai2024daefr} \hspace{-4mm} & 
    PGDiff~\cite{yang2023pgdiff} \hspace{-4mm} &
    DifFace~\cite{yue2024difface} \hspace{-4mm} &
    DiffBIR~\cite{lin2024diffbir} \hspace{-4mm} &
    OSEDiff*~\cite{wu2024osediff} \hspace{-4mm} &
    OSDFace (ours) \hspace{-4mm}
    \\
    \end{tabular}
    \end{adjustbox}
}
\vspace{-3.5mm}
\caption{Visual comparison of the synthetic CelebA-Test dataset in challenging cases. Please zoom in for a better view.}
\label{fig:vis-celeba}
\vspace{-3.5mm}
\end{figure*}
\vspace{-8mm}
\begin{figure*}[t]
\begin{minipage}[t]{0.78\textwidth}
    \vspace{-4mm}
    \begin{table}[H]
    \setlength{\tabcolsep}{0.5mm}
    \scriptsize
    \newcolumntype{C}{>{\centering\arraybackslash}X}
    \newcolumntype{Y}{>{\centering\arraybackslash}X}
    
    \begin{center}
    \begin{tabularx}{\textwidth}{l|l|CC|CC|CCCC}
    \toprule[0.15em]
    \rowcolor{color3} Types & Methods & LPIPS$\downarrow$ & DISTS$\downarrow$ & MUSIQ$\uparrow$ & NIQE$\downarrow$ & Deg.$\downarrow$ & LMD$\downarrow$ & FID(FFHQ)$\downarrow$ & FID(HQ)$\downarrow$ \\
    
    \midrule[0.15em]
    \multirow{4}{*}{\makecell{Non-\\Diffusion}} & RestoreFormer{\tiny{++}}~\cite{wang2023restoreformer++}  & 
    0.4535 & 0.2301 & 72.3612 & \textcolor{cvprblue}{3.9524} & 70.5083 & 8.8019 & 57.3723 & 72.7880 \\
    & VQFR~\cite{gu2022vqfr}  &
    0.3390 & 0.2131 & 73.2969 & 4.8374 & 62.2595 & 5.7297 & 58.1044 & 24.9093 \\
    & CodeFormer~\cite{zhou2022codeformer}  &
    0.3412 & 0.2151 & \textcolor{red}{75.9432} & 4.5157 & 62.1972 & 5.3819 & 62.0280 & 26.8595 \\
    & DAEFR~\cite{tsai2024daefr}  &
    0.3580 & 0.2146 & 74.9853 & 4.4161 & 62.8184 & 5.6332 & 52.0341 & 26.0458 \\
    \midrule
    \multirow{3}{*}{\makecell{Multi-Step\\Diffusion}}&PGDiff~\cite{yang2023pgdiff} (s=1,000)  & 
    0.3866 & \textcolor{cvprblue}{0.1949} & 69.5676 & 4.0010 & 71.5996 & 7.3109 & \textcolor{red}{44.6258} & 26.3694 \\
    &DifFace~\cite{yue2024difface} (s=250) & 
    0.3469 & 0.2126 & 66.7451 & 4.6381 & 63.4511 & 5.4759 & 49.8075 & \textcolor{cvprblue}{22.2370} \\
    &DiffBIR~\cite{lin2024diffbir} (s=50) &
    0.3740 & 0.2340 & 75.6360 & 6.2801 & \textcolor{cvprblue}{61.0238} & \textcolor{red}{5.1042} & 71.7767 & 32.5109 \\
\end{tabularx}
    \end{center}
    \vspace{-6.5mm}
    \caption{Quantitative comparison on the synthetic CelebA-Test dataset with non-diffusion, multi-step diffusion, and one-step diffusion methods. The best and second best results are colored with \textcolor{red}{red} and \textcolor{cvprblue}{blue}. }
    \vspace{-5.5mm}
    \label{table:CelebA-Test}
    \end{table}
\end{minipage}%
\hfill
\begin{minipage}[t]{0.20\textwidth}
    \vspace{-3mm}
    \begin{figure}[H]
    \begin{center}
    \hspace{-0.53cm}
    \begin{adjustbox}{valign=t}
    \scriptsize 
    \begin{tabular}{cc}
    \includegraphics[width=0.44\columnwidth]{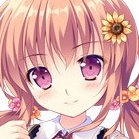} \hspace{-3.6mm} &
    \includegraphics[width=0.44\columnwidth]{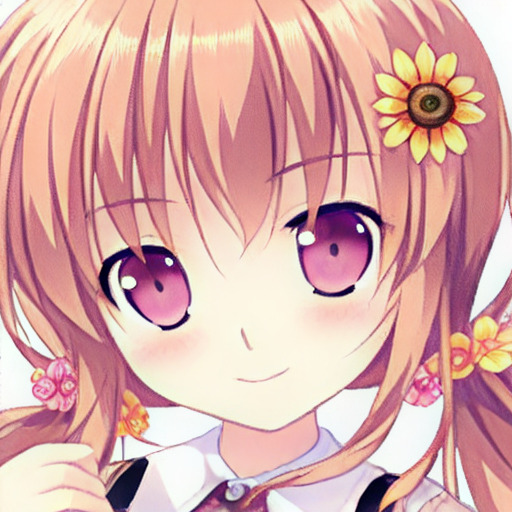} \hspace{-4.6mm} \\
    LQ \hspace{-3.6mm} & OSDFace \hspace{-4.6mm} \\
    \includegraphics[width=0.44\columnwidth]{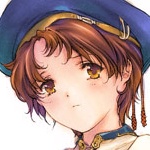} \hspace{-3.6mm} &
    \includegraphics[width=0.44\columnwidth]{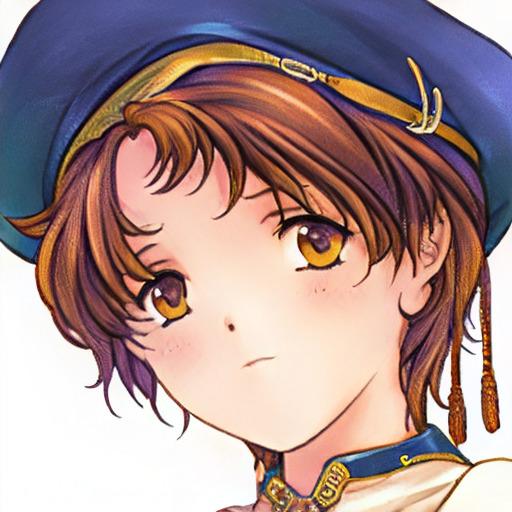} \hspace{-4.6mm} \\ 
    LQ \hspace{-3.6mm} & OSDFace \hspace{-4.6mm} \\
    \end{tabular}
    \end{adjustbox}
    \end{center}
    \vspace{-6.8mm}
    \caption{Zero-shot results of real-world cartoons. }
    \label{fig:cartoon}
    \vspace{-4mm}
    \end{figure}
\end{minipage}
\end{figure*}

\begin{figure*}[t]

\scriptsize
\centering
\scalebox{0.98}{
    \hspace{-0.4cm}
    \begin{adjustbox}{valign=t}
    \begin{tabular}{cccccccccc}
    \includegraphics[width=0.098\textwidth]{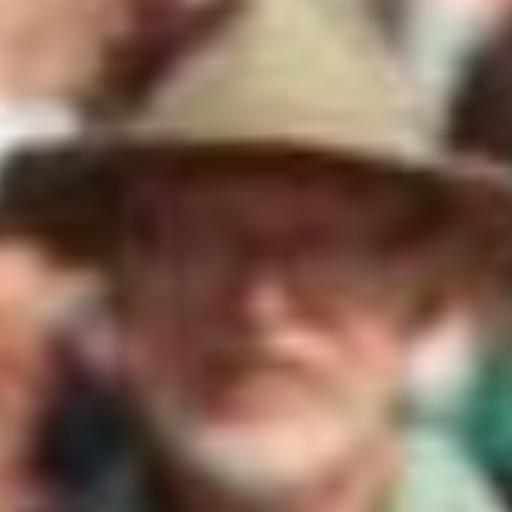} \hspace{-4mm} &
    \includegraphics[width=0.098\textwidth]{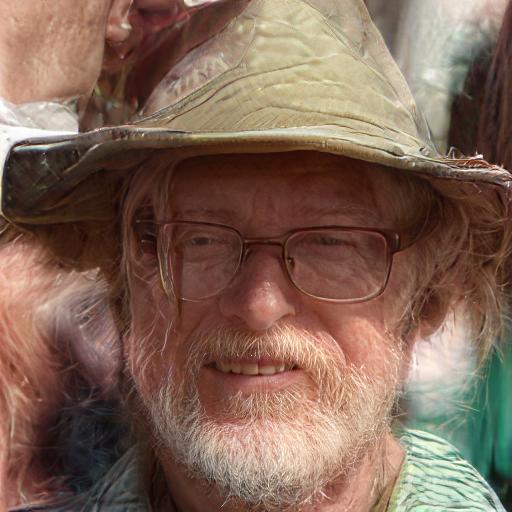} \hspace{-4mm} &
    \includegraphics[width=0.098\textwidth]{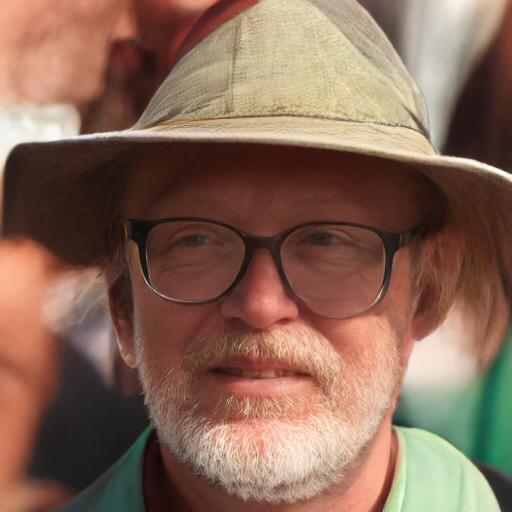} \hspace{-4mm} &
    \includegraphics[width=0.098\textwidth]{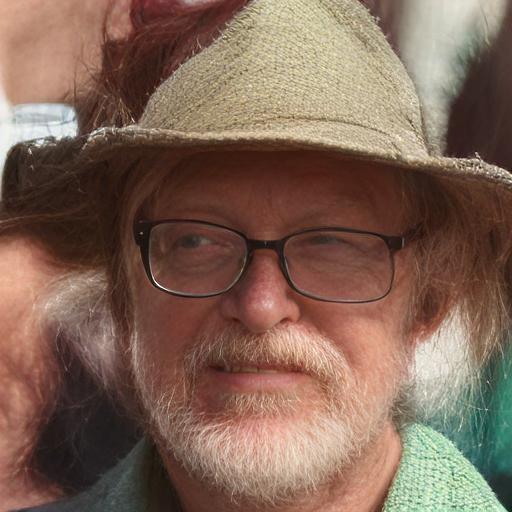} \hspace{-4mm} &
    \includegraphics[width=0.098\textwidth]{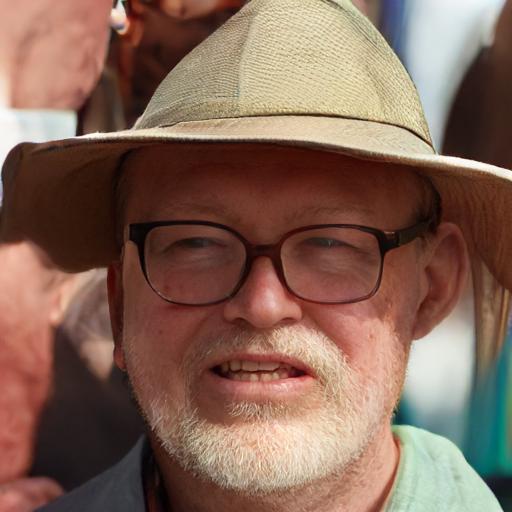} \hspace{-4mm} &
    \includegraphics[width=0.098\textwidth]{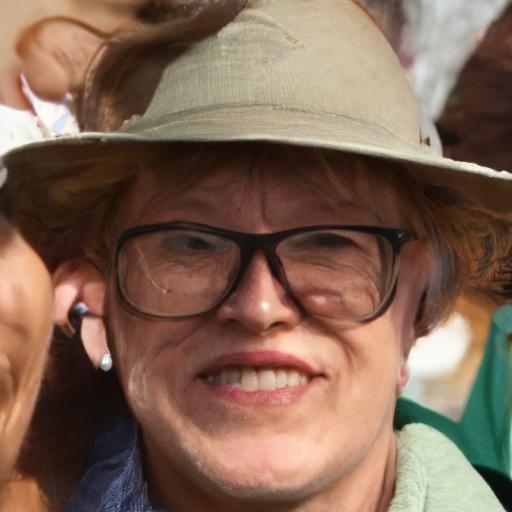} \hspace{-4mm} &
    \includegraphics[width=0.098\textwidth]{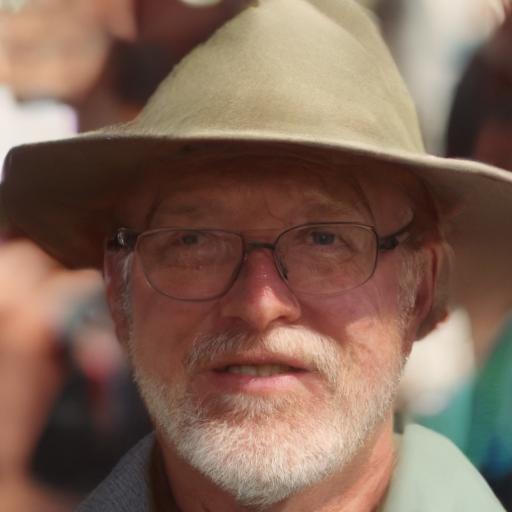} \hspace{-4mm} &
    \includegraphics[width=0.098\textwidth]{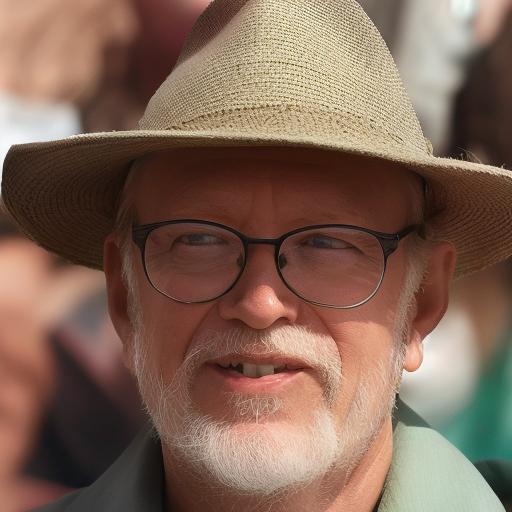} \hspace{-4mm} &
    \includegraphics[width=0.098\textwidth]{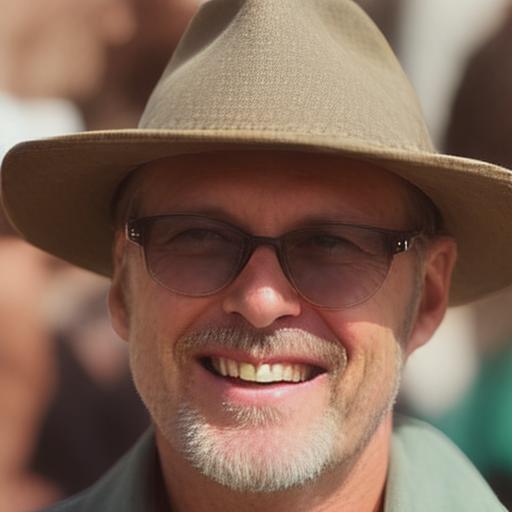} \hspace{-4mm} &
    \includegraphics[width=0.098\textwidth]{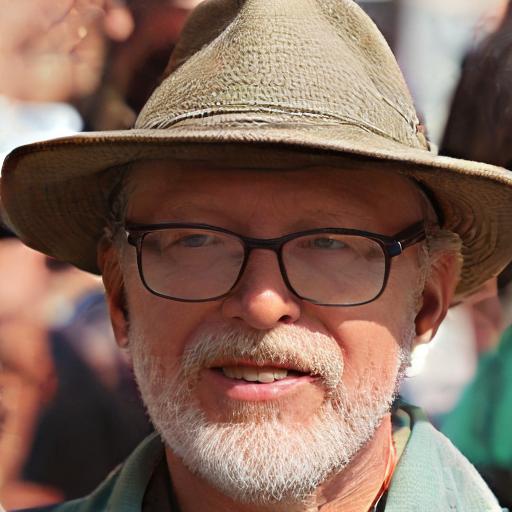} \hspace{-4mm} 
    \\
    Wider 0011 \hspace{-4mm} &
    RF++~\cite{wang2023restoreformer++} \hspace{-4mm} &
    VQFR~\cite{gu2022vqfr} \hspace{-4mm} &
    CodeFormer~\cite{zhou2022codeformer} \hspace{-4mm} &
    DAEFR~\cite{tsai2024daefr} \hspace{-4mm} & 
    PGDiff~\cite{yang2023pgdiff} \hspace{-4mm} &
    DifFace~\cite{yue2024difface} \hspace{-4mm} &
    DiffBIR~\cite{lin2024diffbir} \hspace{-4mm} &
    OSEDiff*~\cite{wu2024osediff} \hspace{-4mm} &
    OSDFace (ours) \hspace{-4mm}
    \\
    \end{tabular}
    \end{adjustbox}
}
\scalebox{0.98}{

    \hspace{-0.4cm}
    \begin{adjustbox}{valign=t}
    \begin{tabular}{cccccccccc}
    \includegraphics[width=0.098\textwidth]{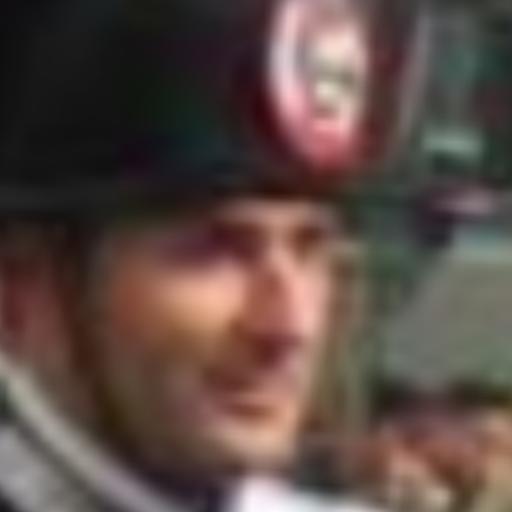} \hspace{-4mm} &
    \includegraphics[width=0.098\textwidth]{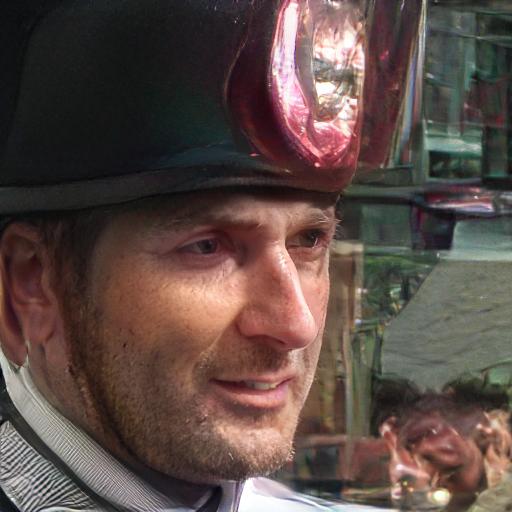} \hspace{-4mm} &
    \includegraphics[width=0.098\textwidth]{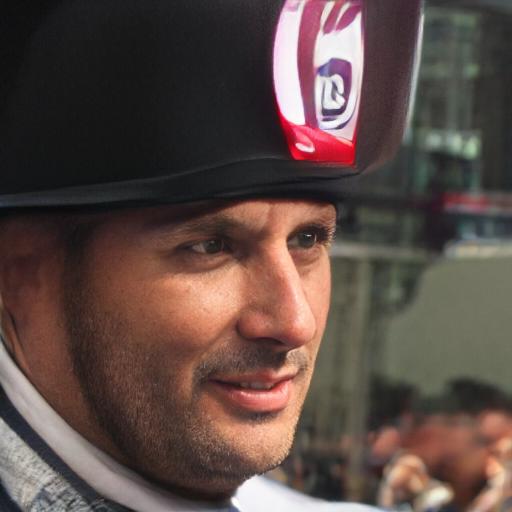} \hspace{-4mm} &
    \includegraphics[width=0.098\textwidth]{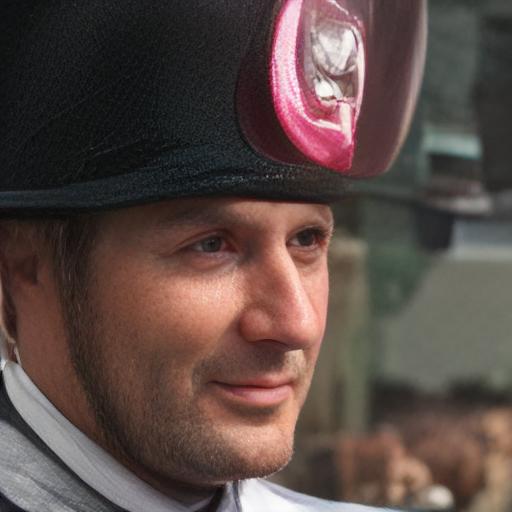} \hspace{-4mm} &
    \includegraphics[width=0.098\textwidth]{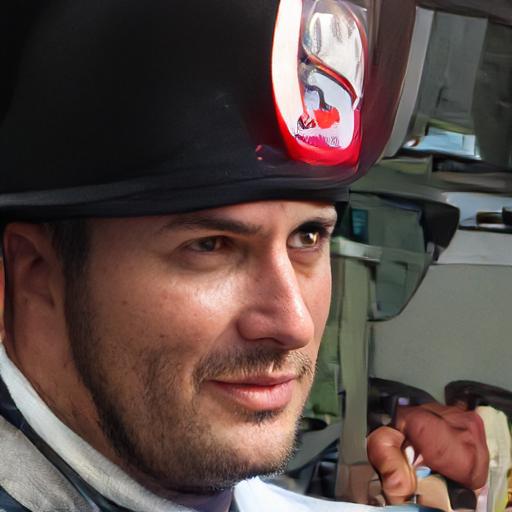} \hspace{-4mm} &
    \includegraphics[width=0.098\textwidth]{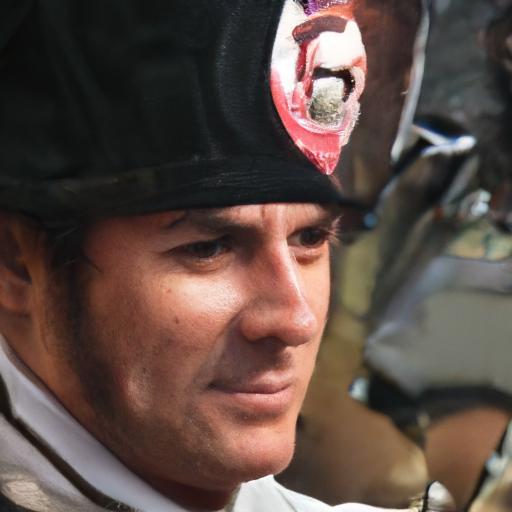} \hspace{-4mm} &
    \includegraphics[width=0.098\textwidth]{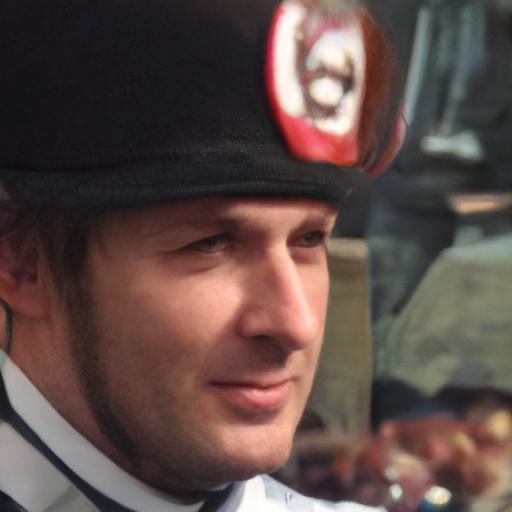} \hspace{-4mm} &
    \includegraphics[width=0.098\textwidth]{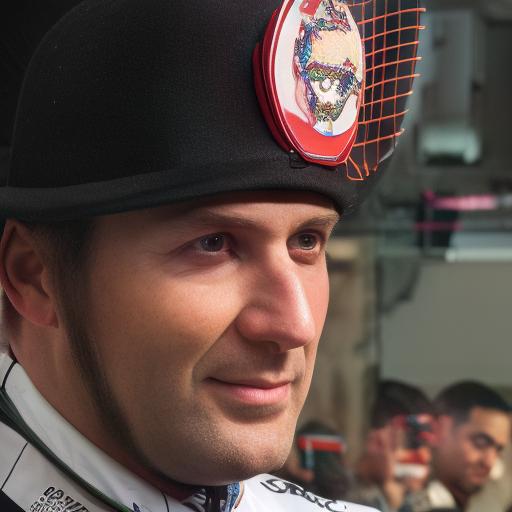} \hspace{-4mm} &
    \includegraphics[width=0.098\textwidth]{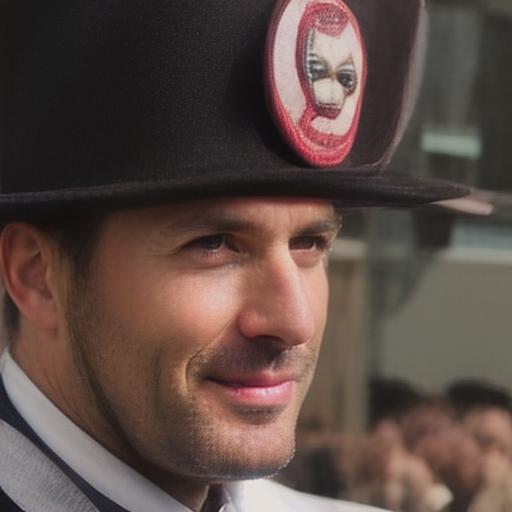} \hspace{-4mm} &
    \includegraphics[width=0.098\textwidth]{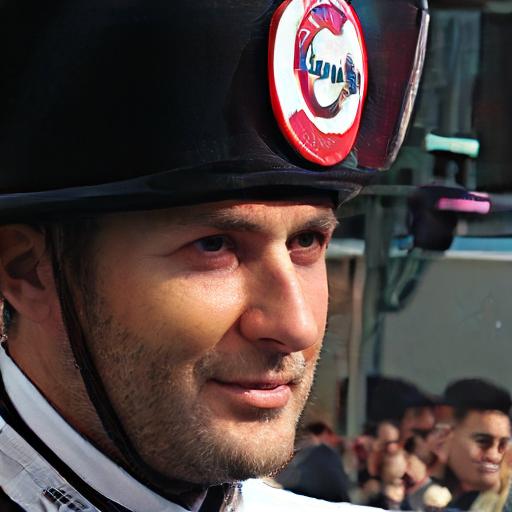} \hspace{-4mm} 
    \\
    Wider 0025 \hspace{-4mm} &
    RF++~\cite{wang2023restoreformer++} \hspace{-4mm} &
    VQFR~\cite{gu2022vqfr} \hspace{-4mm} &
    CodeFormer~\cite{zhou2022codeformer} \hspace{-4mm} &
    DAEFR~\cite{tsai2024daefr} \hspace{-4mm} & 
    PGDiff~\cite{yang2023pgdiff} \hspace{-4mm} &
    DifFace~\cite{yue2024difface} \hspace{-4mm} &
    DiffBIR~\cite{lin2024diffbir} \hspace{-4mm} &
    OSEDiff*~\cite{wu2024osediff} \hspace{-4mm} &
    OSDFace (ours) \hspace{-4mm}
    \\
    \end{tabular}
    \end{adjustbox}
}
\scalebox{0.98}{
    \hspace{-0.4cm}
    \begin{adjustbox}{valign=t}
    \begin{tabular}{cccccccccc}
    \includegraphics[width=0.098\textwidth]{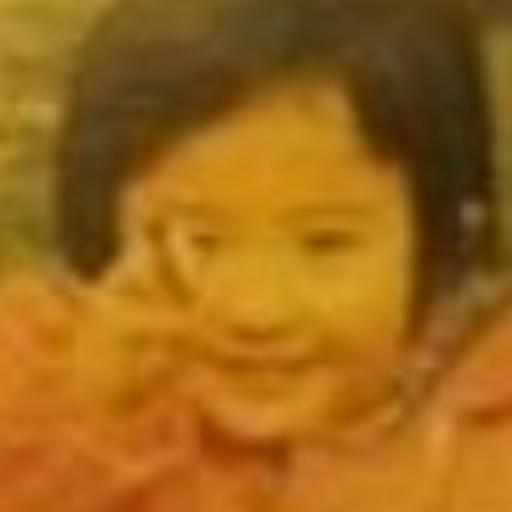} \hspace{-4mm} &
    \includegraphics[width=0.098\textwidth]{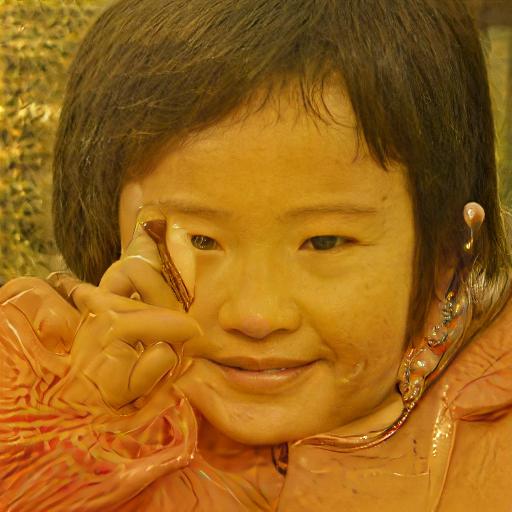} \hspace{-4mm} &
    \includegraphics[width=0.098\textwidth]{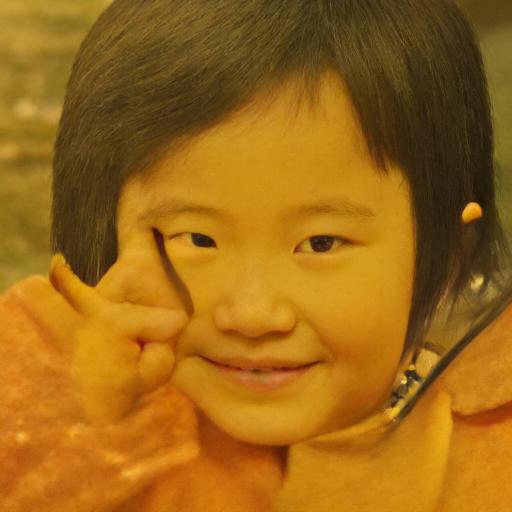} \hspace{-4mm} &
    \includegraphics[width=0.098\textwidth]{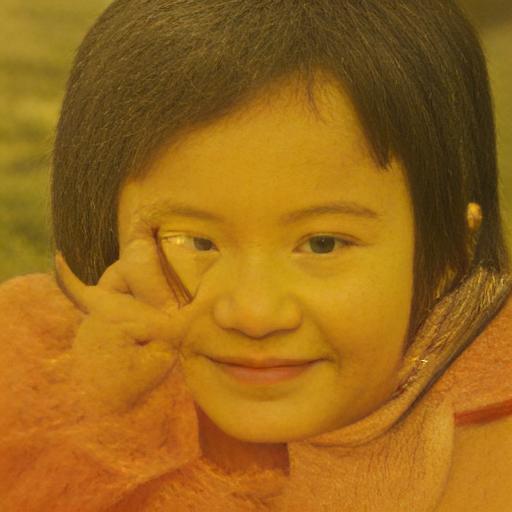} \hspace{-4mm} &
    \includegraphics[width=0.098\textwidth]{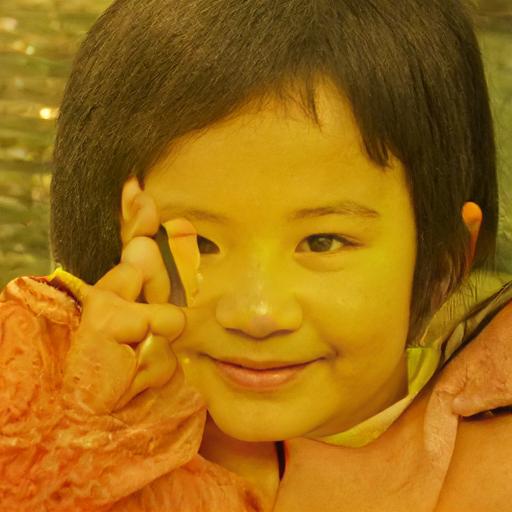} \hspace{-4mm} &
    \includegraphics[width=0.098\textwidth]{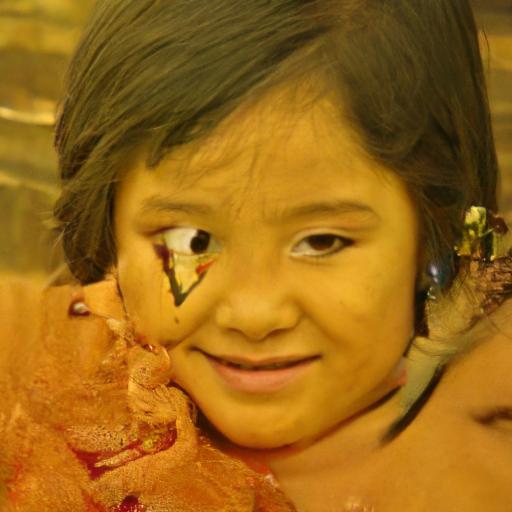} \hspace{-4mm} &
    \includegraphics[width=0.098\textwidth]{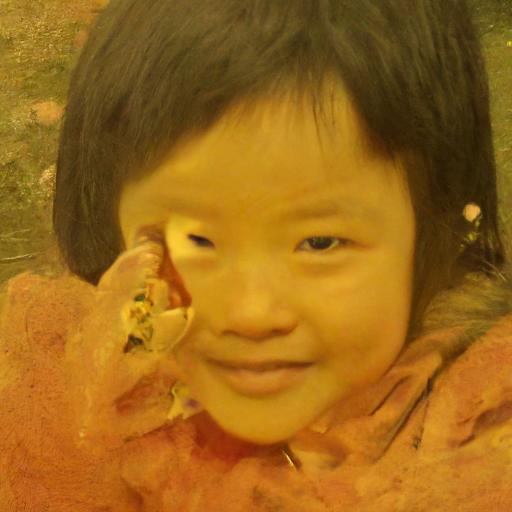} \hspace{-4mm} &
    \includegraphics[width=0.098\textwidth]{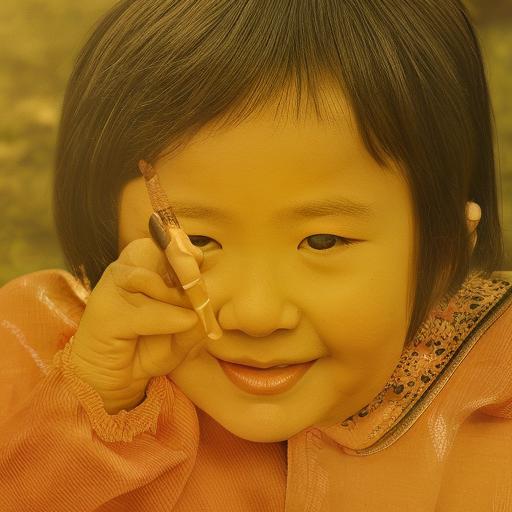} \hspace{-4mm} &
    \includegraphics[width=0.098\textwidth]{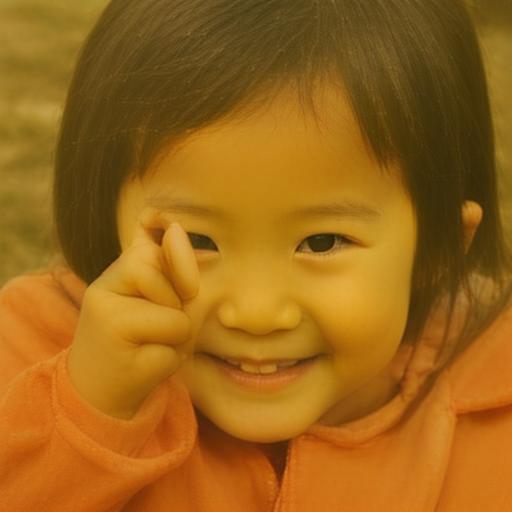} \hspace{-4mm} &
    \includegraphics[width=0.098\textwidth]{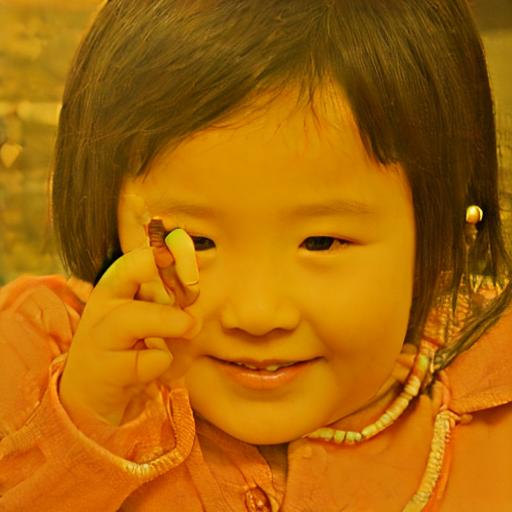} \hspace{-4mm} 
    \\
    WebPhoto 00043 \hspace{-4mm} &
    RF++~\cite{wang2023restoreformer++} \hspace{-4mm} &
    VQFR~\cite{gu2022vqfr} \hspace{-4mm} &
    CodeFormer~\cite{zhou2022codeformer} \hspace{-4mm} &
    DAEFR~\cite{tsai2024daefr} \hspace{-4mm} & 
    PGDiff~\cite{yang2023pgdiff} \hspace{-4mm} &
    DifFace~\cite{yue2024difface} \hspace{-4mm} &
    DiffBIR~\cite{lin2024diffbir} \hspace{-4mm} &
    OSEDiff*~\cite{wu2024osediff} \hspace{-4mm} &
    OSDFace (ours) \hspace{-4mm}
    \\
    \end{tabular}
    \end{adjustbox}
}
\scalebox{0.98}{
    \hspace{-0.4cm}
    \begin{adjustbox}{valign=t}
    \begin{tabular}{cccccccccc}
    \includegraphics[width=0.098\textwidth]{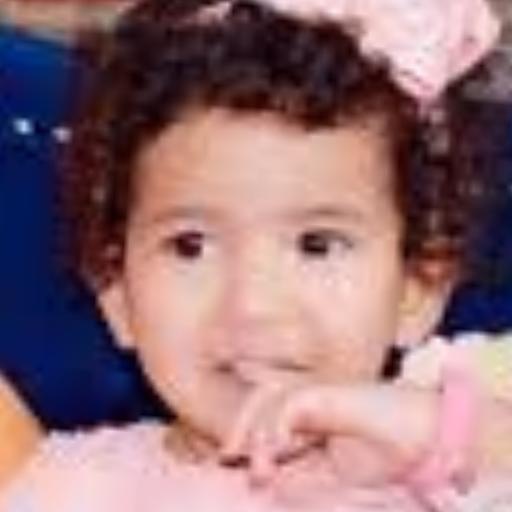} \hspace{-4mm} &
    \includegraphics[width=0.098\textwidth]{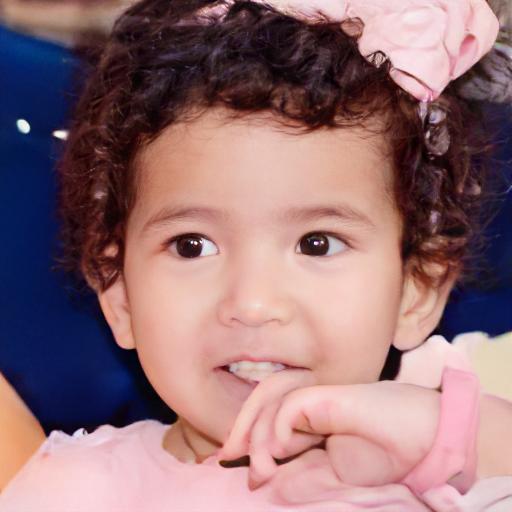} \hspace{-4mm} &
    \includegraphics[width=0.098\textwidth]{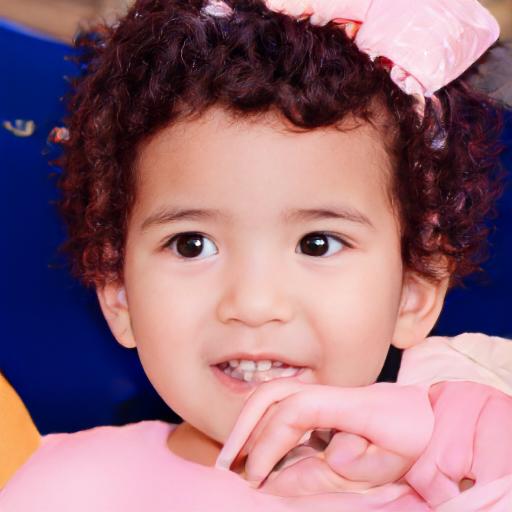} \hspace{-4mm} &
    \includegraphics[width=0.098\textwidth]{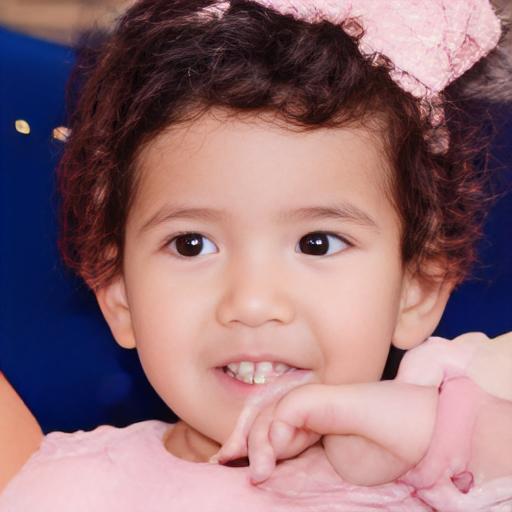} \hspace{-4mm} &
    \includegraphics[width=0.098\textwidth]{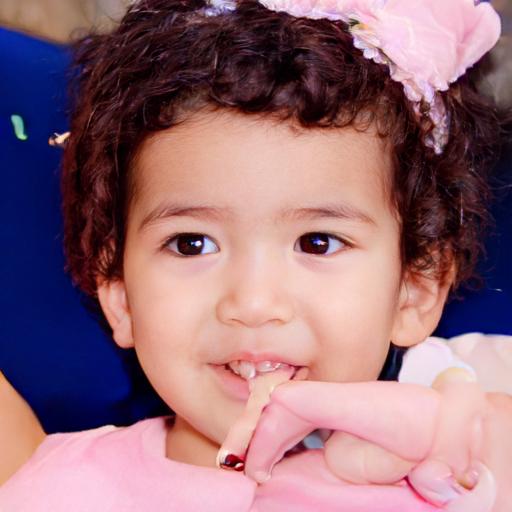} \hspace{-4mm} &
    \includegraphics[width=0.098\textwidth]{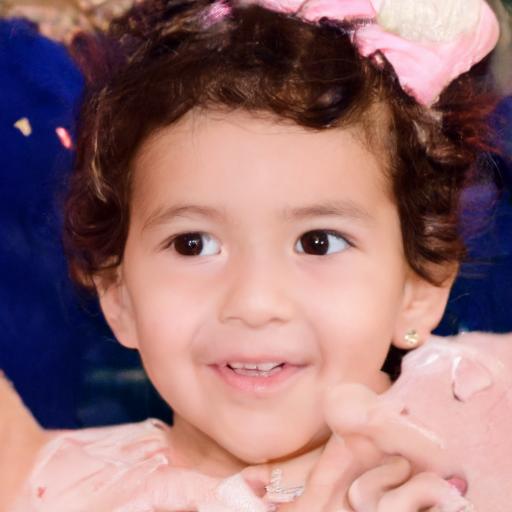} \hspace{-4mm} &
    \includegraphics[width=0.098\textwidth]{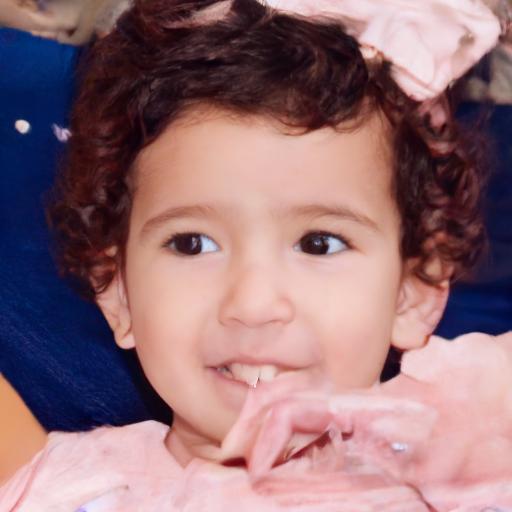} \hspace{-4mm} &
    \includegraphics[width=0.098\textwidth]{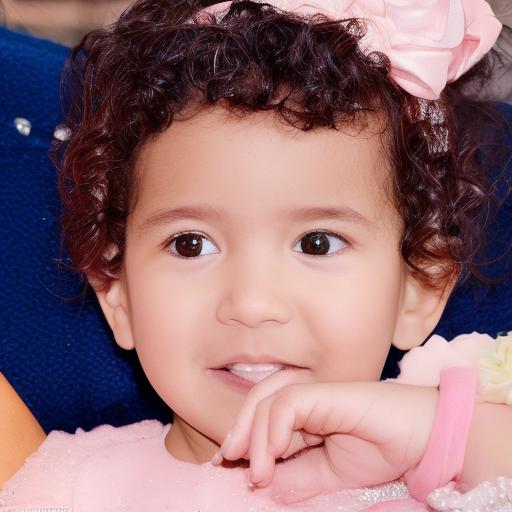} \hspace{-4mm} &
    \includegraphics[width=0.098\textwidth]{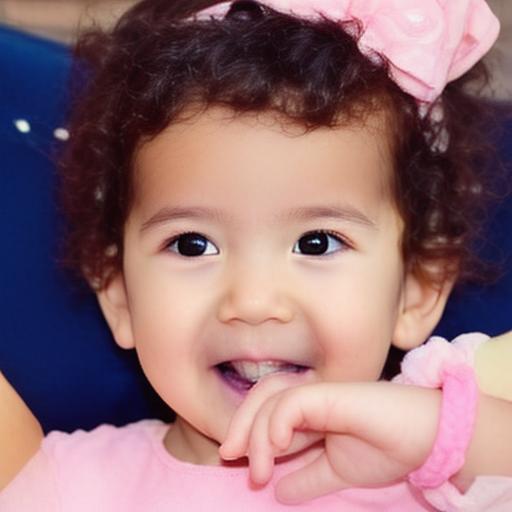} \hspace{-4mm} &
    \includegraphics[width=0.098\textwidth]{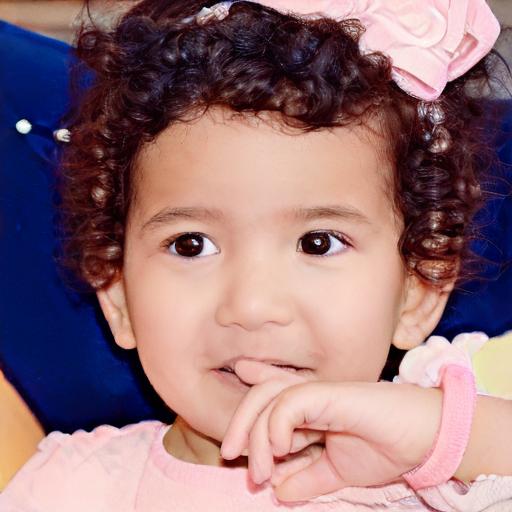} \hspace{-4mm} 
    \\
    WebPhoto 10017 \hspace{-4mm} &
    RF++~\cite{wang2023restoreformer++} \hspace{-4mm} &
    VQFR~\cite{gu2022vqfr} \hspace{-4mm} &
    CodeFormer~\cite{zhou2022codeformer} \hspace{-4mm} &
    DAEFR~\cite{tsai2024daefr} \hspace{-4mm} & 
    PGDiff~\cite{yang2023pgdiff} \hspace{-4mm} &
    DifFace~\cite{yue2024difface} \hspace{-4mm} &
    DiffBIR~\cite{lin2024diffbir} \hspace{-4mm} &
    OSEDiff*~\cite{wu2024osediff} \hspace{-4mm} &
    OSDFace (ours) \hspace{-4mm}
    \\
    \end{tabular}
    \end{adjustbox}
}
\scalebox{0.98}{
    \hspace{-0.4cm}
    \begin{adjustbox}{valign=t}
    \begin{tabular}{cccccccccc}
    \includegraphics[width=0.098\textwidth]{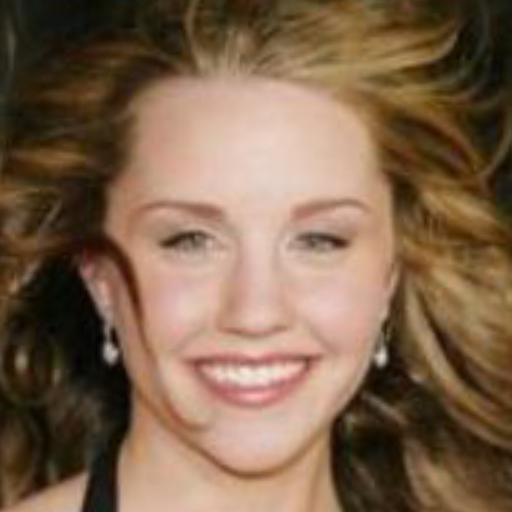} \hspace{-4mm} &
    \includegraphics[width=0.098\textwidth]{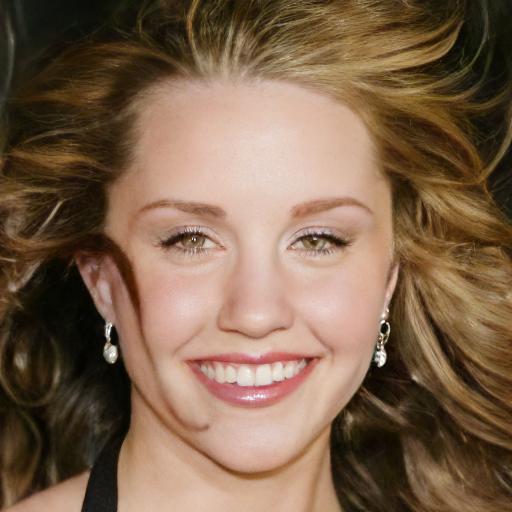} \hspace{-4mm} &
    \includegraphics[width=0.098\textwidth]{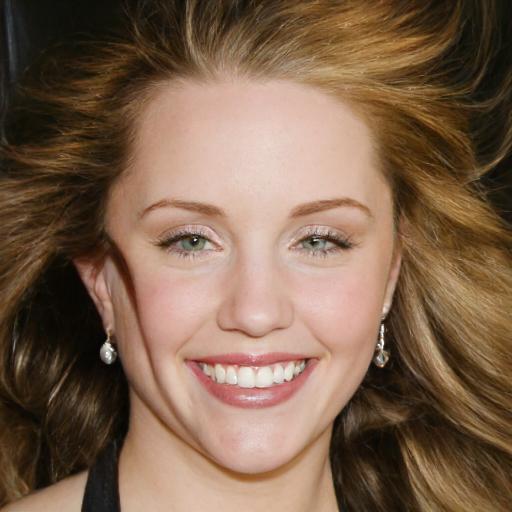} \hspace{-4mm} &
    \includegraphics[width=0.098\textwidth]{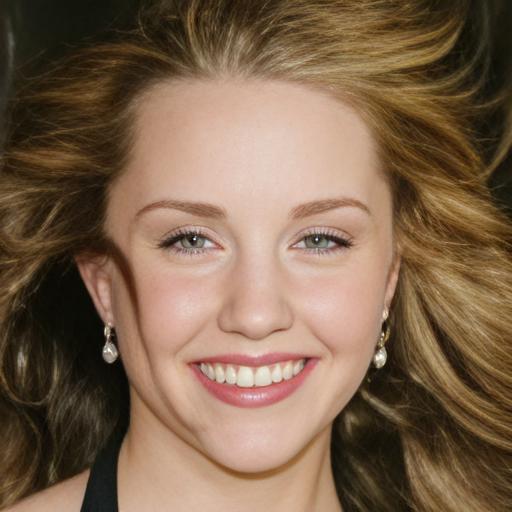} \hspace{-4mm} &
    \includegraphics[width=0.098\textwidth]{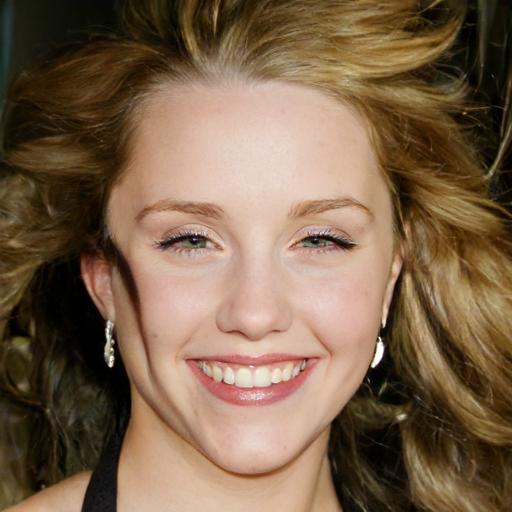} \hspace{-4mm} &
    \includegraphics[width=0.098\textwidth]{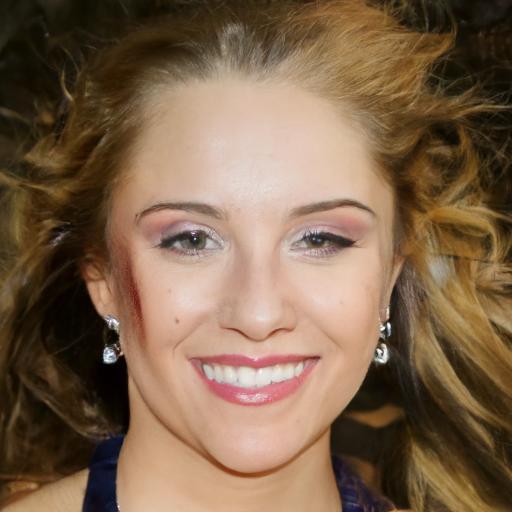} \hspace{-4mm} &
    \includegraphics[width=0.098\textwidth]{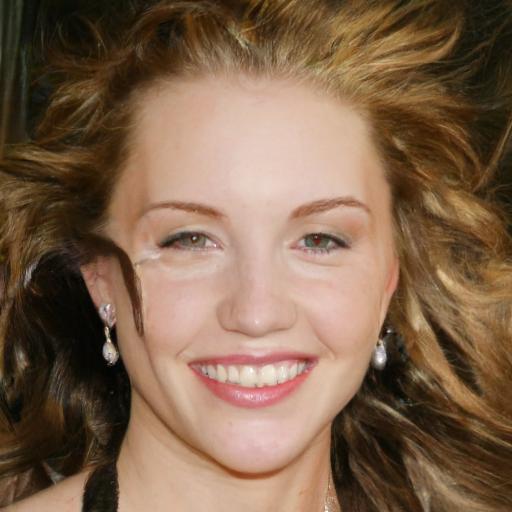} \hspace{-4mm} &
    \includegraphics[width=0.098\textwidth]{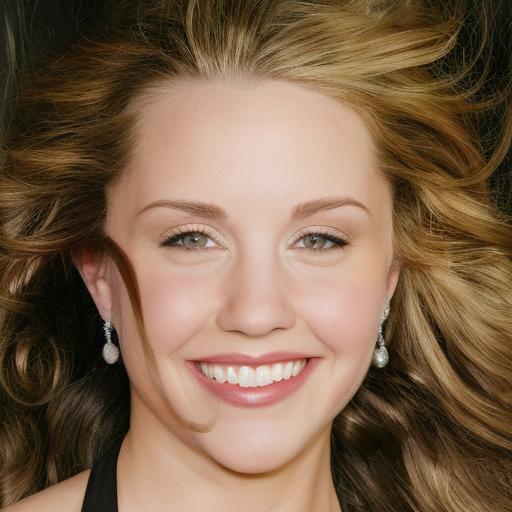} \hspace{-4mm} &
    \includegraphics[width=0.098\textwidth]{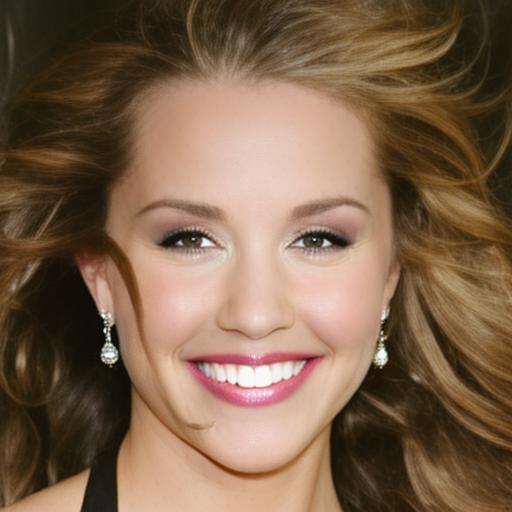} \hspace{-4mm} &
    \includegraphics[width=0.098\textwidth]{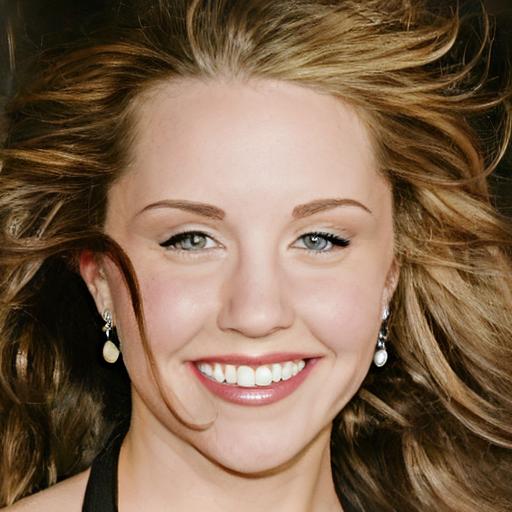} \hspace{-4mm} 
    \\
    LFW A. Bynes \hspace{-4mm} &
    RF++~\cite{wang2023restoreformer++} \hspace{-4mm} &
    VQFR~\cite{gu2022vqfr} \hspace{-4mm} &
    CodeFormer~\cite{zhou2022codeformer} \hspace{-4mm} &
    DAEFR~\cite{tsai2024daefr} \hspace{-4mm} & 
    PGDiff~\cite{yang2023pgdiff} \hspace{-4mm} &
    DifFace~\cite{yue2024difface} \hspace{-4mm} &
    DiffBIR~\cite{lin2024diffbir} \hspace{-4mm} &
    OSEDiff*~\cite{wu2024osediff} \hspace{-4mm} &
    OSDFace (ours) \hspace{-4mm}
    \\
    \end{tabular}
    \end{adjustbox}
}
\scalebox{0.98}{
    \hspace{-0.4cm}
    \begin{adjustbox}{valign=t}
    \begin{tabular}{cccccccccc}
    \includegraphics[width=0.098\textwidth]{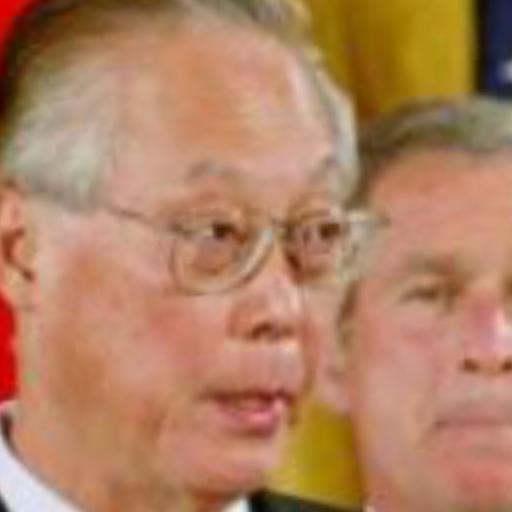} \hspace{-4mm} &
    \includegraphics[width=0.098\textwidth]{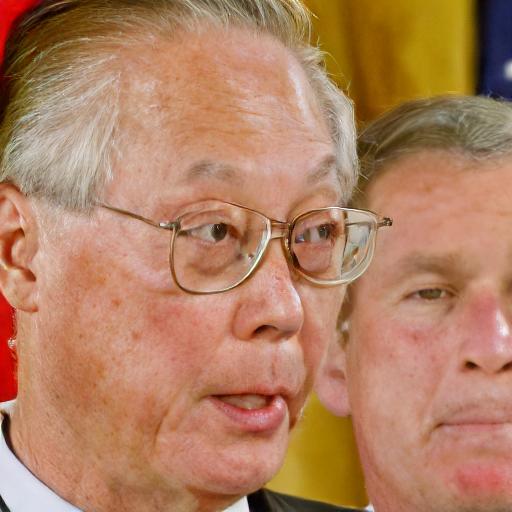} \hspace{-4mm} &
    \includegraphics[width=0.098\textwidth]{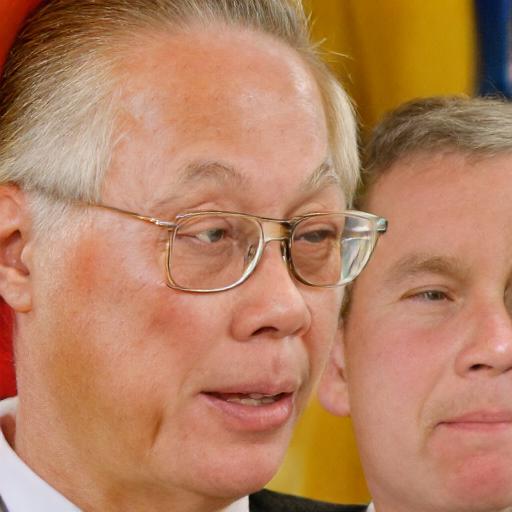} \hspace{-4mm} &
    \includegraphics[width=0.098\textwidth]{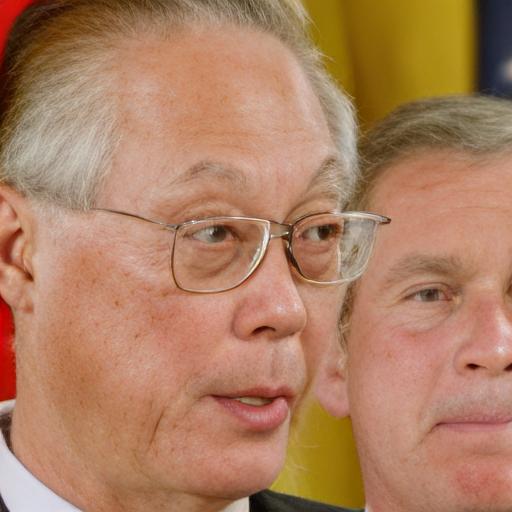} \hspace{-4mm} &
    \includegraphics[width=0.098\textwidth]{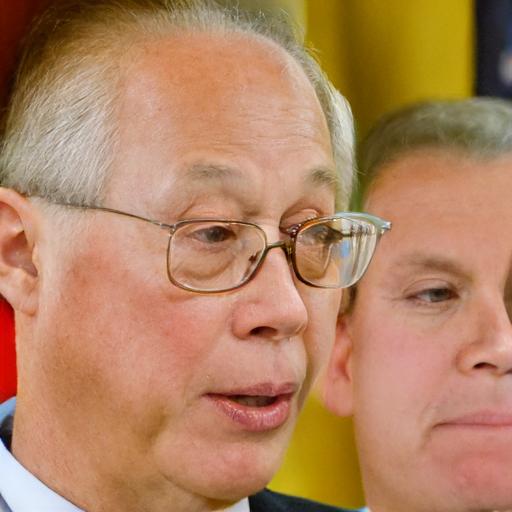} \hspace{-4mm} &
    \includegraphics[width=0.098\textwidth]{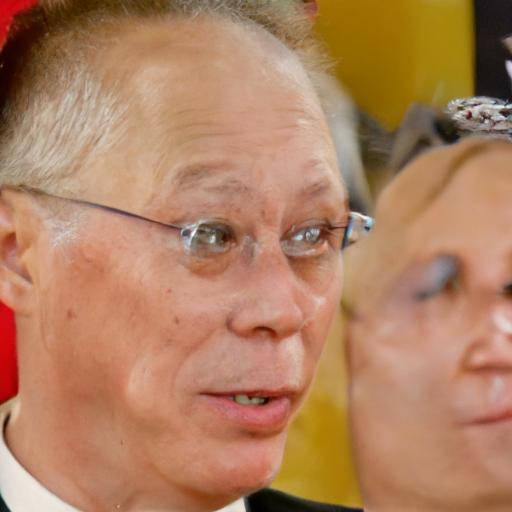} \hspace{-4mm} &
    \includegraphics[width=0.098\textwidth]{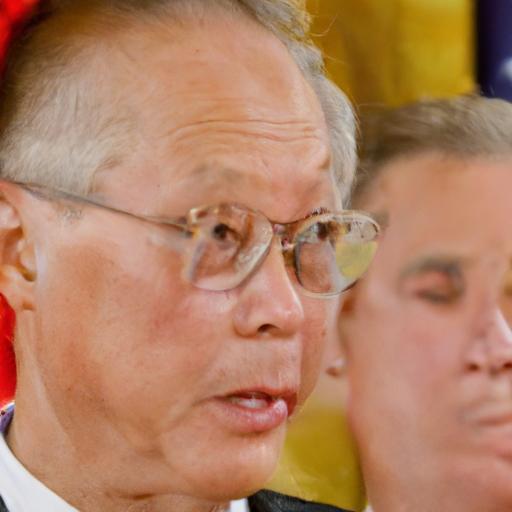} \hspace{-4mm} &
    \includegraphics[width=0.098\textwidth]{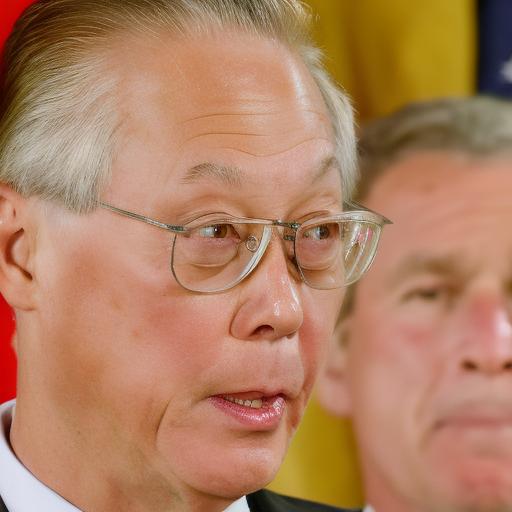} \hspace{-4mm} &
    \includegraphics[width=0.098\textwidth]{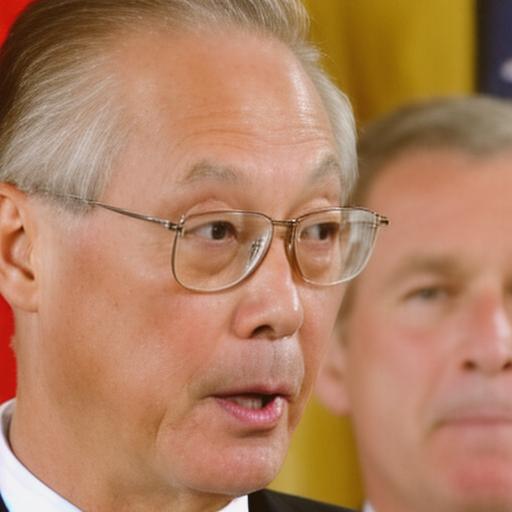} \hspace{-4mm} &
    \includegraphics[width=0.098\textwidth]{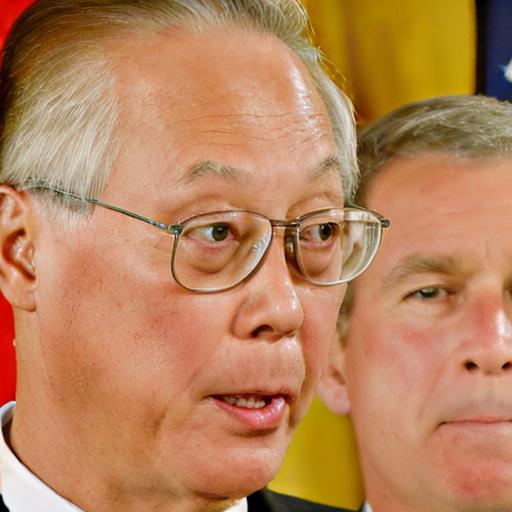} \hspace{-4mm} 
    \\
    LFW Chok T.G. \hspace{-4mm} &
    RF++~\cite{wang2023restoreformer++} \hspace{-4mm} &
    VQFR~\cite{gu2022vqfr} \hspace{-4mm} &
    CodeFormer~\cite{zhou2022codeformer} \hspace{-4mm} &
    DAEFR~\cite{tsai2024daefr} \hspace{-4mm} & 
    PGDiff~\cite{yang2023pgdiff} \hspace{-4mm} &
    DifFace~\cite{yue2024difface} \hspace{-4mm} &
    DiffBIR~\cite{lin2024diffbir} \hspace{-4mm} &
    OSEDiff*~\cite{wu2024osediff} \hspace{-4mm} &
    OSDFace (ours) \hspace{-4mm}
    \\
    \end{tabular}
    \end{adjustbox}
}
\vspace{-3.3mm}
\caption{Visual comparison of the real-world datasets in challenging cases. Please zoom in for a better view.}
\label{fig:vis-realworld}
\vspace{-3.5mm}
\end{figure*}

\begin{table*}[t]
\scriptsize
\setlength{\tabcolsep}{0.4mm}

\newcolumntype{?}{!{\vrule width 1pt}}
\newcolumntype{C}{>{\centering\arraybackslash}X}

\begin{center}
\begin{tabularx}{\textwidth}{l|*{5}{C}|*{5}{C}|*{5}{C}}
\toprule[0.15em]
\rowcolor{color3} & \multicolumn{5}{c|}{Wider-Test} & \multicolumn{5}{c|}{LFW-Test} & \multicolumn{5}{c}{WebPhoto-Test} \\
\rowcolor{color3}
\multirow{-2}{*}{Methods} & C-IQA$\uparrow$ & M-IQA$\uparrow$ & MUSIQ$\uparrow$ & NIQE$\downarrow$ & FID$\downarrow$ & C-IQA$\uparrow$ & M-IQA$\uparrow$ & MUSIQ$\uparrow$ & NIQE$\downarrow$ & FID$\downarrow$ & C-IQA$\uparrow$ & M-IQA$\uparrow$ & MUSIQ$\uparrow$ & NIQE$\downarrow$ & FID$\downarrow$ \\
\midrule[0.15em]

RestoreFormer{\tiny{++}}~\cite{wang2023restoreformer++} & 
0.7159 & 0.4767 & 71.332 & \textcolor{cvprblue}{3.7231} & 45.398 &
0.7024 & 0.5108 & 72.250 & 3.8434 & 50.253 &
0.6950 & 0.4902 & 71.484 & 4.0202 & \textcolor{red}{75.071} \\

VQFR~\cite{gu2022vqfr} & 
0.7069 & 0.5044 & 71.417 & 4.0357 & 37.866 &
0.7098 & 0.5339 & 74.385 & \textcolor{cvprblue}{3.8356} & 49.800 &
0.6769 & 0.4909 & 70.906 & 4.6095 & 84.776 \\

CodeFormer~\cite{zhou2022codeformer} & 
0.6986 & 0.4958 & 73.406 & 4.1188 & 38.765 &
0.6890 & 0.5266 & 75.484 & 4.4377 & 52.341 &
0.6918 & 0.5034 & \textcolor{red}{74.001} & 4.6273 & 83.197 \\

DAEFR~\cite{tsai2024daefr} & 
0.6975 & 0.5205 & 74.143 & \textcolor{red}{3.5701} & 36.701 &
0.6964 & 0.5420 & \textcolor{cvprblue}{75.838} & \textcolor{red}{3.4788} & 47.527 &
0.6696 & 0.4934 & 72.698 & \textcolor{red}{3.9333} & \textcolor{cvprblue}{75.474} \\
\midrule
PGDiff~\cite{yang2023pgdiff} (s=1,000) & 
0.5824 & 0.4531 & 68.135 & 3.9315 & 35.862 &
0.5975 & 0.4858 & 71.244 & 4.0112 & \textcolor{cvprblue}{41.209} &
0.5653 & 0.4460 & 68.599 & 3.9930 & 86.954 \\

DifFace~\cite{yue2024difface} (s=250) & 
0.5924 & 0.4299 & 64.907 & 4.2380 & 37.099 &
0.6075 & 0.4577 & 69.617 & 3.9016 & 46.127 &
0.5737 & 0.4189 & 65.116 & 4.2474 & 79.556 \\
\end{tabularx}
\end{center}
\vspace{-6.3mm}
\caption{Quantitative comparison on real-world datasets. C-IQA stands for CLIPIQA, and M-IQA stands for MANIQA. The best and second best results are colored with \textcolor{red}{red} and \textcolor{cvprblue}{blue}. OSEDiff* is retrained on FFHQ dataset for reference.}
\label{table:RealWorld}
\vspace{-6mm}
\end{table*}

\vspace{-1mm}
\section{Experiments}
\label{sec:exp}
\vspace{-1mm}
\subsection{Experimental Settings}
\vspace{-1mm}
\noindent\textbf{Training Datasets.} Our model is trained on FFHQ~\cite{karras2019ffhq} and its retouched version~\cite{Shafaei2021ffhqr}, containing 70,000 different high-quality face images. Images are resized to 512$\times$512 pixels. Synthetic training data is generated using a dual-stage degradation model, with parameters following WaveFace~\cite{miao2024waveface}. This dual-stage degradation process closely aligns the synthetic data with real-world degradation scenarios, handling both mild and severe degradations. 

\noindent\textbf{Testing Datasets.} We evaluate our method on the synthetic dataset CelebA-Test~\cite{karras2018celeba} from DAEFR~\cite{tsai2024daefr} and three widely used real-world datasets: Wider-Test~\cite{zhou2022codeformer}, LFW-Test~\cite{huang2008lfw}, and WebPhoto-Test~\cite{wang2021gfpgan}, following the settings in previous literature~\cite{zhou2022codeformer,wang2023restoreformer++,gu2022vqfr}. These datasets exhibit diverse and complex degradations. For example, the degradations in LFW-Test are relatively slight and regular, whereas those in Wider-Test are stronger. 

\noindent\textbf{Metrics.} For the \emph{Synthetic Dataset}, we employ LPIPS~\cite{zhang2018lpips} and DISTS~\cite{ding2020dists} as reference-based perceptual quality measures, together with MUSIQ~\cite{ke2021musiq} and NIQE~\cite{zhang2015niqe} as no-reference image quality measures.
To evaluate the distribution similarity between the ground truth and restored faces, we calculate FID~\cite{heusel2017fid} with both FFHQ and CelebA-Test HQ, showing the ability to recover “real” faces and HQ images.
Following previous works~\cite{wang2021gfpgan,gu2022vqfr,tsai2024daefr}, we use the embedding angle of ArcFace~\cite{deng2019arcface}, namely ``Deg.'', and landmark distance, namely ``LMD'', as additional identity and fidelity metrics. For the \emph{Real-world Datasets}, we employ non-reference metrics, including FID (with FFHQ as reference), as well as CLIPIQA~\cite{wang2022clipiqa}, MANIQA~\cite{yang2022maniqa}, MUSIQ~\cite{ke2021musiq}, and NIQE~\cite{zhang2015niqe}.
We utilize the evaluation codes provided by pyiqa~\cite{pyiqa} and VQFR~\cite{gu2022vqfr} for all metrics. 

\noindent\textbf{Implementation Details.} In the first stage, we train the VRE using the Adam optimizer~\cite{kingma2014adam} with a learning rate of $1.44$$\times$$10^{-4}$ and a batch size of $32$. In our implementation, the input face image has dimensions of 512$\times$512, and the quantized feature map has dimensions of 256$\times$512 after flattening. The VQ dictionaries contain $N$$=$$1,024$ code items, each with a channel size of 512. The HQ and LQ dictionaries are trained for 50 and 10 epochs, respectively. The association parameter $\lambda_{\text{assoc}}$ is initially set to 0, followed by an additional 10 epochs with $\lambda_{\text{assoc}}$$=$$1$. The VRE is trained on 8 NVIDIA A800 GPUs. 

In the second stage, our OSDFace is trained by the AdamW optimizer~\cite{loshchilov2018AdamW}, with learning rate $1$$\times10$$^{-4}$ and batch size $2$ for both the generator and discriminator. The LoRA rank in the UNet is set to $16$. The SD 2.1-base~\cite{sd21} serves as the pretrained OSD model, with the VRE used to construct the prompt embeddings. The setting of the discriminator is followed by D$^3$SR~\cite{li2024dfosd}, and the generator and discriminator are trained alternately. Training is performed for 150K iterations on 2 NVIDIA A6000 GPUs.

\noindent\textbf{Compared State-of-the-Art (SOTA) Methods.} We compare OSDFace with several SOTA methods, including RestoreFormer++~\cite{wang2023restoreformer++}, VQFR~\cite{gu2022vqfr}, CodeFormer~\cite{zhou2022codeformer},  DAEFR~\cite{tsai2024daefr}, PGDiff~\cite{yang2023pgdiff},  DifFace~\cite{yue2024difface} and DiffBIR~\cite{lin2024diffbir}. We also provide the current SOTA OSD model OSEDiff~\cite{wu2024osediff}, and retrain it on FFHQ~\cite{karras2019ffhq}, namely OSEDiff*.

\vspace{-1.5mm}
\subsection{Performance on Synthetic Data}
\vspace{-1.5mm}
\noindent\textbf{Quantitative Results.} The results in Tab.~\ref{table:CelebA-Test} demonstrate that OSDFace outperforms other methods in Deg., indicating better alignment with the ID features of HQ images. It also achieves higher scores on LPIPS and DISTS, highlighting superior perceptual coherence and detail preservation. Performance in FID(HQ) shows that OSDFace effectively recovers the distribution of the original HQ dataset.

\noindent\textbf{Qualitative Results.} Visual comparisons in Fig.~\ref{fig:vis-celeba} illustrate that our method closely resembles real images. OSDFace enhances realism in elements such as earrings, eyelashes, eyebrows, hair strands, and subtle skin textures. Our approach restores hair most naturally, avoiding additional streaks or other regular textures, and maintains consistent hair color without random changes.
\vspace{-1mm}
\subsection{Performance on Real-world Data}
\vspace{-1mm}
\noindent\textbf{Quantitative Results.} In Tab.~\ref{table:RealWorld}, we can see that OSDFace performs significantly better than another OSD method across all metrics. OSDFace outperforms non-diffusion models in most metrics. For multi-step diffusion models, we generally surpass PGDiff and DifFace. For DiffBIR, we maintain a small gap and even exceed it in certain metrics.

\noindent\textbf{Qualitative Results.} Figure~\ref{fig:vis-realworld} visualizes representative images in three real-world datasets. Existing face restoration methods often emphasize facial features while overlooking the precise reconstruction of facial accessories like glasses, hair, hats, headwear, and background details. Some methods, especially DiffBIR~\cite{lin2024diffbir}, are over-smoothing and lack fine facial textures. OSDFace naturally and clearly restores multiple faces within a single image, showing its effectiveness for unaligned faces. OSDFace ensures high-fidelity restoration with balanced and consistent visual quality throughout the image. We also test its zero-shot ability on cartoon images in Fig.~\ref{fig:cartoon}, using Anime Face Dataset~\cite{chao2019/online}. 

\begin{table}[t] 
\scriptsize
\centering
\resizebox{\columnwidth}{!}{
\setlength{\tabcolsep}{0.5mm}
\begin{tabular}{l|ccccc}
\toprule
\rowcolor{color3} Methods & PGDiff~\cite{yang2023pgdiff} & DifFace~\cite{yue2024difface} & DiffBIR~\cite{lin2024diffbir} & OSEDiff~\cite{wu2024osediff} & OSDFace\\ \midrule[0.15em]
 Step    & 1,000   &  250    & 50      & 1        & \hspace{0.5mm} 1     \hspace{0.5mm} \\
 Time (s)   & 85.81   &  7.05   & 9.03    & 0.13     & \hspace{0.5mm} 0.10  \hspace{0.5mm} \\ 
 Param (M)  & 176.4   &  175.4  & 3,042   & 1,302    & \hspace{0.5mm} 978.4 \hspace{0.5mm} \\ 
 MACs (G)   & 480,997 &  18,682 & 189,208 & 2,269    & \hspace{0.5mm} 2,132 \hspace{0.5mm} \\ 
\bottomrule
\end{tabular}
}
\vspace{-3.2mm}
\caption{Complexity comparison during inference. All models are tested with an input image size of 512$\times$512.}
\label{tab:speed}
\vspace{-3.5mm}
\end{table}

\begin{table}[t]
    \centering
    \scriptsize

\resizebox{\columnwidth}{!}{
\setlength{\tabcolsep}{1.6mm}

    \begin{tabular}{l|ccccc} 
        \toprule[0.15em]
        \rowcolor{color3} Methods & C-IQA$\uparrow$ & M-IQA$\uparrow$ & MUSIQ$\uparrow$ & NIQE$\downarrow$ & FID$\downarrow$ \\
        \midrule[0.15em]
        Learnable & 
            0.6714 & 0.5312 &                           75.0339 &           4.0855 &            45.4041 \\
        DAPE~\cite{wu2024seesr}& 
        \textcolor{cvprblue}{0.6863} & \textcolor{red}{0.5622} & 74.8107 & 4.3801 & 45.9380 \\
        VRE (w/o~assoc)& 
        0.6791                       &                   0.5180 & \textcolor{cvprblue}{75.2328} & \textcolor{red}{3.8228} &\textcolor{cvprblue}{44.3360} \\
        VRE (w/~assoc)& 
        \textcolor{red}{0.6946} & \textcolor{cvprblue}{0.5356} & \textcolor{red}{75.2911} & \textcolor{cvprblue}{3.8793} & \textcolor{red}{41.9502} \\
        \bottomrule[0.15em]
    \end{tabular}
}
    \vspace{-3mm} 
    \caption{Ablation study of prompt embedding generation. The best and second best results are colored with \textcolor{red}{red} and \textcolor{cvprblue}{blue}, respectively.}
    \label{table:ablation_prompt}
    \vspace{-7mm}
\end{table}

\vspace{-1mm}
\subsection{Complexity Analysis.} 
\vspace{-1mm}
Table~\ref{tab:speed} provides a detailed comparison of model complexity, including number of steps, inference time, parameter number, and multiply-accumulate operations (MACs) during inference. All evaluations are carried out on an NVIDIA A6000 GPU for consistency. \textbf{First}, by relying on one-step diffusion process, OSDFace significantly outperforms multi-step diffusion models in both inference time and MACs, demonstrating remarkable efficiency. \textbf{Furthermore}, OSDFace maintains an advantage over other OSD model. We do not incorporate text encoders (\eg, the DAPE used by OSEDiff~\cite{wu2024osediff}), nor do we introduce additional modules that increase Stable Diffusion model complexity (\eg, the ControlNet used by DiffBIR~\cite{lin2024diffbir}). Instead, we utilize a more efficient image encoder, VRE. \textbf{Consequently}, OSDFace achieves the lowest MACs count and fastest inference speed among all evaluated diffusion-based methods. Compared to the other OSD model, OSEDiff~\cite{wu2024osediff}, OSDFace demonstrates a 25\% reduction in parameters and a 23\% improvement in inference speed. 

\begin{table}[t]
    \centering
    \scriptsize
    \setlength{\tabcolsep}{0.5mm} 
    \newcolumntype{C}{>{\centering\arraybackslash}X}

    \newcolumntype{S}{>{\centering\arraybackslash}c}

    \begin{tabularx}{\columnwidth}{SSSS|*{5}{C}} 
        \toprule[0.15em]
        \rowcolor{color3} $\mathcal{L}_\text{ID}$ & $\mathcal{L}_{\text{EA-DISTS}}$ & $\mathcal{L}_\mathcal{G}$ & $\mathcal{L}_{\text{EA-LPIPS}}$ & C-IQA$\uparrow$ & M-IQA$\uparrow$ & MUSIQ$\uparrow$ & NIQE$\downarrow$ & FID$\downarrow$ \\
        \midrule[0.15em]
        & \checkmark & \checkmark & & \textcolor{cvprblue}{0.6724} & 0.5243 & 74.5986 & 4.0190 & \textcolor{red}{41.7842} \\
        \checkmark & & \checkmark & & 0.6710 & \textcolor{cvprblue}{0.5387} & 74.1060 & 4.3223 & 55.1016 \\
        \checkmark &  & \checkmark & \checkmark & 0.6674 & \textcolor{red}{0.5470} & 75.2021 & 4.1484 & 46.1684 \\
        \checkmark & \checkmark & & & 0.6590 & 0.5081 & \textcolor{cvprblue}{75.2336} & \textcolor{cvprblue}{3.9857} & 45.7834 \\
        \checkmark & \checkmark & \checkmark & & \textcolor{red}{0.6946} & 0.5356 & \textcolor{red}{75.2911} & \textcolor{red}{3.8793} & \textcolor{cvprblue}{41.9502} \\
        \bottomrule[0.15em]
    \end{tabularx}

    \vspace{-3mm}
    \caption{Ablation studies on different components. The best and second best results are colored with \textcolor{red}{red} and \textcolor{cvprblue}{blue}, respectively.}
    \label{table:ablation_loss}
    \vspace{-6mm}
\end{table}

\vspace{-1mm}
\subsection{Ablation Studies}
\vspace{-1mm}

\noindent\textbf{Visual Representation Embedder (VRE).} We evaluated the impact of various prompt embedding generation methods on the training of OSDFace, including VRE, learnable prompt embeddings, and Degradation-Aware Prompt Extractor (DAPE)~\cite{wu2024seesr}. As shown in Tab.~\ref{table:ablation_prompt}, the experimental results indicate that our method using VRE outperforms others across most metrics. As shown in Fig.~\ref{fig:embedding_ablation}, using VRE significantly enhances the facial details and the coherence of non-facial areas in the generated faces.

Furthermore, we trained the VRE using two approaches: with and without the HQ-LQ feature association loss, further validating the effectiveness of our method. As shown in the last two rows of Tab.~\ref{table:ablation_prompt}, training with association loss achieves superior performance. This demonstrates that our VRE can extract valuable information from LQ images to guide face restoration effectively. \textbf{Notably}, with the inclusion of the feature association loss, the VRE gains the ability to identify and reconstruct certain HQ-like categories from LQ inputs, enhancing face restoration quality.

\noindent\textbf{Realistic Face Alignment Loss.} For the series of losses used in training the OSDFace, such as facial identity loss, EA-DISTS, and adversarial loss, we conducted extensive ablation studies. \textbf{Additionally}, we introduce the EA-LPIPS loss, defined similarly to EA-DISTS in Sec.~\ref{sec:loss}. This loss first processes the image with the Sobel operator, followed by calculating the LPIPS loss. As shown in Tab.~\ref{table:ablation_loss}, the losses we utilized significantly contribute to model performance, with most metrics reaching the best scores.

\vspace{-1mm}
\section{Conclusion}
\vspace{-1mm}
We propose OSDFace, a high-speed diffusion model that enhances face image quality in one step. Our method bridges the gap between the powerful generative capabilities of text-to-image diffusion models and the specific requirements of face restoration. By introducing a visual prompt consistent with the model's original conditional input, we effectively utilize prior information from the low-quality faces themselves. This enables the model to generate more realistic and high-fidelity face images. Our approach not only addresses the limitations of previous methods but also sets a new direction for incorporating visual priors into diffusion models for face restoration. Extensive experiments demonstrate that OSDFace achieves superior performance compared to recent state-of-the-art methods.

\section*{Acknowledgments} 
We thank the reviewers for their valuable feedback. This work was supported by the Shanghai Municipal Science and Technology Major Project (2021SHZDZX0102) and the Fundamental Research Funds for the Central Universities. Additionally, this work was supported in part by NSFC grant 2024YFC3017100, YDZX20253100004004, 62302299, Shanghai Science and Technology Development Funds under Grant 23YF1420500, and Huawei Explore X funds. Special thanks to Jianze Li for his technical support.

\small
\bibliographystyle{ieeenat_fullname}
\bibliography{main}

\end{document}


\maketitle

\renewcommand{\thesection}{\Alph{section}}

\vspace{-0.5mm}
\section{Overall}
\vspace{-0.5mm}

This supplementary material provides additional results to support the main manuscript. First, in Section~\ref{sec:time}, we analyze the parameter size of VRE and inference time. Next, in Section~\ref{sec:osediff}, we present experiments that integrate VRE into OSEDiff~\cite{wu2024osediff}. These experiments demonstrate VRE's strong image understanding capabilities. In Section~\ref{sec:face_recog}, we validate our method on downstream face recognition tasks. The results show that our method outperforms others when used as a preprocessing step. Additionally, we analysis the limitation and future work in Section~\ref{sec:limit}. Finally, in Section~\ref{sec:vis-comp}, we provide more visual comparisons with state-of-the-art methods.

\vspace{-0.5mm}
\section{Parameters and Inference Time}
\label{sec:time}
\vspace{-0.5mm}

Table~\ref{tab:supp-speed} clearly shows that OSDFace achieves high inference speed and low computational cost compared to other one-step diffusion models. The VRE prompt embedder in OSDFace significantly reduces the parameter count and MACs. This reduction is notable when compared to the prompt embedder used in OSEDiff~\cite{wu2024osediff}, \ie, DAPE~\cite{wu2024seesr} with CLIP text encoder.

Additionally, generating text embeddings from input images does not conflict with generating latent vectors through a VAE encoder. Therefore, we can introduce a parallel mechanism that could speed up both OSEDiff~\cite{wu2024osediff} and OSDFace. Using parallel acceleration, our OSDFace could further reduce inference time by 14\% on top of its fast performance. All tests are conducted on an NVIDIA A6000 GPU.

\vspace{-0.5mm}
\section{Integrating VRE into OSEDiff}
\label{sec:osediff}
\vspace{-0.5mm}

The existing representative one-step diffusion (OSD) image restoration model, OSEDiff~\cite{wu2024osediff}, does not focus on face restoration tasks. In order to assess its applicability to face restoration, we retrained it using the same dataset and experimental settings as OSDFace, resulting in OSEDiff*. Furthermore, we integrated the proposed VRE into OSEDiff*, creating the enhanced model OSEDiff*+VRE.

\begin{figure}[t]
\scriptsize
\begin{center}

\hspace{-0.4cm}
\begin{adjustbox}{valign=t}
\begin{tabular}{cccc}
\includegraphics[width=0.23\columnwidth]{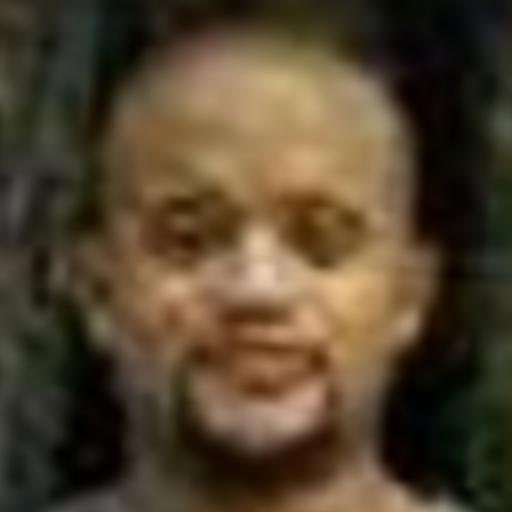} \hspace{-4mm} &
\includegraphics[width=0.23\columnwidth]{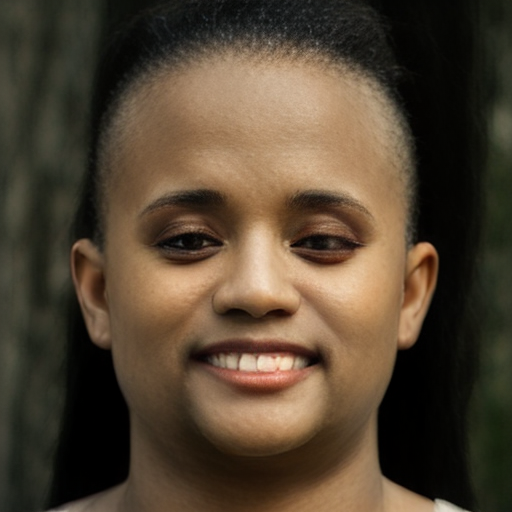} \hspace{-4mm} &
\includegraphics[width=0.23\columnwidth]{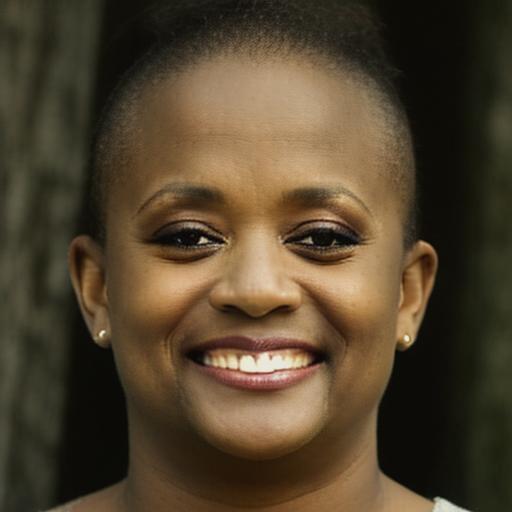} \hspace{-4mm} &
\includegraphics[width=0.23\columnwidth]{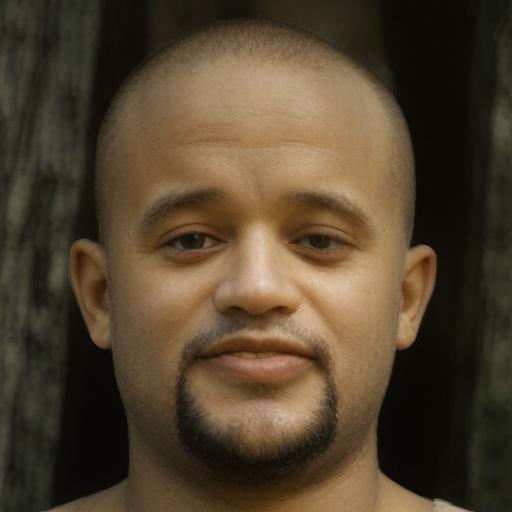} \hspace{-4mm} 
\\
Wider 0504 \hspace{-4mm} &
OSEDiff \hspace{-4mm} &
OSEDiff* \hspace{-4mm} &
OSEDiff*+VRE \hspace{-4mm} 
\\
\includegraphics[width=0.23\columnwidth]{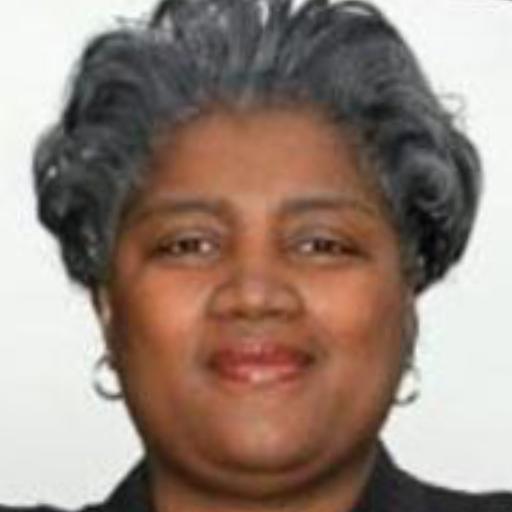} \hspace{-4mm} &
\includegraphics[width=0.23\columnwidth]{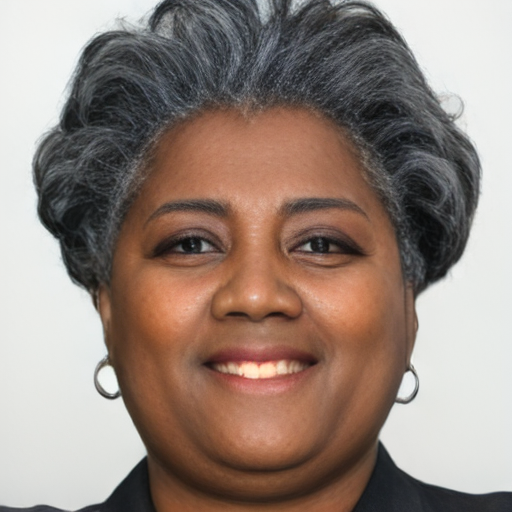} \hspace{-4mm} &
\includegraphics[width=0.23\columnwidth]{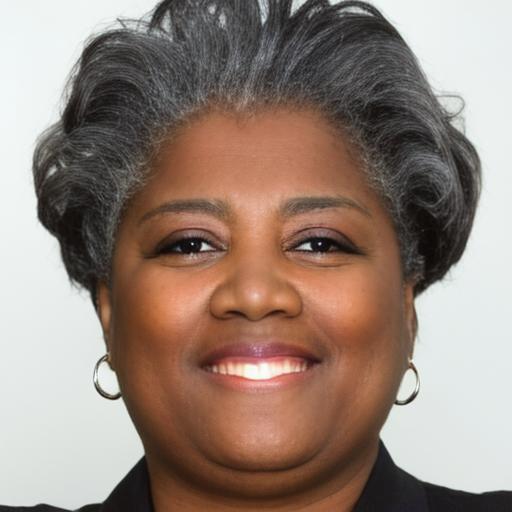} \hspace{-4mm} &
\includegraphics[width=0.23\columnwidth]{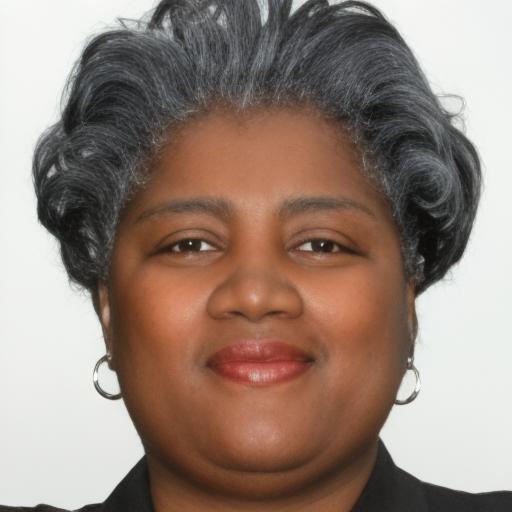} \hspace{-4mm} 
\\
LFW D. Brazile \hspace{-4mm} &
OSEDiff \hspace{-4mm} &
OSEDiff* \hspace{-4mm} &
OSEDiff*+VRE \hspace{-4mm} 
\\
\includegraphics[width=0.23\columnwidth]{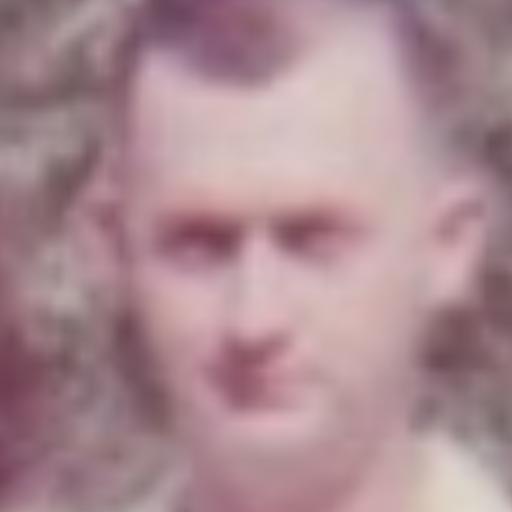} \hspace{-4mm} &
\includegraphics[width=0.23\columnwidth]{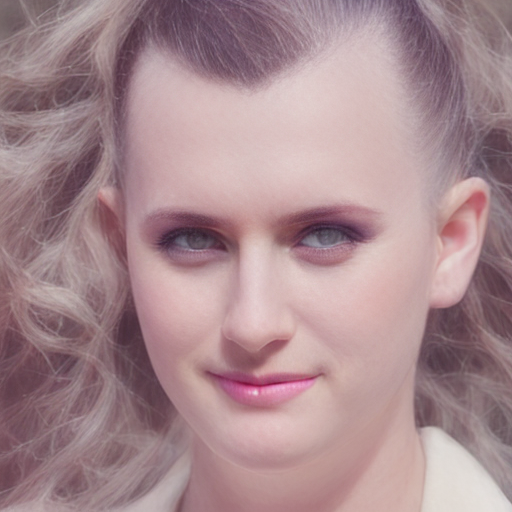} \hspace{-4mm} &
\includegraphics[width=0.23\columnwidth]{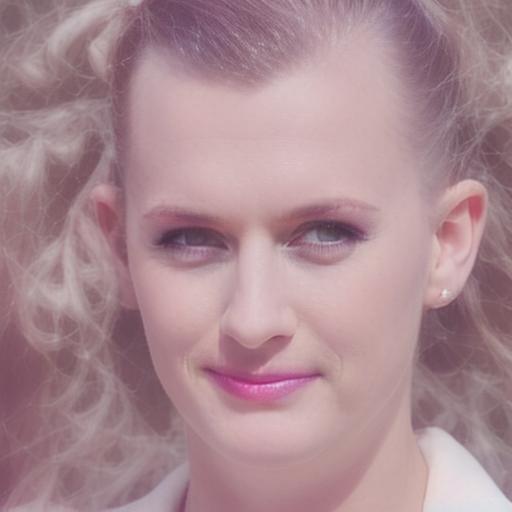} \hspace{-4mm} &
\includegraphics[width=0.23\columnwidth]{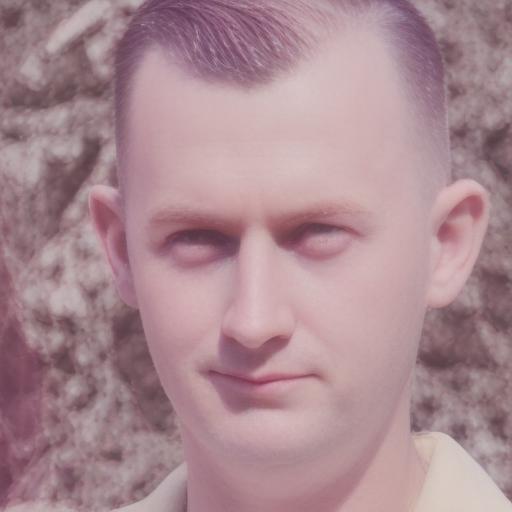} \hspace{-4mm} 
\\
WebPhoto 10038 \hspace{-4mm} &
OSEDiff \hspace{-4mm} &
OSEDiff* \hspace{-4mm} &
OSEDiff*+VRE \hspace{-4mm} 
\end{tabular}
\end{adjustbox}
\end{center}
\vspace{-6mm}
\caption{Visual comparisons of various versions of OSEDiff~\cite{wu2024osediff}. OSEDiff*+VRE shows enhanced visual quality.}
\label{fig:osediff}
\vspace{-4mm}
\end{figure}

\begin{table}[t]
\scriptsize
\setlength{\tabcolsep}{0.4mm} 
\newcolumntype{?}{!{\vrule width 1pt}}
\newcolumntype{C}{>{\centering\arraybackslash}X}
\centering
\begin{tabularx}{\columnwidth}{l|*{2}{C}|*{3}{C}}
\toprule[0.15em]
\rowcolor{color3} & \multicolumn{2}{c|}{ Prompt Embedder} & \multicolumn{2}{c}{ Inference Time (ms)} \\
\rowcolor{color3} 
\multirow{-2}{*}{ Methods} &  Param (M) &  MACs (G) &  Serialized &   Parallelized \\ \midrule[0.15em]
 \hspace{0.8mm}OSEDiff~\cite{wu2024osediff} & 353.41  &  141.45  &  130.75   &  125.54     \\
 \hspace{0.8mm}OSDFace (ours) \hspace{1mm}  & 28.63   &  99.47   &  119.00   &  102.83     \\ 
\bottomrule[0.15em]
\end{tabularx}

\vspace{-2mm}
\caption{Complexity comparison during inference. “parallelized” refers to the parallel execution of the prompt embedder and VAE encoder, while “serialized” denotes a fully sequential execution approach. We provide the number of parameters (Param), multiply-accumulate operations (MACs), and time during inference. All models are evaluated with  512$\times$512 input image.}
\label{tab:supp-speed}
\vspace{-6mm}
\end{table}

As shown in Tab.~\ref{table:OSEDiff-RealWorld}, Tab.~\ref{table:OSEDiff-CelebA-Test}, and Fig.~\ref{fig:osediff}, OSEDiff*+VRE performs well in both quantitative metrics and visual quality. The incorporation of VRE significantly reduces information loss during the image-text-embedding process, ensuring more accurate data representation. Visual results indicate that OSEDiff*+VRE prevents common issues like gender misclassification and unwanted artifacts. Additionally, it reliably captures subtle facial expressions from the input images. Besides, the IQA metrics demonstrate a competitive advantage by consistently reducing the distribution differences from the reference data. These experimental results demonstrate that our proposed VRE substantially enhances face restoration performance, particularly when applied to OSD models.

\begin{table*}[t]
\scriptsize
\setlength{\tabcolsep}{0.4mm} 
\vspace{-6mm}
\newcolumntype{?}{!{\vrule width 1pt}}
\newcolumntype{C}{>{\centering\arraybackslash}X}

\begin{center}
\begin{tabularx}{\textwidth}{l|*{5}{C}|*{5}{C}|*{5}{C}}
\toprule[0.15em]
\rowcolor{color3} & \multicolumn{5}{c|}{Wider-Test} & \multicolumn{5}{c|}{LFW-Test} & \multicolumn{5}{c}{WebPhoto-Test} \\
\rowcolor{color3}
\multirow{-2}{*}{Methods} & C-IQA$\uparrow$ & M-IQA$\uparrow$ & MUSIQ$\uparrow$ & NIQE$\downarrow$ & FID$\downarrow$ & C-IQA$\uparrow$ & M-IQA$\uparrow$ & MUSIQ$\uparrow$ & NIQE$\downarrow$ & FID$\downarrow$ & C-IQA$\uparrow$ & M-IQA$\uparrow$ & MUSIQ$\uparrow$ & NIQE$\downarrow$ & FID$\downarrow$ \\
\midrule[0.15em]
OSEDiff~\cite{wu2024osediff} & 
0.6298 & \textcolor{cvprblue}{0.4951} & \textcolor{cvprblue}{70.559} & \textcolor{cvprblue}{4.9388} & 50.274 &
0.6326 & \textcolor{cvprblue}{0.5037} & \textcolor{cvprblue}{73.401} & \textcolor{cvprblue}{4.7196} & 57.800 &
\textcolor{cvprblue}{0.6457} & \textcolor{cvprblue}{0.5108} & \textcolor{cvprblue}{72.593} & \textcolor{cvprblue}{5.2611} & 117.510 \\

OSEDiff*  & 
0.6193 & 0.4752 & 69.101 & 5.0869 & 47.883 &
0.6186 & 0.4879 & 71.707 & 4.8002 & 51.048 &
0.6254 & 0.4823 & 69.816 & 5.3253 & 109.236 \\

OSEDiff*+VRE  & 
\textcolor{cvprblue}{0.6637} & 0.4834 & 68.259 & 5.0490 & \textcolor{cvprblue}{41.490} &
\textcolor{cvprblue}{0.6608} & 0.5015 & 70.826 & 4.8956 & \textcolor{cvprblue}{46.911} &
0.6410 & 0.4646 & 66.912 & 5.5233 & \textcolor{cvprblue}{95.566} \\

\textbf{OSDFace} (ours) \hspace{0.5mm} & 
\textcolor{red}{0.7284} & \textcolor{red}{0.5229} & \textcolor{red}{74.601} & \textcolor{red}{3.7741} & \textcolor{red}{34.648} &
\textcolor{red}{0.7203} & \textcolor{red}{0.5493} & \textcolor{red}{75.354} & \textcolor{red}{3.8710} & \textcolor{red}{44.629} &
\textcolor{red}{0.7106} & \textcolor{red}{0.5162} & \textcolor{red}{73.935} & \textcolor{red}{3.9864} & \textcolor{red}{84.597} \\

\bottomrule[0.15em]
\end{tabularx}
\end{center}
\vspace{-6.5mm}
\caption{Quantitative comparison on real-world datasets with one-step diffusion methods. C-IQA stands for CLIPIQA, and M-IQA stands for MANIQA. The best and second best results are colored with \textcolor{red}{red} and \textcolor{cvprblue}{blue}, respectively.}
\label{table:OSEDiff-RealWorld}
\vspace{-4mm}
\end{table*}

\begin{table*}[t]
\centering
\begin{minipage}[t]{0.65\textwidth}
\vspace{-18mm}
\scriptsize
\setlength{\tabcolsep}{0.4mm} 
\newcolumntype{?}{!{\vrule width 1pt}}
\newcolumntype{C}{>{\centering\arraybackslash}X}

\begin{center}
\begin{tabularx}{\textwidth}{l|CC|CC|CCCC}
\toprule[0.15em]
\rowcolor{color3}  Methods & LPIPS$\downarrow$ & DISTS$\downarrow$ & MUSIQ$\uparrow$ & NIQE$\downarrow$ & Deg.$\downarrow$ & LMD$\downarrow$ & FID(FFHQ)$\downarrow$ & FID(HQ)$\downarrow$ \\

\midrule[0.15em]

OSEDiff~\cite{wu2024osediff} &
\textcolor{red}{0.3306} & \textcolor{cvprblue}{0.2170} & \textcolor{cvprblue}{71.467} & \textcolor{cvprblue}{5.1241} & 67.390 & \textcolor{cvprblue}{6.4141} & 73.484 & 37.210 \\
OSEDiff* &
0.3496 & 0.2200 & 69.981 & 5.3280 & 67.403 & 7.4082 & 81.362 & 37.131 \\
OSEDiff*+VRE &
0.3368 & 0.2420 & 69.089 & 5.3241 & \textcolor{cvprblue}{63.758} & 6.5365 & \textcolor{cvprblue}{67.785} & \textcolor{cvprblue}{36.356} \\
\textbf{OSDFace} (ours) \hspace{0.5mm} &
\textcolor{cvprblue}{0.3365} & \textcolor{red}{0.1773} & \textcolor{red}{75.640} & \textcolor{red}{3.8840} & \textcolor{red}{60.071} & \textcolor{red}{5.2867} & \textcolor{red}{45.415} & \textcolor{red}{17.062} \\
\bottomrule[0.15em]
\end{tabularx}
\end{center}
\vspace{-6.5mm}
\captionof{table}{Quantitative comparison on the synthetic CelebA-Test dataset with one-step diffusion methods. The best and second best results are colored with \textcolor{red}{red} and \textcolor{cvprblue}{blue}, respectively.}
\label{table:OSEDiff-CelebA-Test}
\end{minipage}%
\hfill
\hspace{-1mm}
\begin{minipage}[t]{0.33\textwidth}
    \centering

    \includegraphics[width=0.323\textwidth]{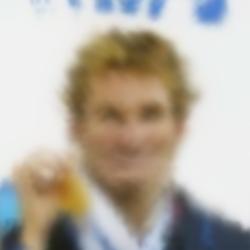} 
    \includegraphics[width=0.323\textwidth]{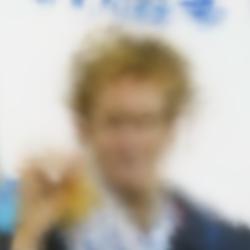}
    \includegraphics[width=0.323\textwidth]{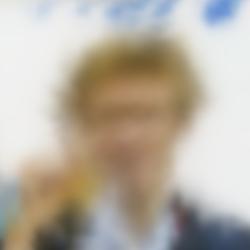}
    \vspace{-6.5mm}
    \captionof{figure}{Visualization of the atmospheric turbulence~\cite{chimitt2020tur_simram} range from 20,000 to 40,000.}
    \label{fig:tur-Visualization}
\end{minipage}
\vspace{-5mm}
\end{table*}

\vspace{-0.5mm}
\section{Validation on Face Recognition}
\label{sec:face_recog}
\vspace{-0.5mm}

Face restoration, as a fundamental low-level vision task, could enhance downstream face recognition tasks to achieve better performance. We use the LFW~\cite{huang2008lfw} dataset as a benchmark for comparison, which includes 3,000 positive pairs and 3,000 negative pairs. Following DAEFR~\cite{tsai2024daefr}, we evaluate the face recognition accuracy using the ArcFace~\cite{deng2019arcface} model under different degradation levels. Specifically, we employ unseen atmospheric turbulence degradation~\cite{chimitt2020tur_simram} to simulate diverse degradation levels, with propagation lengths ranging from 20,000 to 40,000, as illustrated in Fig.~\ref{fig:tur-Visualization}.

\begin{figure*}[t]
\newcommand{\sheetwidth}{0.32}
\vspace{-7mm}
\begin{center}
\hspace{-1mm}
\includegraphics[width=\sheetwidth\textwidth]{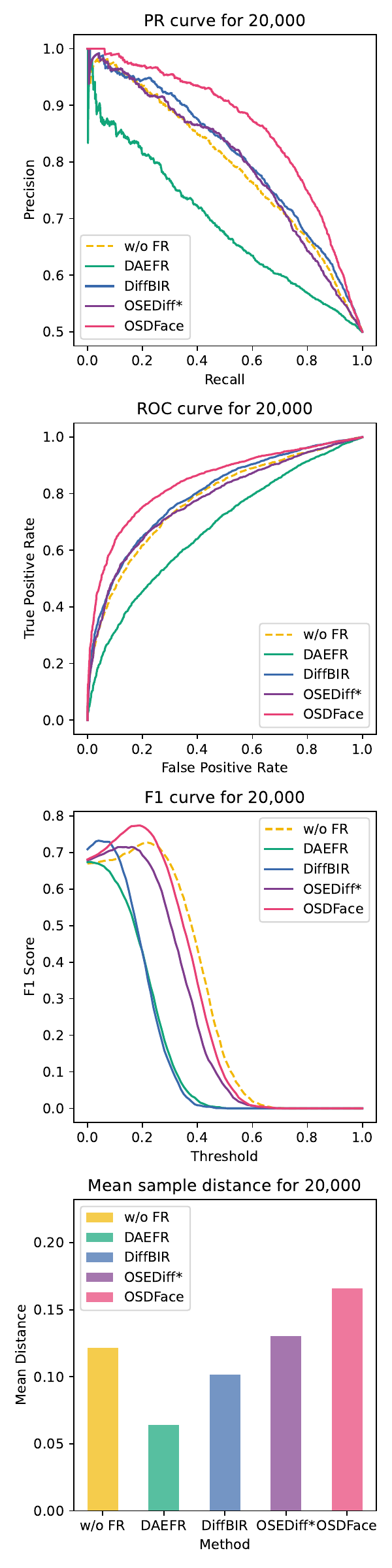}       \hspace{-4mm}
\includegraphics[width=\sheetwidth\textwidth]{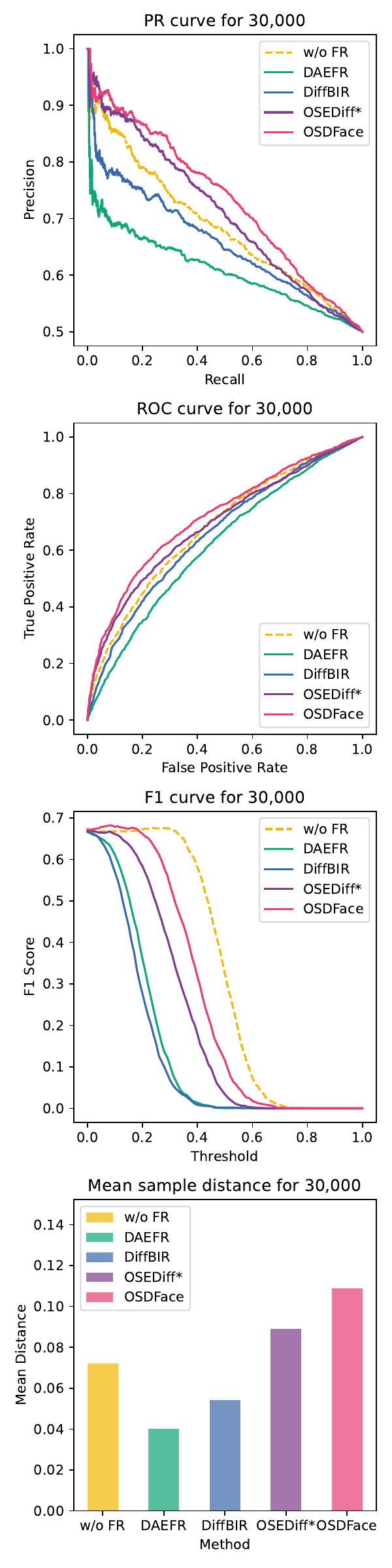}       \hspace{-4mm}
\includegraphics[width=\sheetwidth\textwidth]{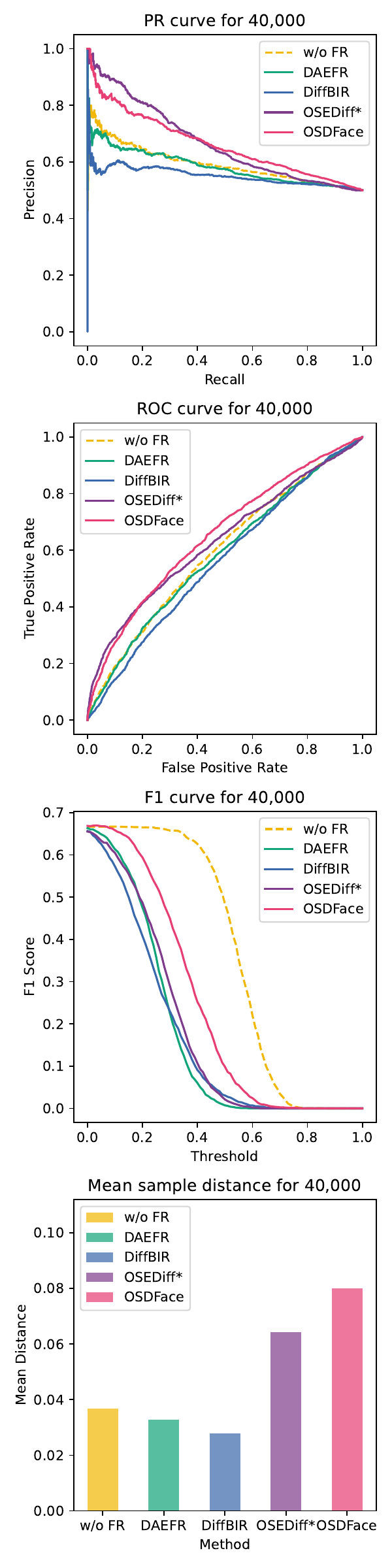}       \hspace{-4mm}
\end{center}
\vspace{-8mm}
\caption{\small Quantitative results on the LFW dataset~\cite{huang2008lfw} for face recognition using the official ArcFace~\cite{deng2019arcface} MS1MV3 R50 model. The evaluated metrics include precision-recall (PR) curves, receiver operating characteristic (ROC) curves, F1 scores, and mean sample distance histograms. The mean sample distance is defined as the difference between the average cosine similarity of predicted positive pairs and predicted negative pairs. “w/o FR” refers to the absence of the face restoration process. Atmospheric turbulence parameters range from 20,000 to 40,000.}
\label{fig:Face_Recognition}
\end{figure*}

The experimental results in Fig.~\ref{fig:Face_Recognition} demonstrate the superior performance of our method across various degradation levels. As degradation severity increases, our method significantly improves precision at the same recall level. The ROC curve shows that OSDFace makes fewer errors at specific true positive rates. Besides, OSDFace widens the gap between positive and negative predictions, thereby improving classifier performance. These findings indicate that our method provides substantial enhancements to downstream face recognition tasks.

\section{Limitations and Future Work} \label{sec:limit}

We briefly analyze the limitations and future work. \textbf{(1)} Color shift: OSDFace sometimes over-enhances contrast or saturation in less degraded regions, causing color shifts in restored face images. Although AdaIN~\cite{huang2017adain} can fix this during inference, we aim for an end-to-end, color shift-free restoration. Future work will explore content-aware color regularization to improve color preservation. \textbf{(2)} Texture in complex regions: OSDFace struggles with realistic skin textures and fine details in complex regions like limbs or fingers. This arises from the model's focus on face features, with limited training data for non-facial parts with similar skin textures. Future work will explore semantic information extraction and domain-specific priors to improve the handling of these areas. \textbf{(3)} Generalization to low-degradation images: OSDFace was not trained on minimal degradation images, but still shows some generalization. However, finer skin texture restoration remains a focus, requiring higher resolution input and output faces, HD training data, and texture-sensitive architectures.

\section{Additional Visual Comparisons}
\label{sec:vis-comp}
These comparisons demonstrate that our proposed OSDFace generates high-quality faces and effectively preserves identities, even with severely degraded input images. Compared to other methods, OSDFace more accurately recovers finer details and produces more realistic faces. To illustrate these advantages further, we select various representative images with unique characteristics, which can be regarded as different face categories. These images are briefly analyzed below. 

\begin{figure*}[t]
\scriptsize
\centering
\newcommand{\widthscale}{0.169}
\newcommand{\dataset}{CelebA}
\newcommand{\leftspace}{-8mm}
\newcommand{\imgid}{00000063}
\newcommand{\imgnote}{0063}

    \hspace{\leftspace}
    \begin{adjustbox}{valign=t}
\begin{tabular}{cccccc}
    \includegraphics[width=\widthscale\textwidth]{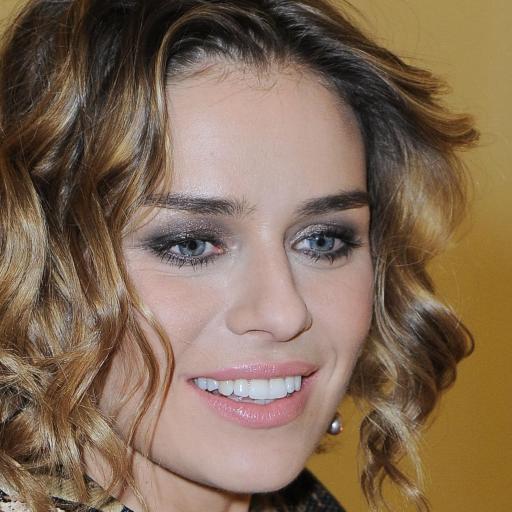} \hspace{-4mm} &
    \includegraphics[width=\widthscale\textwidth]{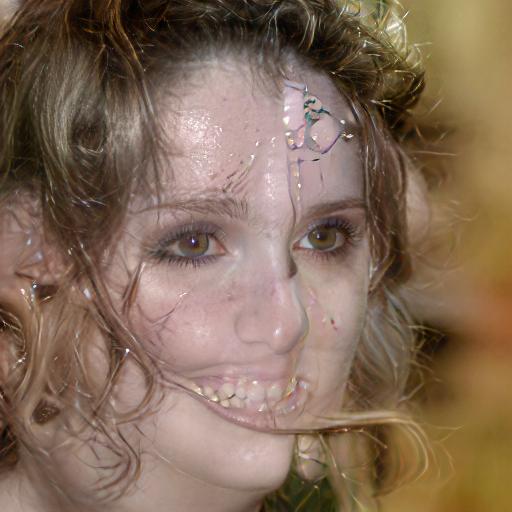} \hspace{-4mm} &
    \includegraphics[width=\widthscale\textwidth]{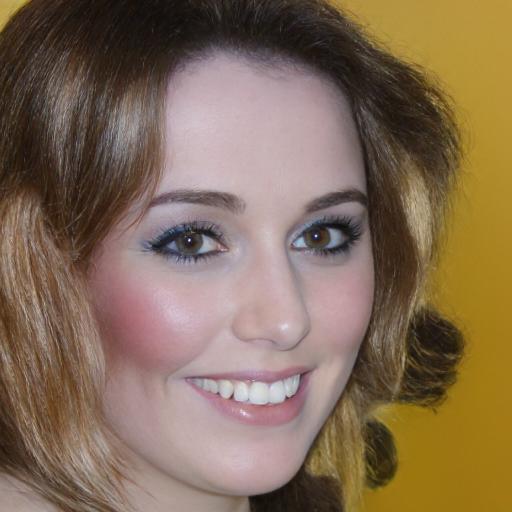} \hspace{-4mm} &
    \includegraphics[width=\widthscale\textwidth]{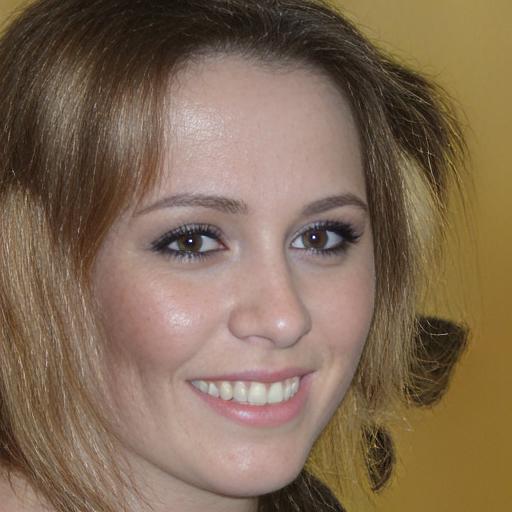} \hspace{-4mm} &
    \includegraphics[width=\widthscale\textwidth]{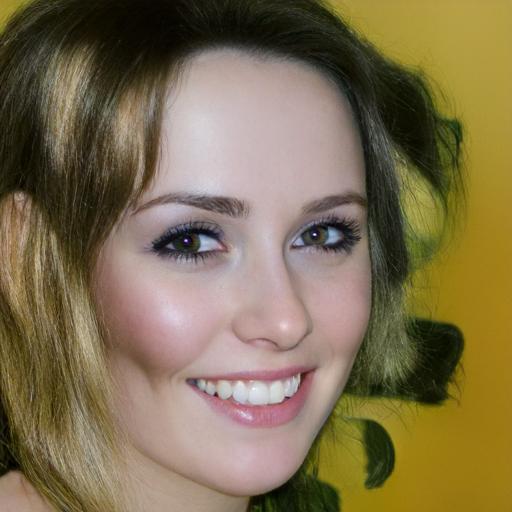} \hspace{-4mm} &
    \includegraphics[width=\widthscale\textwidth]{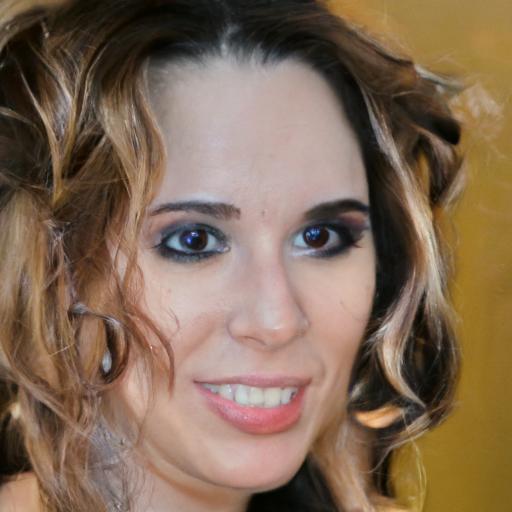} \hspace{-4mm} 
    \\
    HQ - \dataset~\imgnote \hspace{-4mm} &
    RestoreFormer++~\cite{wang2023restoreformer++} \hspace{-4mm} &
    VQFR~\cite{gu2022vqfr} \hspace{-4mm} &
    CodeFormer~\cite{zhou2022codeformer} \hspace{-4mm} &
    DAEFR~\cite{tsai2024daefr} \hspace{-4mm} &
    PGDiff~\cite{yang2023pgdiff} \hspace{-4mm}
    \\
    \includegraphics[width=\widthscale\textwidth]{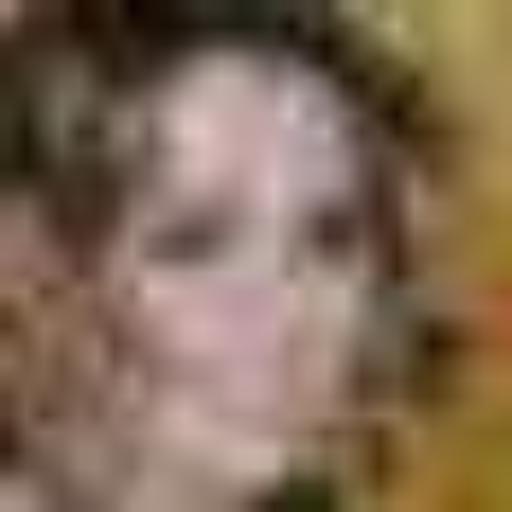} \hspace{-4mm} &
    \includegraphics[width=\widthscale\textwidth]{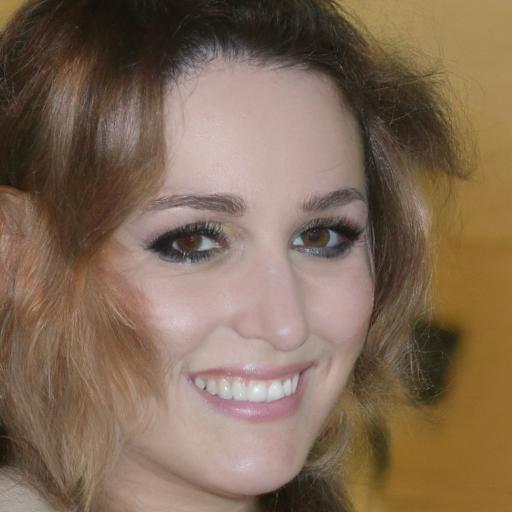} \hspace{-4mm} &
    \includegraphics[width=\widthscale\textwidth]{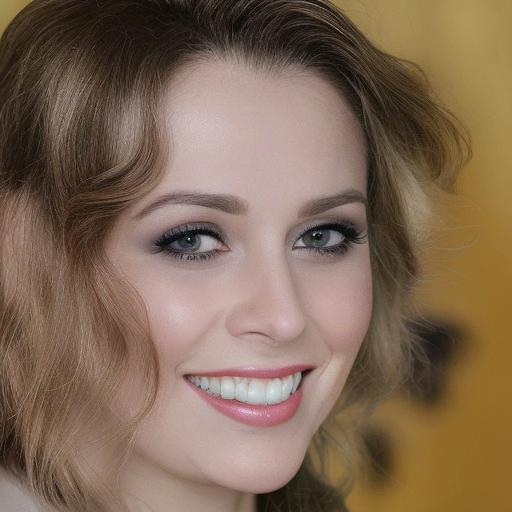} \hspace{-4mm} &
    \includegraphics[width=\widthscale\textwidth]{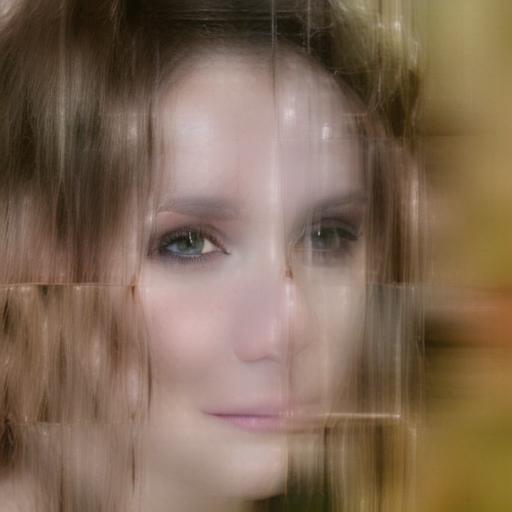} \hspace{-4mm} &
    \includegraphics[width=\widthscale\textwidth]{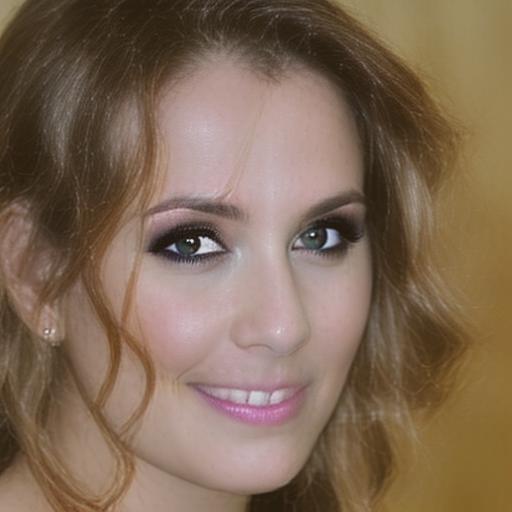} \hspace{-4mm} &
    \includegraphics[width=\widthscale\textwidth]{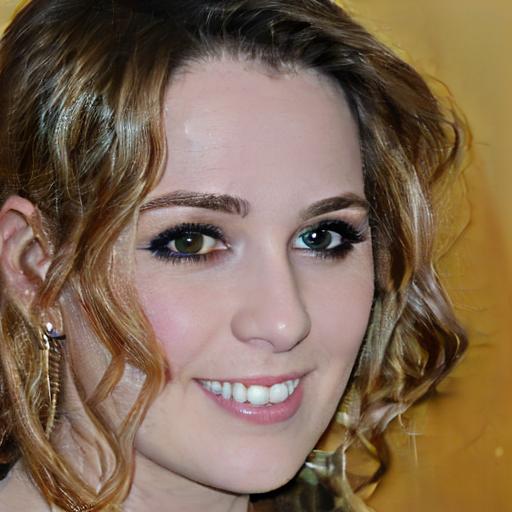} \hspace{-4mm} 
    \\ 
    LQ - \dataset~\imgnote \hspace{-4mm} &
    DifFace~\cite{yue2024difface} \hspace{-4mm} &
    DiffBIR~\cite{lin2024diffbir} \hspace{-4mm} &
    OSEDiff~\cite{wu2024osediff}  \hspace{-4mm} &
    OSEDiff*~\cite{wu2024osediff}  \hspace{-4mm} &
    OSDFace (ours) \hspace{-4mm}
    \\
    \end{tabular}
    \end{adjustbox}

\renewcommand{\imgid}{00000064}
\renewcommand{\imgnote}{0064}

    \hspace{\leftspace}
    \begin{adjustbox}{valign=t}
    \begin{tabular}{cccccc}
    \includegraphics[width=\widthscale\textwidth]{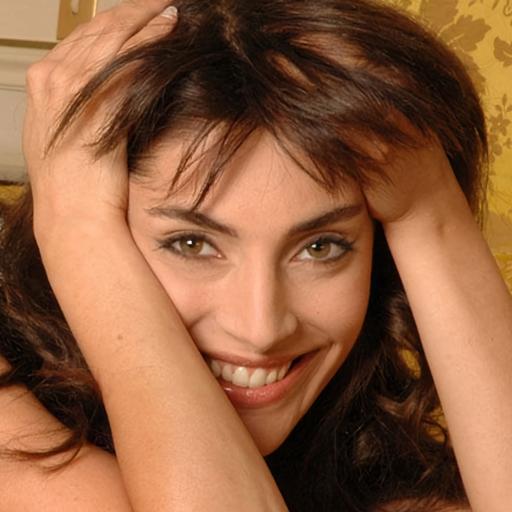} \hspace{-4mm} &
    \includegraphics[width=\widthscale\textwidth]{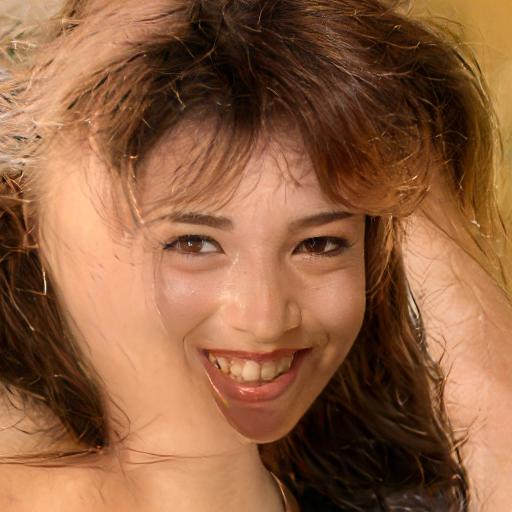} \hspace{-4mm} &
    \includegraphics[width=\widthscale\textwidth]{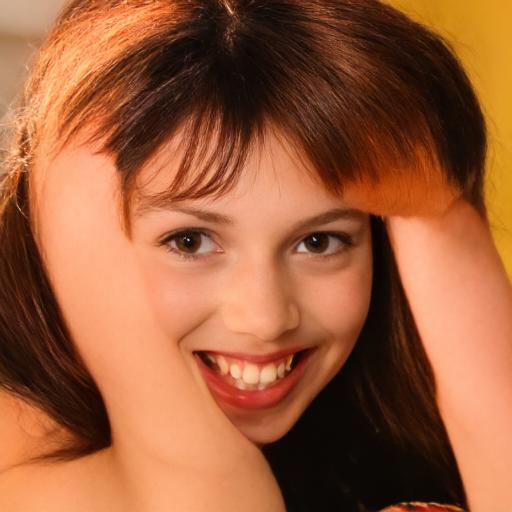} \hspace{-4mm} &
    \includegraphics[width=\widthscale\textwidth]{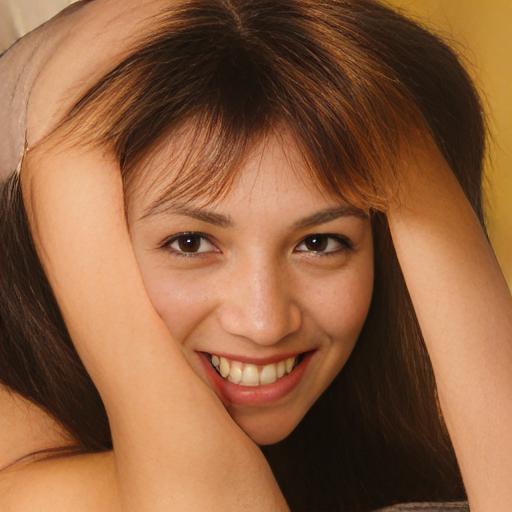} \hspace{-4mm} &
    \includegraphics[width=\widthscale\textwidth]{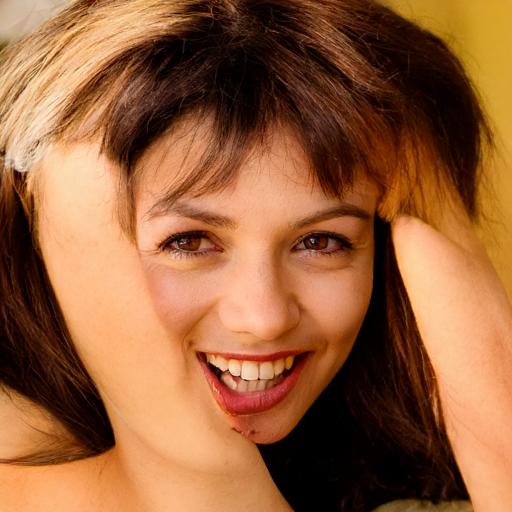} \hspace{-4mm} &
    \includegraphics[width=\widthscale\textwidth]{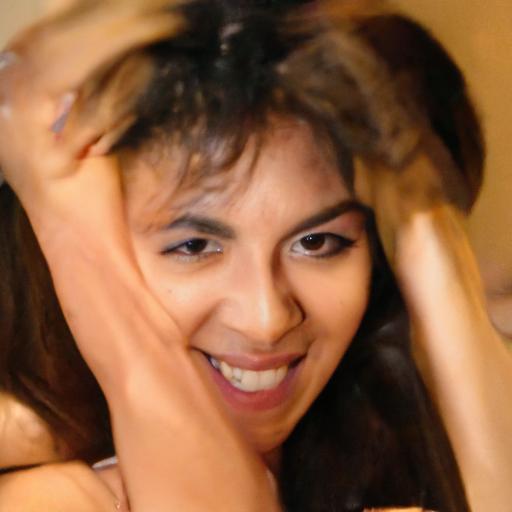} \hspace{-4mm} 
    \\
    HQ - \dataset~\imgnote \hspace{-4mm} &
    RestoreFormer++~\cite{wang2023restoreformer++} \hspace{-4mm} &
    VQFR~\cite{gu2022vqfr} \hspace{-4mm} &
    CodeFormer~\cite{zhou2022codeformer} \hspace{-4mm} &
    DAEFR~\cite{tsai2024daefr} \hspace{-4mm} &
    PGDiff~\cite{yang2023pgdiff} \hspace{-4mm}
    \\
    \includegraphics[width=\widthscale\textwidth]{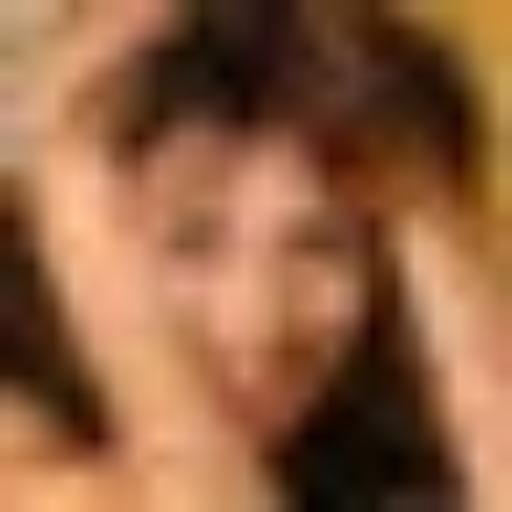} \hspace{-4mm} &
    \includegraphics[width=\widthscale\textwidth]{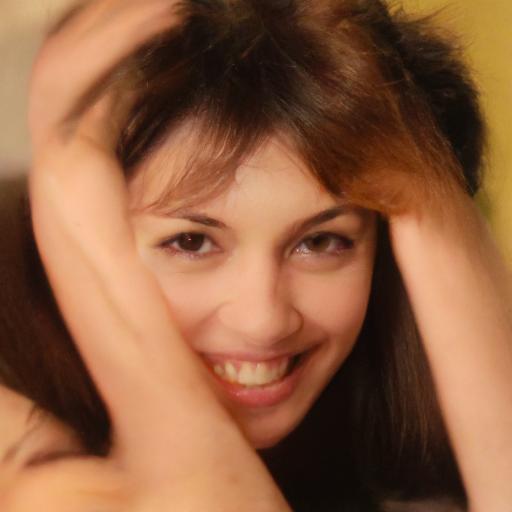} \hspace{-4mm} &
    \includegraphics[width=\widthscale\textwidth]{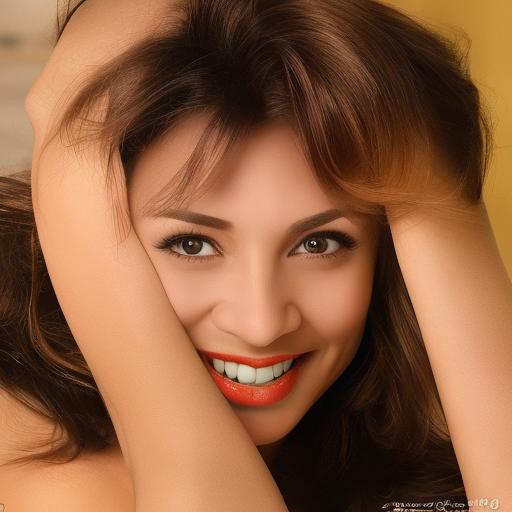} \hspace{-4mm} &
    \includegraphics[width=\widthscale\textwidth]{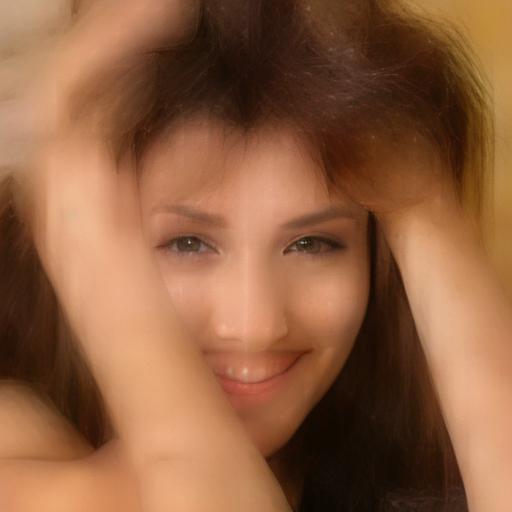} \hspace{-4mm} &
    \includegraphics[width=\widthscale\textwidth]{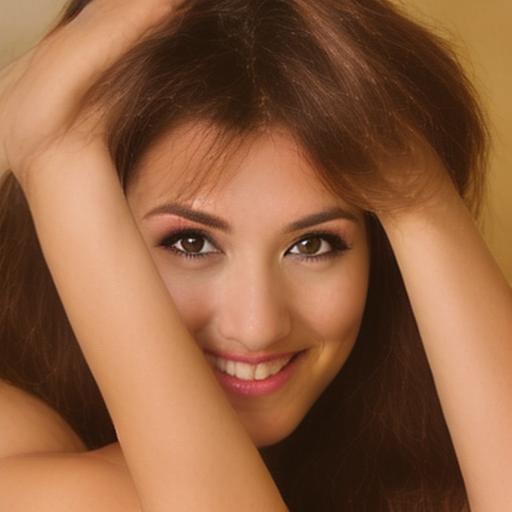} \hspace{-4mm} &
    \includegraphics[width=\widthscale\textwidth]{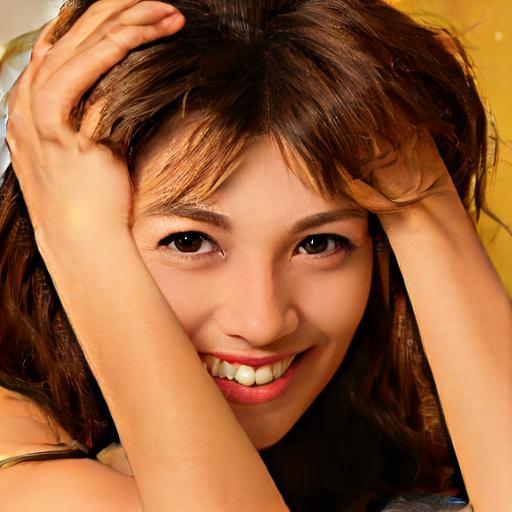} \hspace{-4mm} 
    \\ 
    LQ - \dataset~\imgnote \hspace{-4mm} &
    DifFace~\cite{yue2024difface} \hspace{-4mm} &
    DiffBIR~\cite{lin2024diffbir} \hspace{-4mm} &
    OSEDiff~\cite{wu2024osediff}  \hspace{-4mm} &
    OSEDiff*~\cite{wu2024osediff}  \hspace{-4mm} &
    OSDFace (ours) \hspace{-4mm}
    \\
    \end{tabular}
    \end{adjustbox}

\renewcommand{\imgid}{00000080}
\renewcommand{\imgnote}{0080}

    \hspace{\leftspace}
    \begin{adjustbox}{valign=t}
    \begin{tabular}{cccccc}
    \includegraphics[width=\widthscale\textwidth]{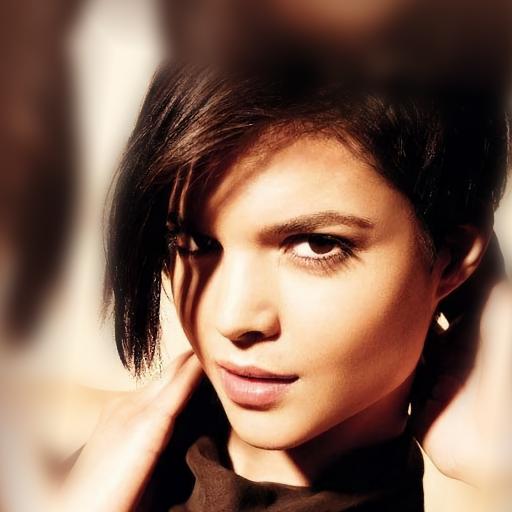} \hspace{-4mm} &
    \includegraphics[width=\widthscale\textwidth]{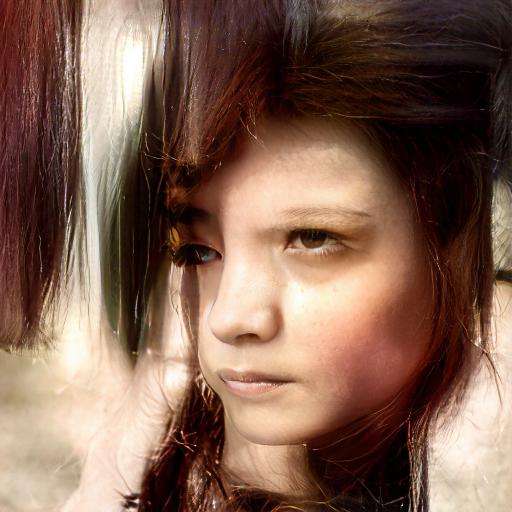} \hspace{-4mm} &
    \includegraphics[width=\widthscale\textwidth]{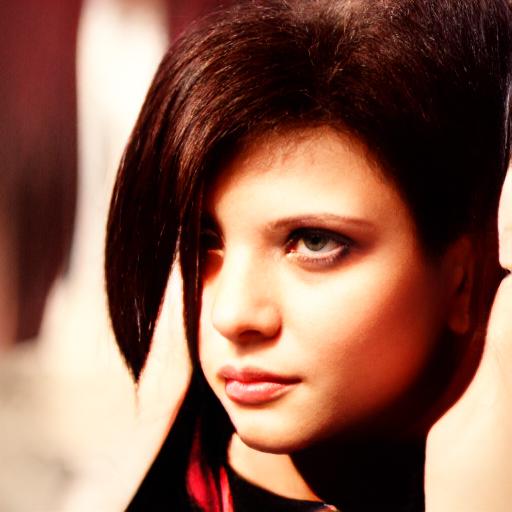} \hspace{-4mm} &
    \includegraphics[width=\widthscale\textwidth]{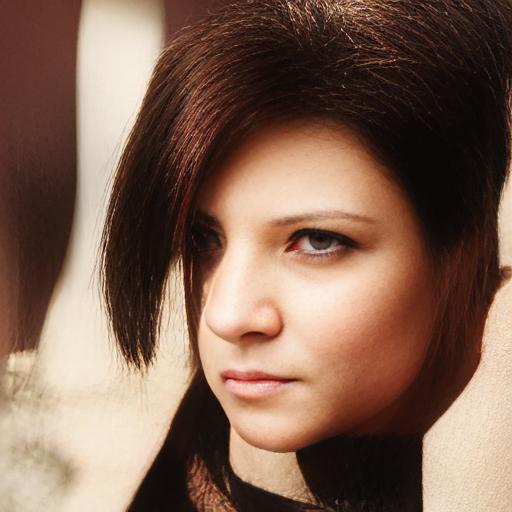} \hspace{-4mm} &
    \includegraphics[width=\widthscale\textwidth]{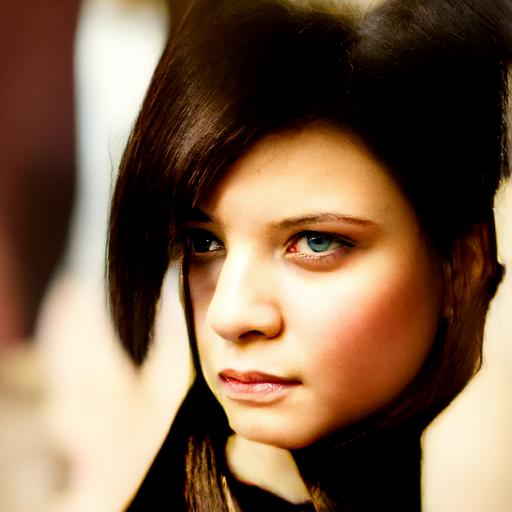} \hspace{-4mm} &
    \includegraphics[width=\widthscale\textwidth]{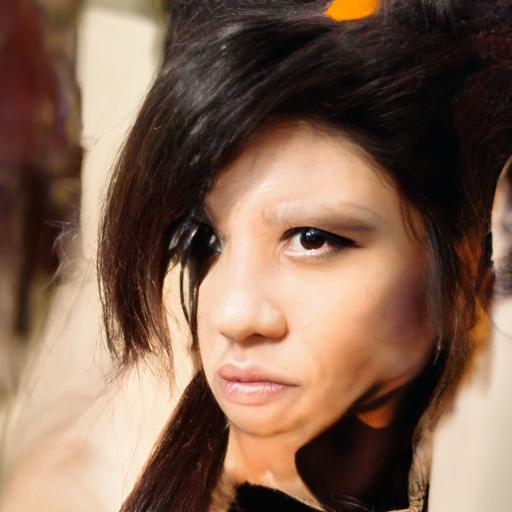} \hspace{-4mm} 
    \\
    HQ - \dataset~\imgnote \hspace{-4mm} &
    RestoreFormer++~\cite{wang2023restoreformer++} \hspace{-4mm} &
    VQFR~\cite{gu2022vqfr} \hspace{-4mm} &
    CodeFormer~\cite{zhou2022codeformer} \hspace{-4mm} &
    DAEFR~\cite{tsai2024daefr} \hspace{-4mm} &
    PGDiff~\cite{yang2023pgdiff} \hspace{-4mm}
    \\
    \includegraphics[width=\widthscale\textwidth]{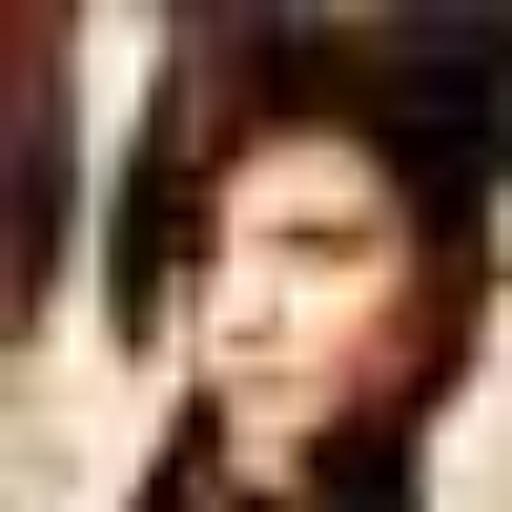} \hspace{-4mm} &
    \includegraphics[width=\widthscale\textwidth]{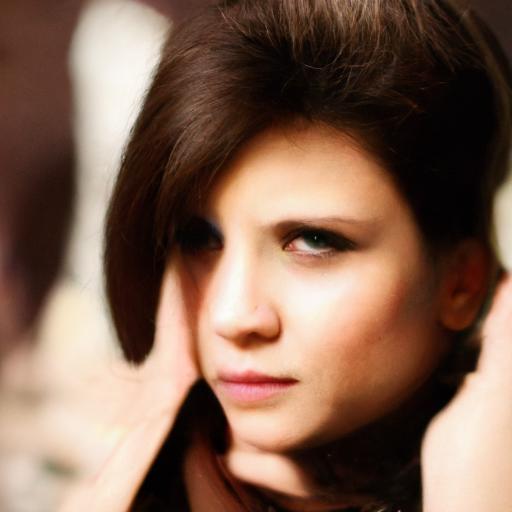} \hspace{-4mm} &
    \includegraphics[width=\widthscale\textwidth]{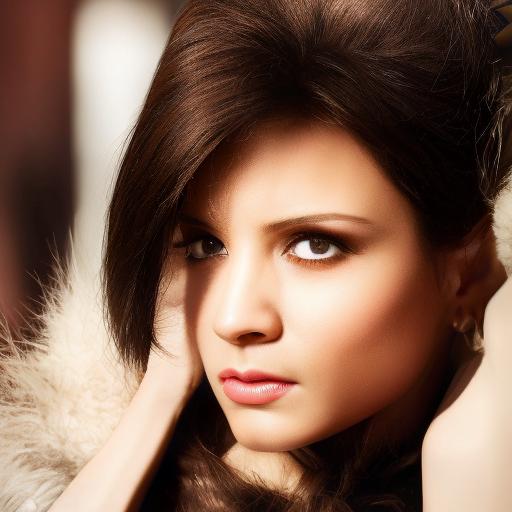} \hspace{-4mm} &
    \includegraphics[width=\widthscale\textwidth]{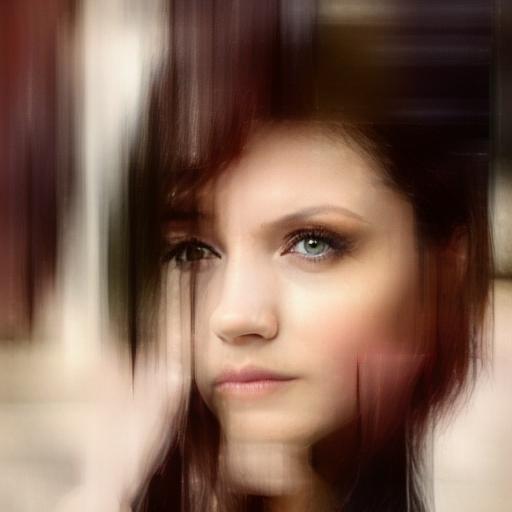} \hspace{-4mm} &
    \includegraphics[width=\widthscale\textwidth]{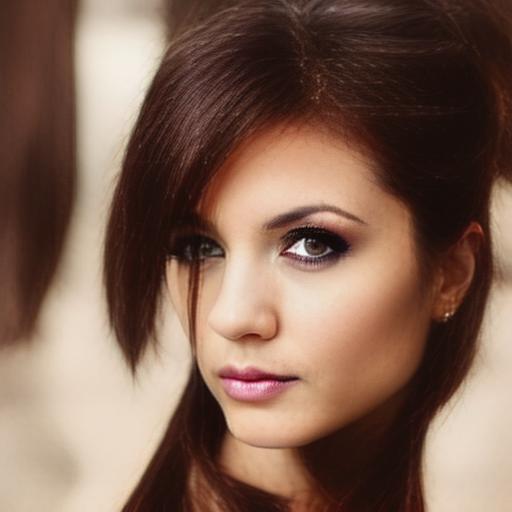} \hspace{-4mm} &
    \includegraphics[width=\widthscale\textwidth]{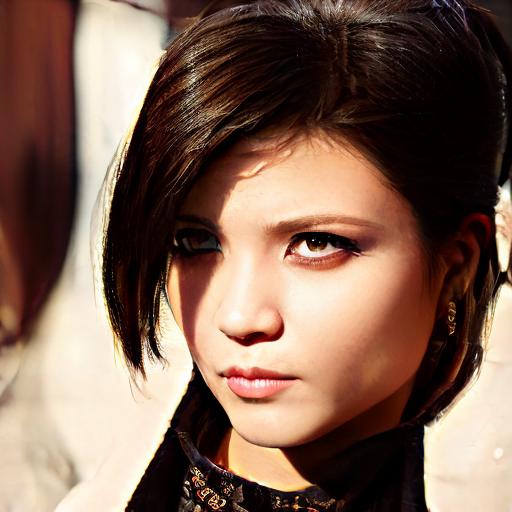} \hspace{-4mm} 
    \\ 
    LQ - \dataset~\imgnote \hspace{-4mm} &
    DifFace~\cite{yue2024difface} \hspace{-4mm} &
    DiffBIR~\cite{lin2024diffbir} \hspace{-4mm} &
    OSEDiff~\cite{wu2024osediff}  \hspace{-4mm} &
    OSEDiff*~\cite{wu2024osediff}  \hspace{-4mm} &
    OSDFace (ours) \hspace{-4mm}
    \\
    \end{tabular}
    \end{adjustbox}

\vspace{-2mm}
\caption{More visual comparison of the synthetic CelebA-Test dataset in challenging cases. Please zoom in for a better view.}
\label{fig:vis-supp-celeba-1}
\vspace{-2mm}
\end{figure*}

\begin{figure*}[t]
\scriptsize
\centering
\newcommand{\widthscale}{0.169}
\newcommand{\dataset}{CelebA}
\newcommand{\leftspace}{-8mm}
\newcommand{\imgid}{00000124}
\newcommand{\imgnote}{0124}

    \hspace{\leftspace}
    \begin{adjustbox}{valign=t}
    \begin{tabular}{cccccc}
    \includegraphics[width=\widthscale\textwidth]{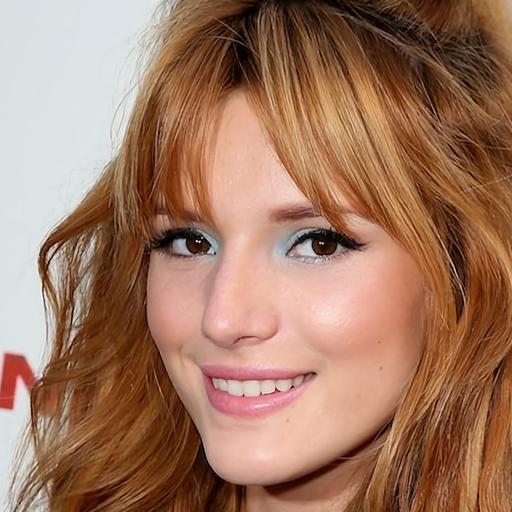} \hspace{-4mm} &
    \includegraphics[width=\widthscale\textwidth]{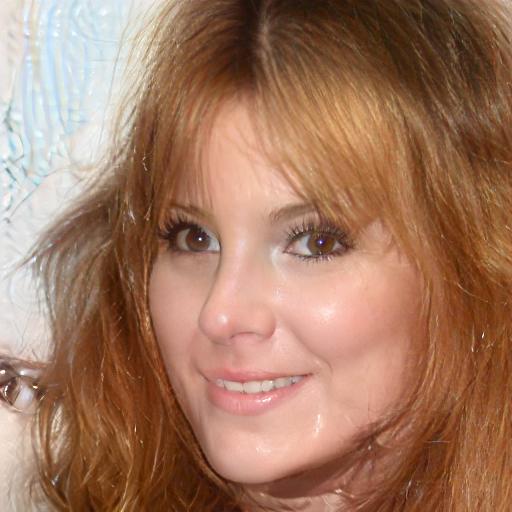} \hspace{-4mm} &
    \includegraphics[width=\widthscale\textwidth]{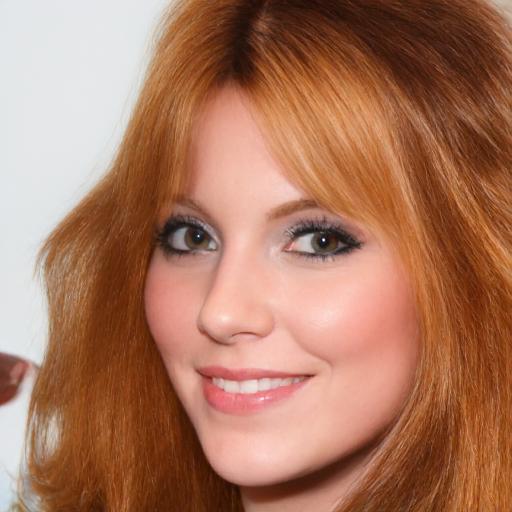} \hspace{-4mm} &
    \includegraphics[width=\widthscale\textwidth]{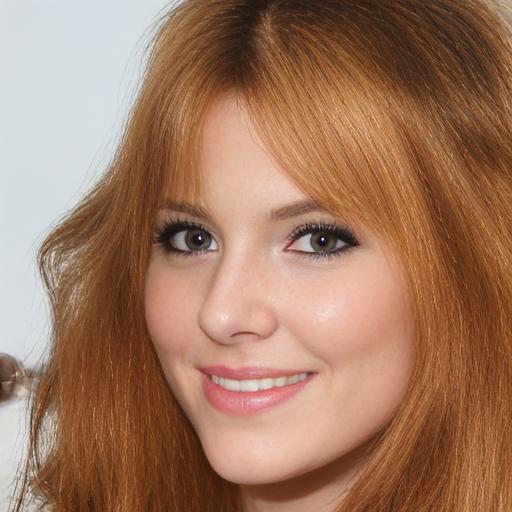} \hspace{-4mm} &
    \includegraphics[width=\widthscale\textwidth]{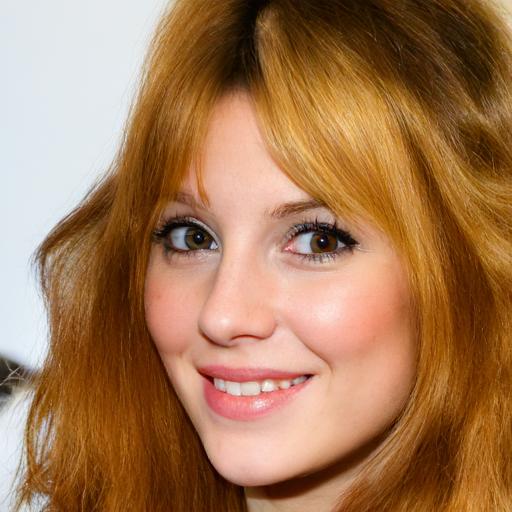} \hspace{-4mm} &
    \includegraphics[width=\widthscale\textwidth]{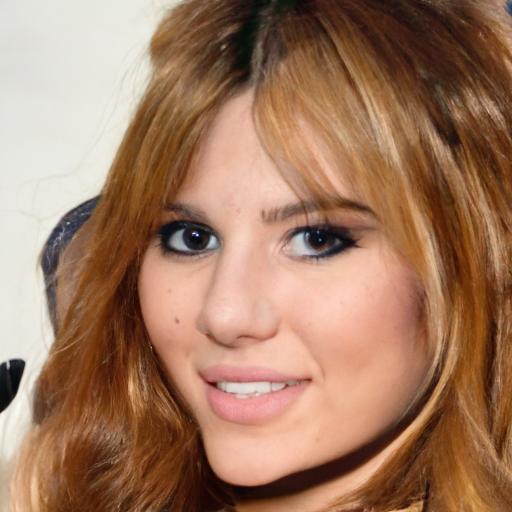} \hspace{-4mm} 
    \\
    HQ - \dataset~\imgnote \hspace{-4mm} &
    RestoreFormer++~\cite{wang2023restoreformer++} \hspace{-4mm} &
    VQFR~\cite{gu2022vqfr} \hspace{-4mm} &
    CodeFormer~\cite{zhou2022codeformer} \hspace{-4mm} &
    DAEFR~\cite{tsai2024daefr} \hspace{-4mm} &
    PGDiff~\cite{yang2023pgdiff} \hspace{-4mm}
    \\
    \includegraphics[width=\widthscale\textwidth]{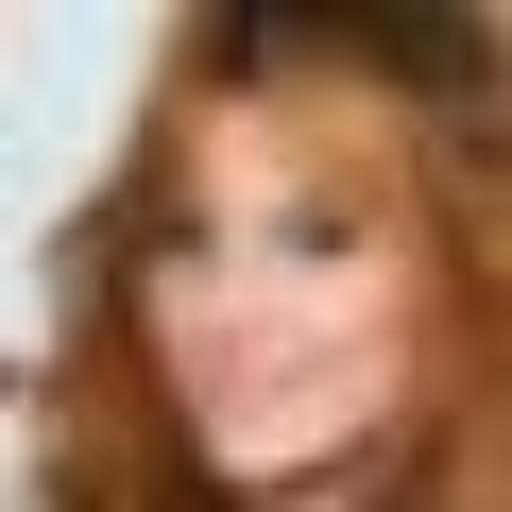} \hspace{-4mm} &
    \includegraphics[width=\widthscale\textwidth]{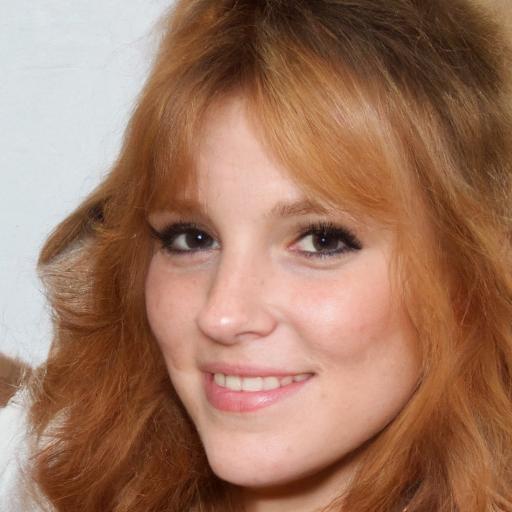} \hspace{-4mm} &
    \includegraphics[width=\widthscale\textwidth]{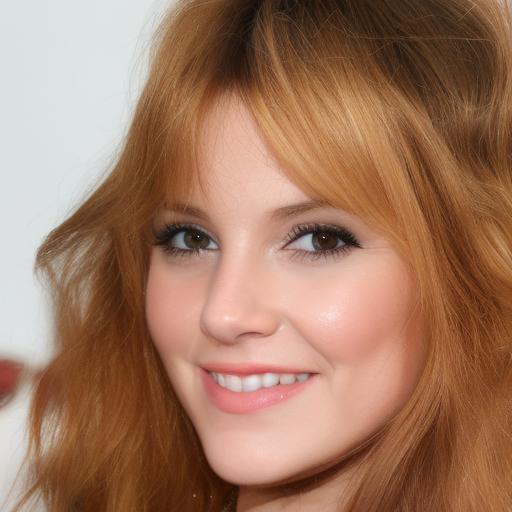} \hspace{-4mm} &
    \includegraphics[width=\widthscale\textwidth]{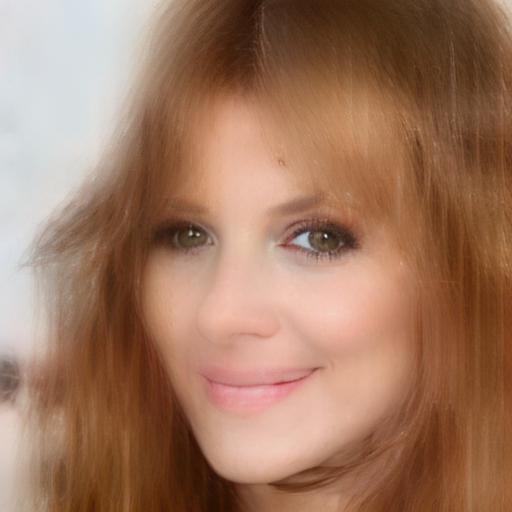} \hspace{-4mm} &
    \includegraphics[width=\widthscale\textwidth]{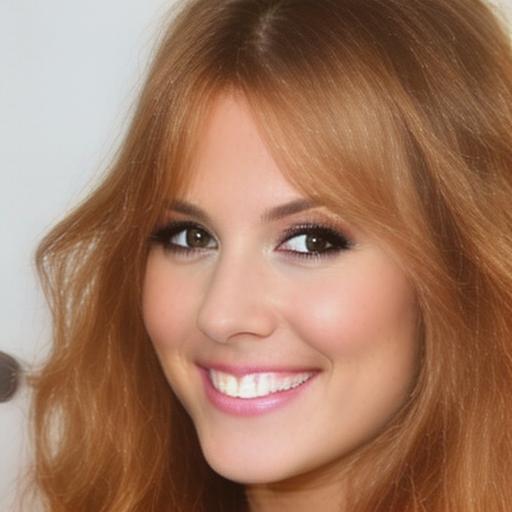} \hspace{-4mm} &
    \includegraphics[width=\widthscale\textwidth]{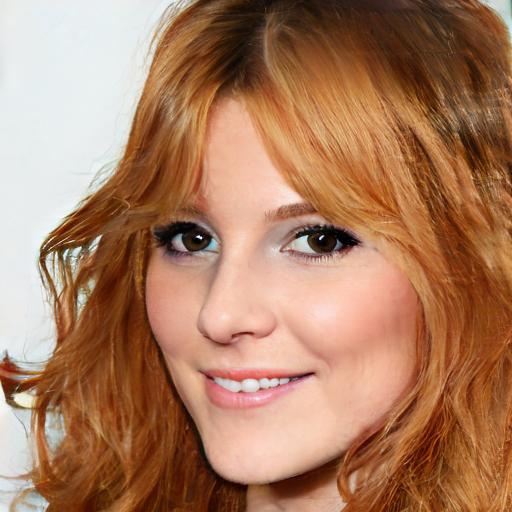} \hspace{-4mm} 
    \\ 
    LQ - \dataset~\imgnote \hspace{-4mm} &
    DifFace~\cite{yue2024difface} \hspace{-4mm} &
    DiffBIR~\cite{lin2024diffbir} \hspace{-4mm} &
    OSEDiff~\cite{wu2024osediff}  \hspace{-4mm} &
    OSEDiff*~\cite{wu2024osediff}  \hspace{-4mm} &
    OSDFace (ours) \hspace{-4mm}
    \\
    \end{tabular}
    \end{adjustbox}

\renewcommand{\imgid}{00000125}
\renewcommand{\imgnote}{0125}

    \hspace{\leftspace}
    \begin{adjustbox}{valign=t}
    \begin{tabular}{cccccc}
    \includegraphics[width=\widthscale\textwidth]{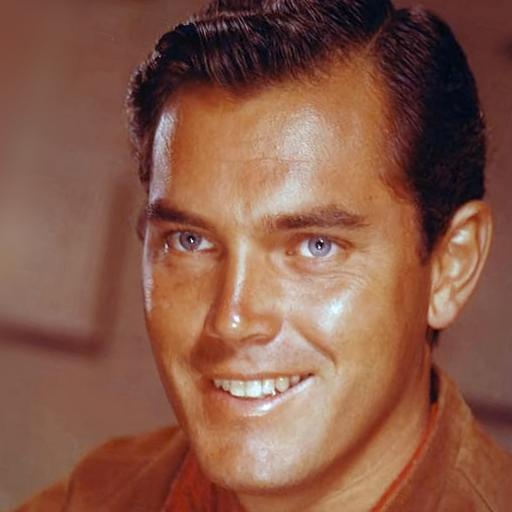} \hspace{-4mm} &
    \includegraphics[width=\widthscale\textwidth]{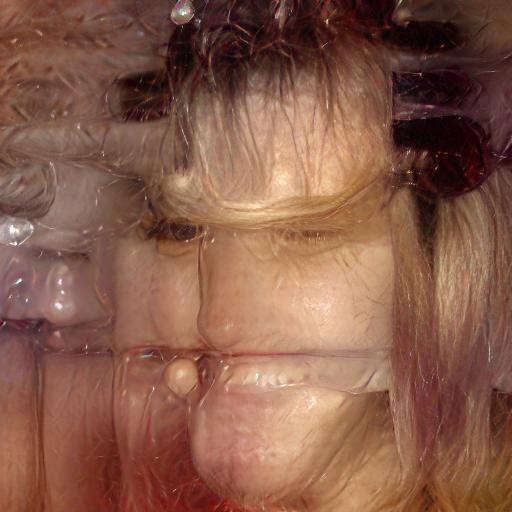} \hspace{-4mm} &
    \includegraphics[width=\widthscale\textwidth]{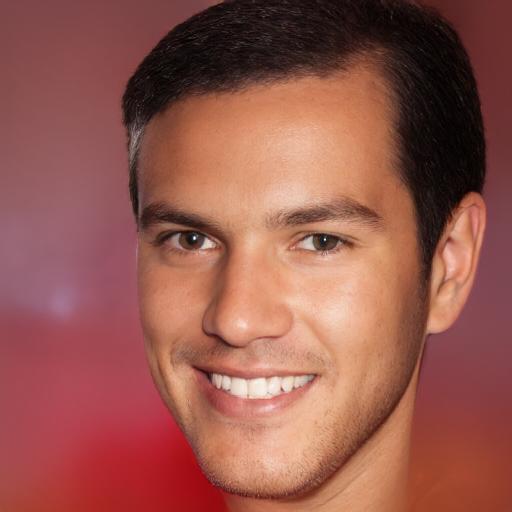} \hspace{-4mm} &
    \includegraphics[width=\widthscale\textwidth]{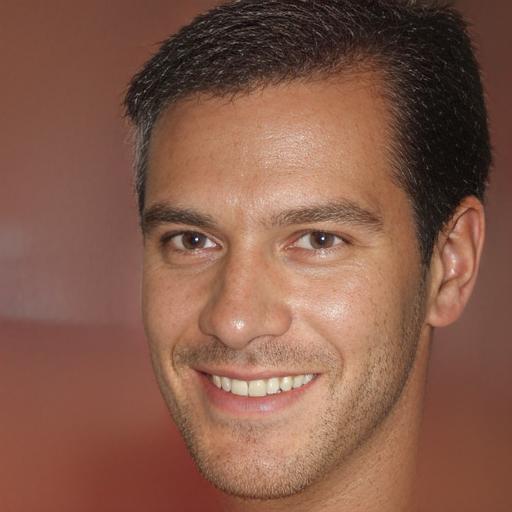} \hspace{-4mm} &
    \includegraphics[width=\widthscale\textwidth]{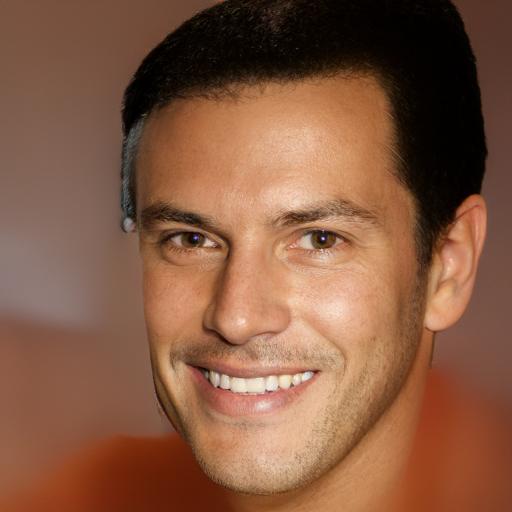} \hspace{-4mm} &
    \includegraphics[width=\widthscale\textwidth]{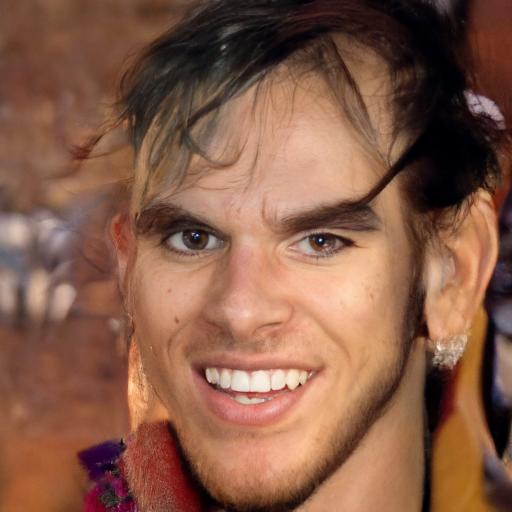} \hspace{-4mm} 
    \\
    HQ - \dataset~\imgnote \hspace{-4mm} &
    RestoreFormer++~\cite{wang2023restoreformer++} \hspace{-4mm} &
    VQFR~\cite{gu2022vqfr} \hspace{-4mm} &
    CodeFormer~\cite{zhou2022codeformer} \hspace{-4mm} &
    DAEFR~\cite{tsai2024daefr} \hspace{-4mm} &
    PGDiff~\cite{yang2023pgdiff} \hspace{-4mm}
    \\
    \includegraphics[width=\widthscale\textwidth]{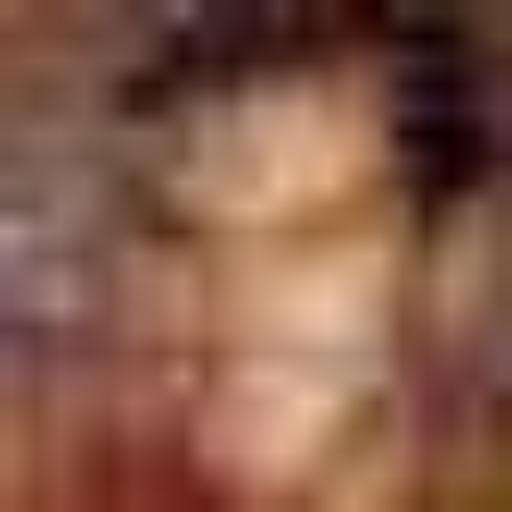} \hspace{-4mm} &
    \includegraphics[width=\widthscale\textwidth]{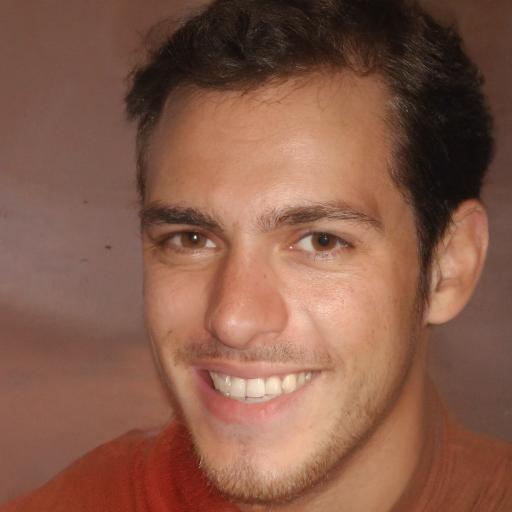} \hspace{-4mm} &
    \includegraphics[width=\widthscale\textwidth]{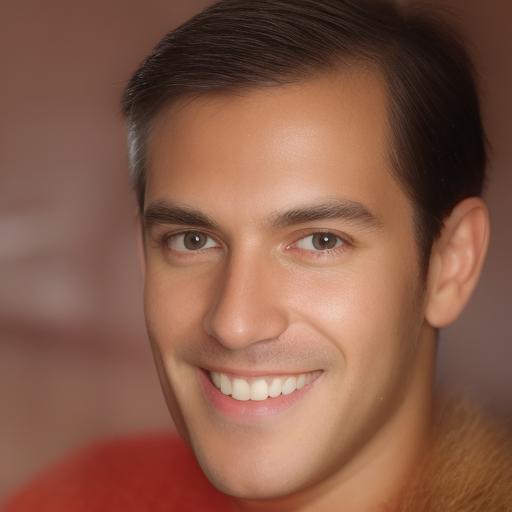} \hspace{-4mm} &
    \includegraphics[width=\widthscale\textwidth]{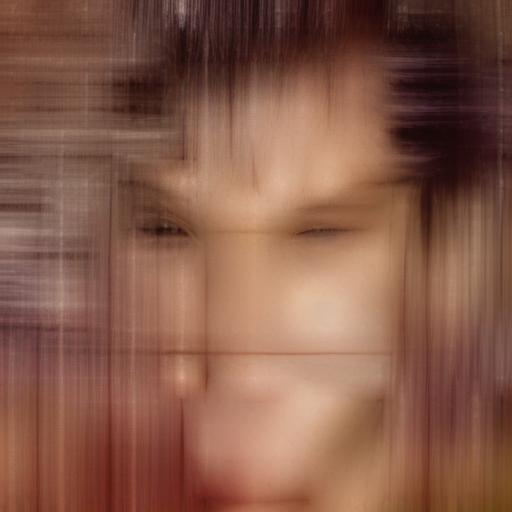} \hspace{-4mm} &
    \includegraphics[width=\widthscale\textwidth]{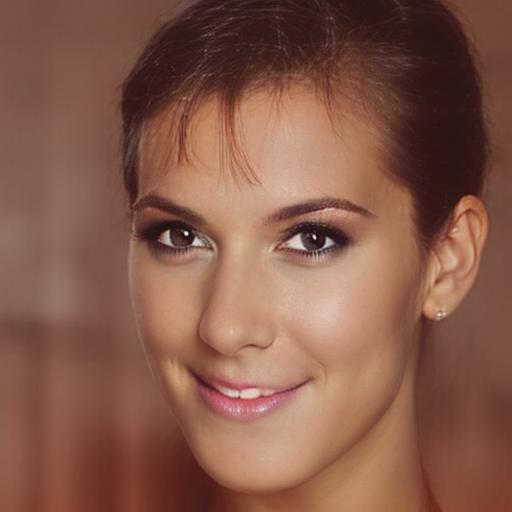} \hspace{-4mm} &
    \includegraphics[width=\widthscale\textwidth]{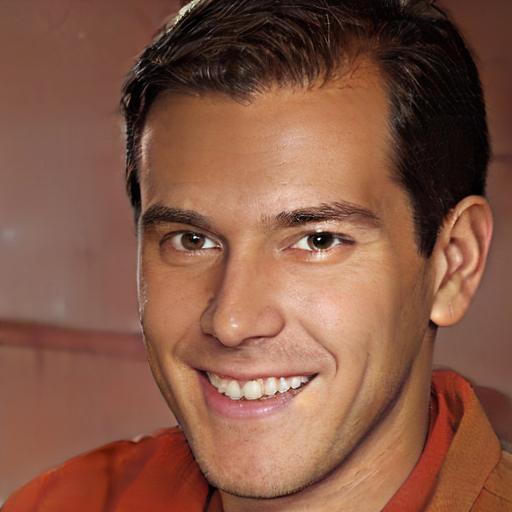} \hspace{-4mm} 
    \\ 
    LQ - \dataset~\imgnote \hspace{-4mm} &
    DifFace~\cite{yue2024difface} \hspace{-4mm} &
    DiffBIR~\cite{lin2024diffbir} \hspace{-4mm} &
    OSEDiff~\cite{wu2024osediff}  \hspace{-4mm} &
    OSEDiff*~\cite{wu2024osediff}  \hspace{-4mm} &
    OSDFace (ours) \hspace{-4mm}
    \\
    \end{tabular}
    \end{adjustbox}

\renewcommand{\imgid}{00000186}
\renewcommand{\imgnote}{0186}

    \hspace{\leftspace}
    \begin{adjustbox}{valign=t}
    \begin{tabular}{cccccc}
    \includegraphics[width=\widthscale\textwidth]{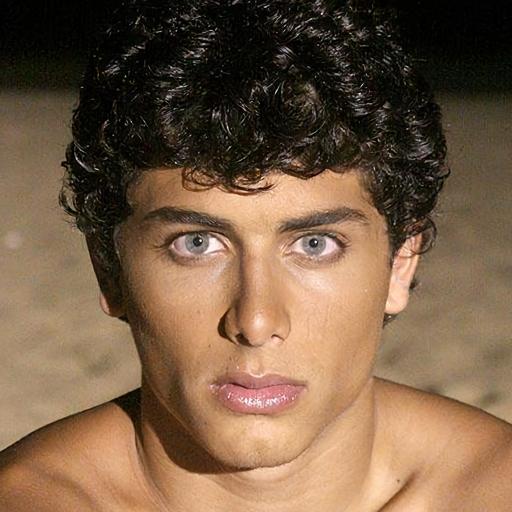} \hspace{-4mm} &
    \includegraphics[width=\widthscale\textwidth]{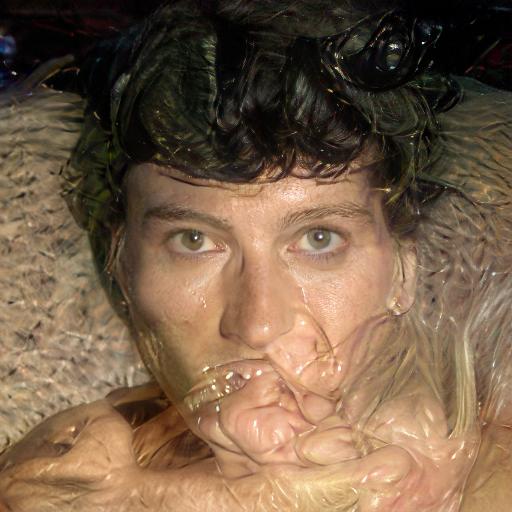} \hspace{-4mm} &
    \includegraphics[width=\widthscale\textwidth]{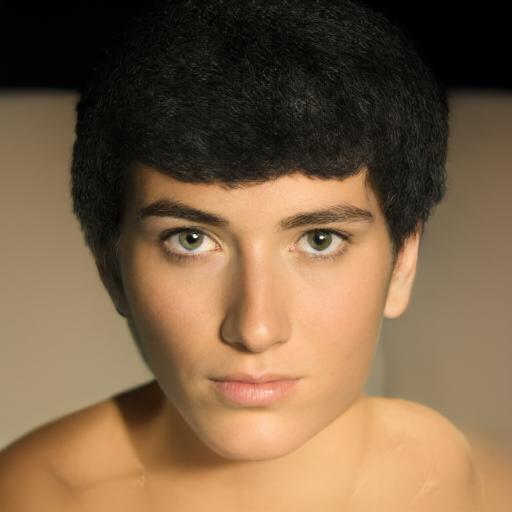} \hspace{-4mm} &
    \includegraphics[width=\widthscale\textwidth]{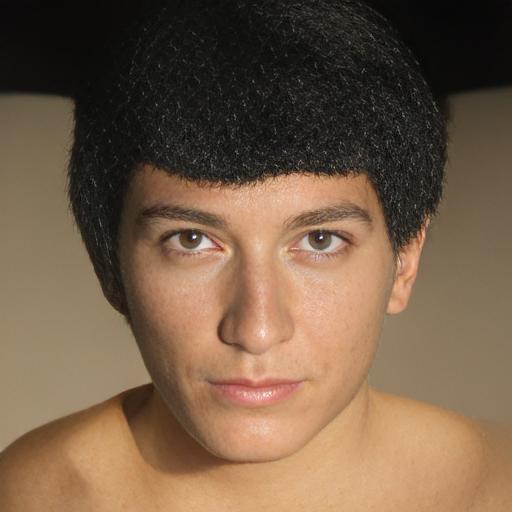} \hspace{-4mm} &
    \includegraphics[width=\widthscale\textwidth]{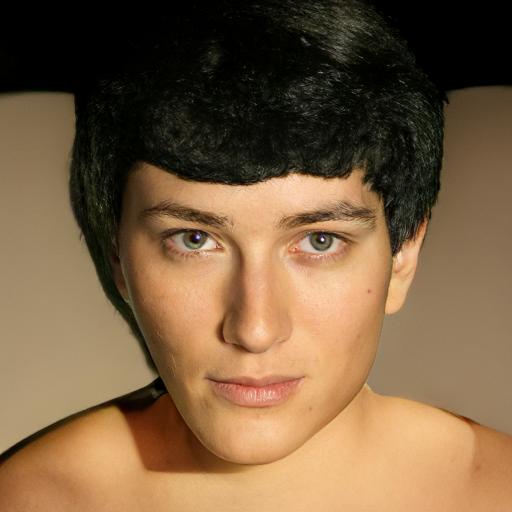} \hspace{-4mm} &
    \includegraphics[width=\widthscale\textwidth]{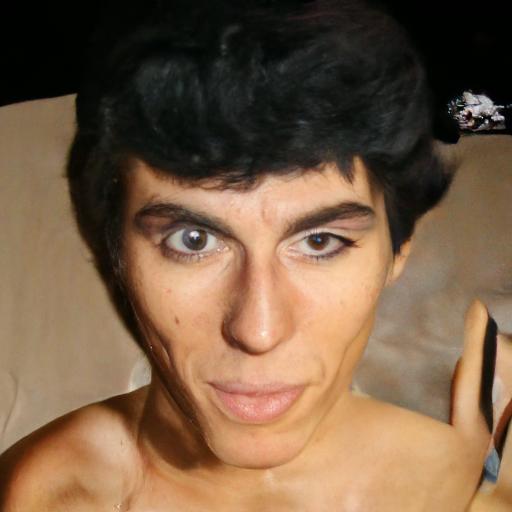} \hspace{-4mm} 
    \\
    HQ - \dataset~\imgnote \hspace{-4mm} &
    RestoreFormer++~\cite{wang2023restoreformer++} \hspace{-4mm} &
    VQFR~\cite{gu2022vqfr} \hspace{-4mm} &
    CodeFormer~\cite{zhou2022codeformer} \hspace{-4mm} &
    DAEFR~\cite{tsai2024daefr} \hspace{-4mm} &
    PGDiff~\cite{yang2023pgdiff} \hspace{-4mm}
    \\
    \includegraphics[width=\widthscale\textwidth]{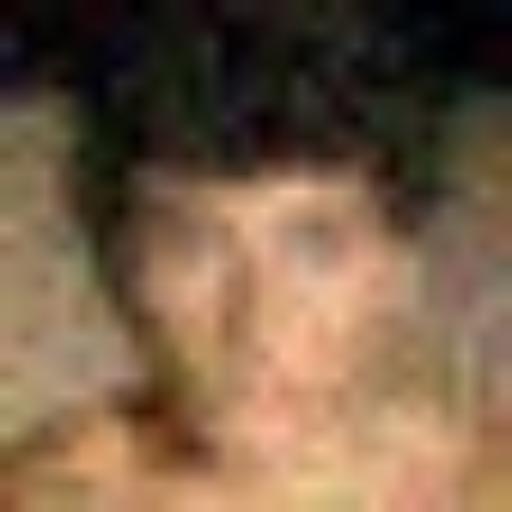} \hspace{-4mm} &
    \includegraphics[width=\widthscale\textwidth]{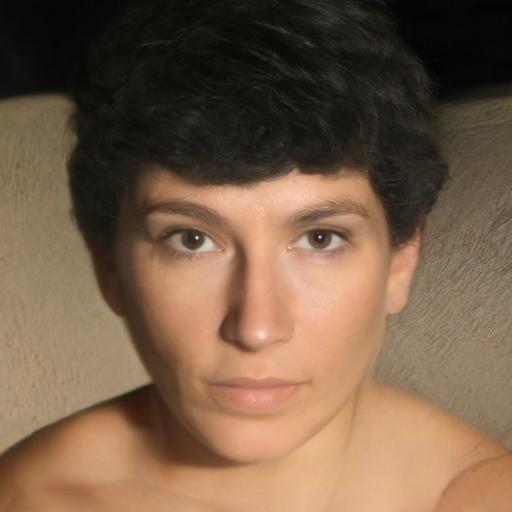} \hspace{-4mm} &
    \includegraphics[width=\widthscale\textwidth]{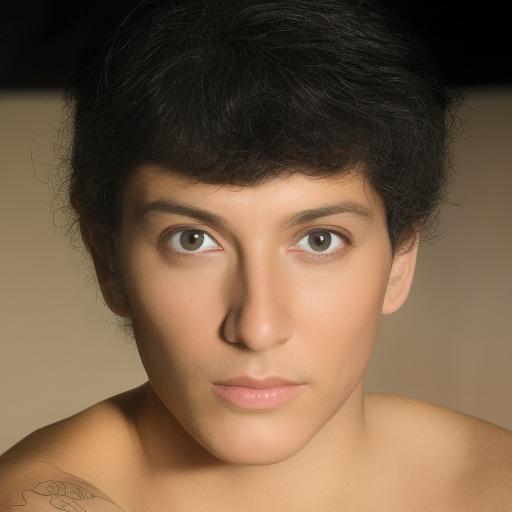} \hspace{-4mm} &
    \includegraphics[width=\widthscale\textwidth]{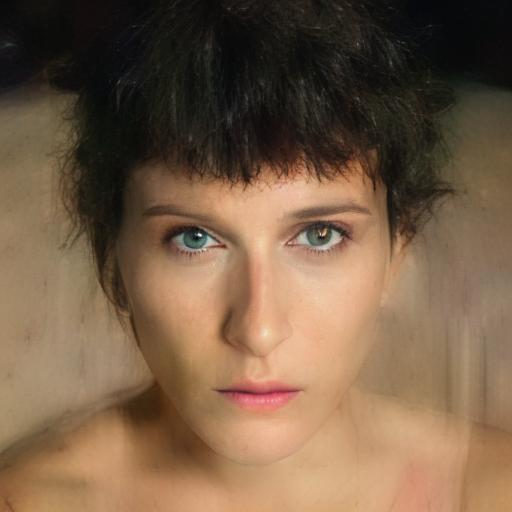} \hspace{-4mm} &
    \includegraphics[width=\widthscale\textwidth]{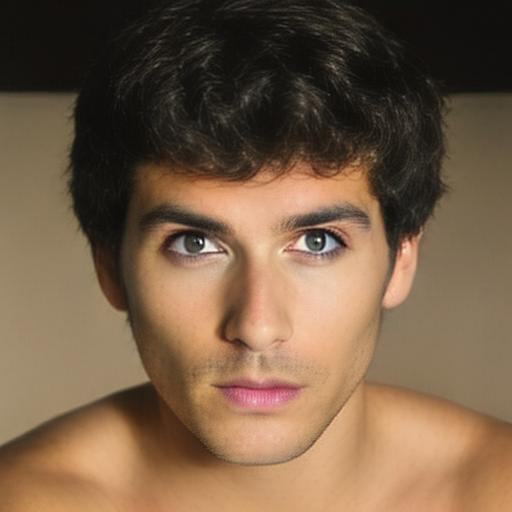} \hspace{-4mm} &
    \includegraphics[width=\widthscale\textwidth]{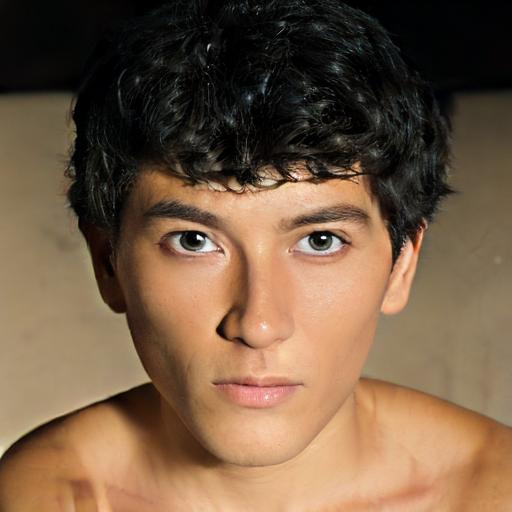} \hspace{-4mm} 
    \\ 
    LQ - \dataset~\imgnote \hspace{-4mm} &
    DifFace~\cite{yue2024difface} \hspace{-4mm} &
    DiffBIR~\cite{lin2024diffbir} \hspace{-4mm} &
    OSEDiff~\cite{wu2024osediff}  \hspace{-4mm} &
    OSEDiff*~\cite{wu2024osediff}  \hspace{-4mm} &
    OSDFace (ours) \hspace{-4mm}
    \\
    \end{tabular}
    \end{adjustbox}

\vspace{-2mm}
\caption{More visual comparison of the synthetic CelebA-Test dataset in challenging cases. Please zoom in for a better view.}
\label{fig:vis-supp-celeba-2}
\vspace{-2mm}
\end{figure*}

\begin{figure*}[t]
\scriptsize
\centering
\newcommand{\widthscale}{0.169}
\newcommand{\dataset}{CelebA}
\newcommand{\leftspace}{-8mm}
\newcommand{\imgid}{00000323}
\newcommand{\imgnote}{0323}

\hspace{\leftspace}
\begin{adjustbox}{valign=t}
\begin{tabular}{cccccc}
\includegraphics[width=\widthscale\textwidth]{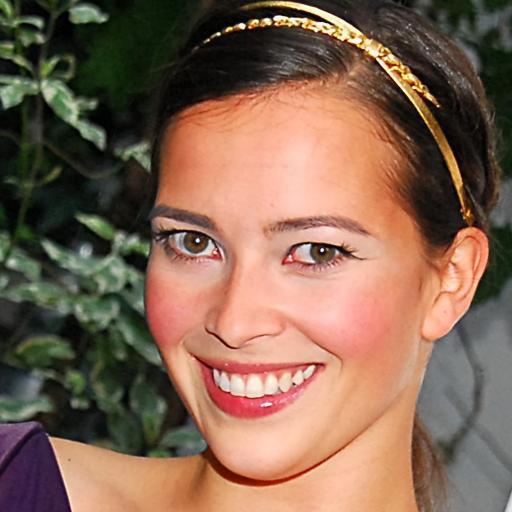} \hspace{-4mm} &
\includegraphics[width=\widthscale\textwidth]{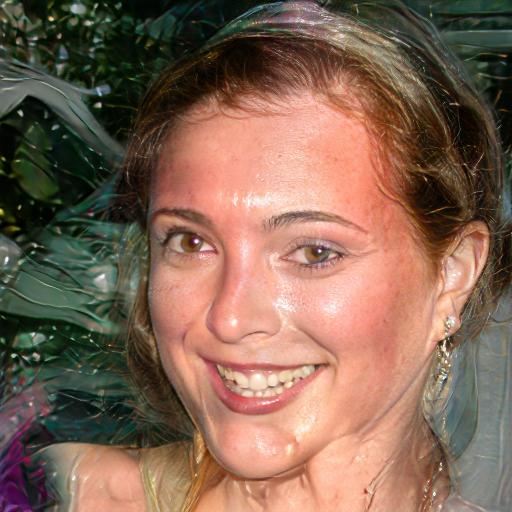} \hspace{-4mm} &
\includegraphics[width=\widthscale\textwidth]{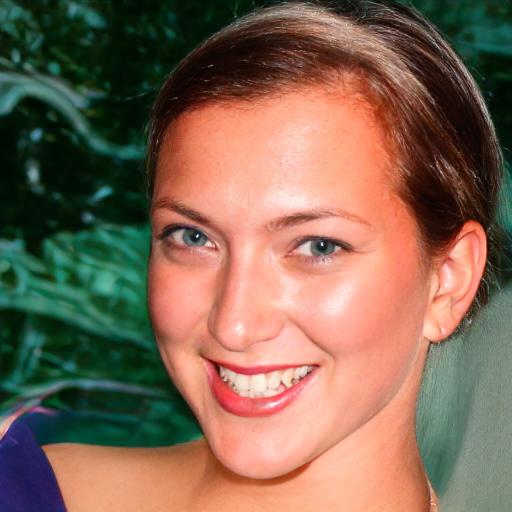} \hspace{-4mm} &
\includegraphics[width=\widthscale\textwidth]{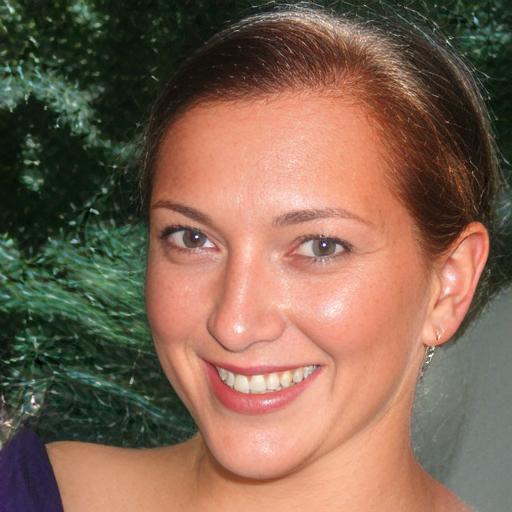} \hspace{-4mm} &
\includegraphics[width=\widthscale\textwidth]{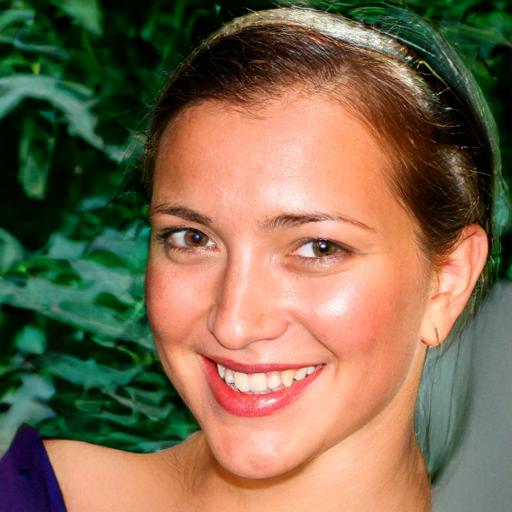} \hspace{-4mm} &
\includegraphics[width=\widthscale\textwidth]{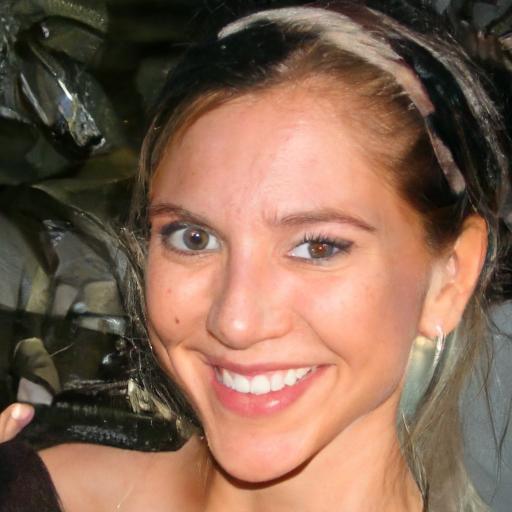} \hspace{-4mm} 
\\
HQ - \dataset~\imgnote \hspace{-4mm} &
RestoreFormer++~\cite{wang2023restoreformer++} \hspace{-4mm} &
VQFR~\cite{gu2022vqfr} \hspace{-4mm} &
CodeFormer~\cite{zhou2022codeformer} \hspace{-4mm} &
DAEFR~\cite{tsai2024daefr} \hspace{-4mm} &
PGDiff~\cite{yang2023pgdiff} \hspace{-4mm}
\\
\includegraphics[width=\widthscale\textwidth]{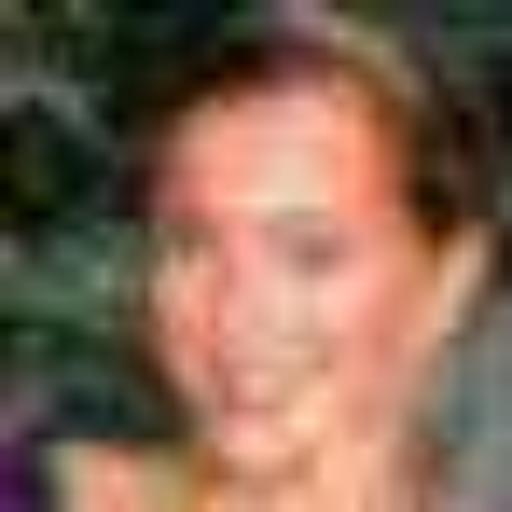} \hspace{-4mm} &
\includegraphics[width=\widthscale\textwidth]{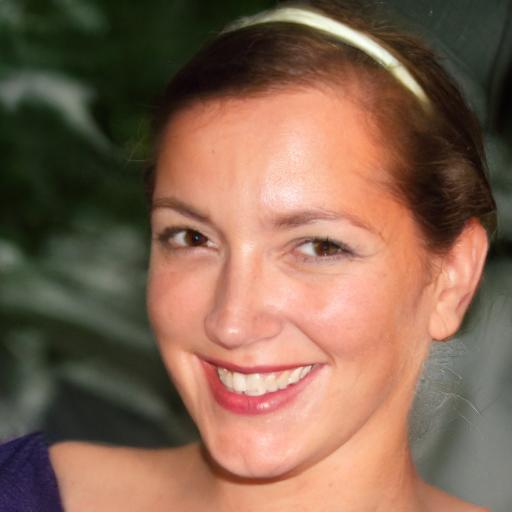} \hspace{-4mm} &
\includegraphics[width=\widthscale\textwidth]{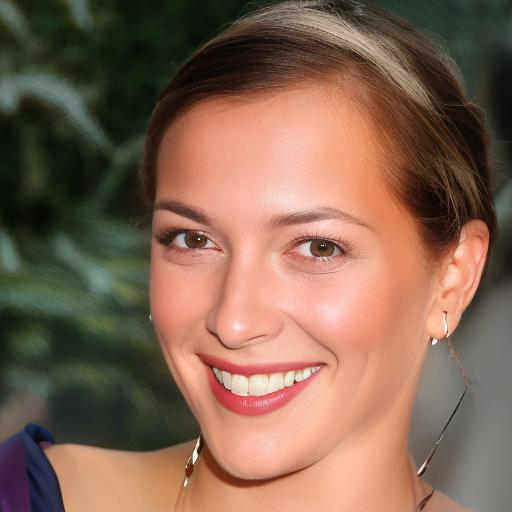} \hspace{-4mm} &
\includegraphics[width=\widthscale\textwidth]{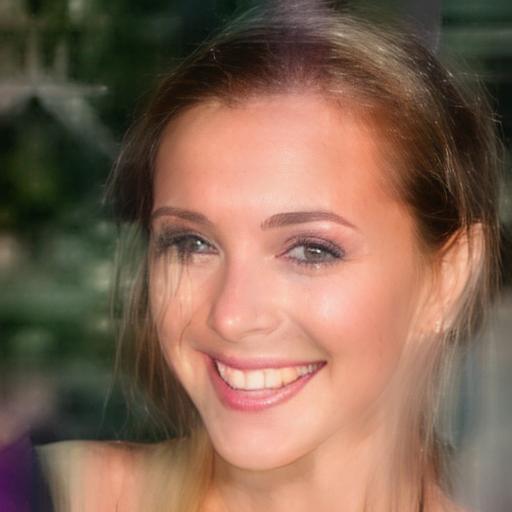} \hspace{-4mm} &
\includegraphics[width=\widthscale\textwidth]{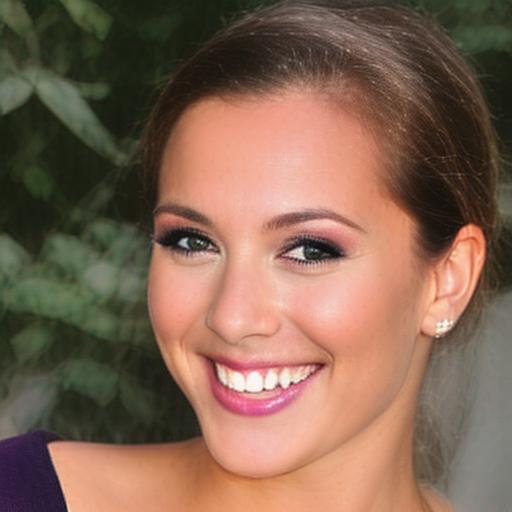} \hspace{-4mm} &
\includegraphics[width=\widthscale\textwidth]{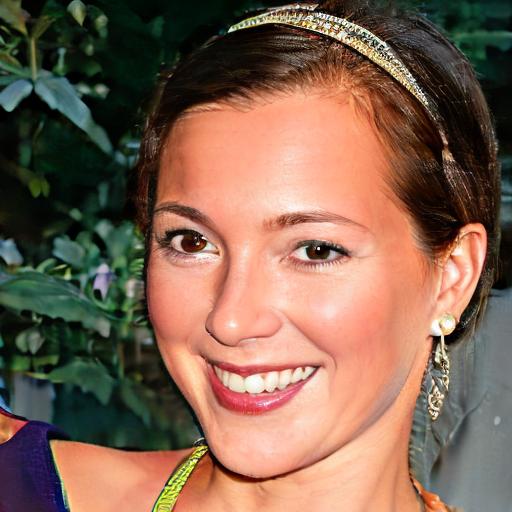} \hspace{-4mm} 
\\ 
LQ - \dataset~\imgnote \hspace{-4mm} &
DifFace~\cite{yue2024difface} \hspace{-4mm} &
DiffBIR~\cite{lin2024diffbir} \hspace{-4mm} &
OSEDiff~\cite{wu2024osediff}  \hspace{-4mm} &
OSEDiff*~\cite{wu2024osediff}  \hspace{-4mm} &
OSDFace (ours) \hspace{-4mm}
\\
\end{tabular}
\end{adjustbox}

\renewcommand{\imgid}{00000448}
\renewcommand{\imgnote}{0448}

\hspace{\leftspace}
\begin{adjustbox}{valign=t}
\begin{tabular}{cccccc}
\includegraphics[width=\widthscale\textwidth]{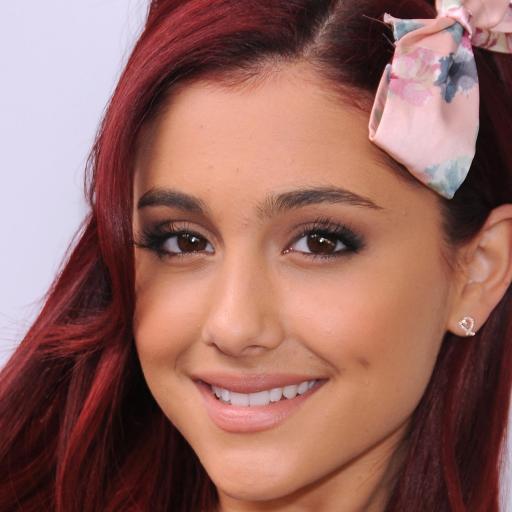} \hspace{-4mm} &
\includegraphics[width=\widthscale\textwidth]{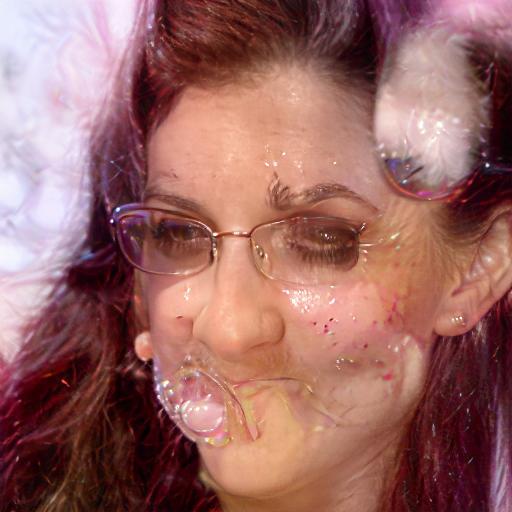} \hspace{-4mm} &
\includegraphics[width=\widthscale\textwidth]{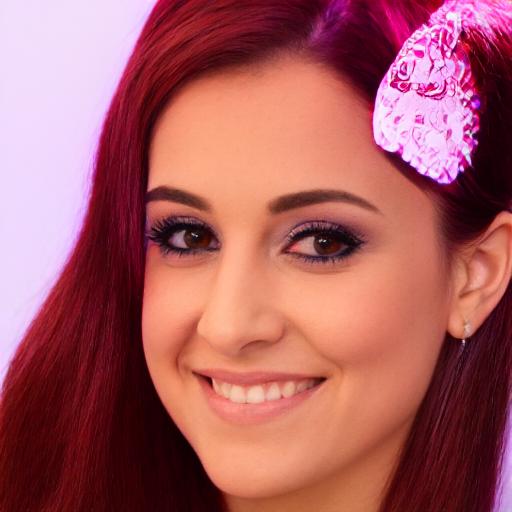} \hspace{-4mm} &
\includegraphics[width=\widthscale\textwidth]{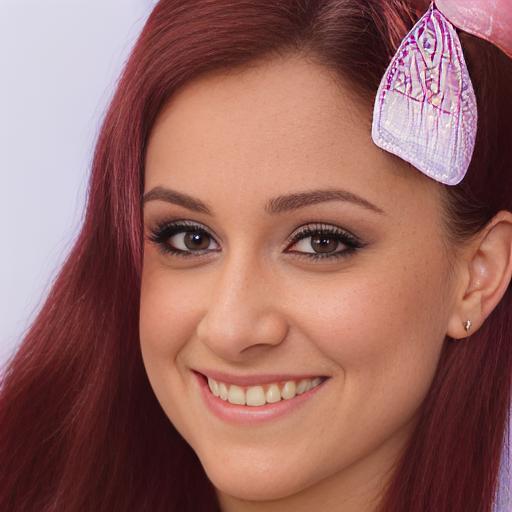} \hspace{-4mm} &
\includegraphics[width=\widthscale\textwidth]{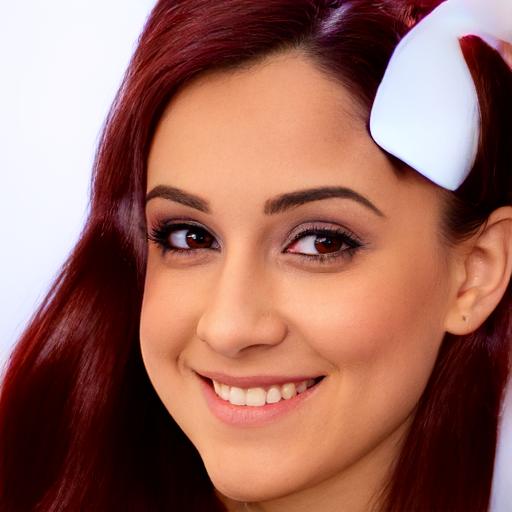} \hspace{-4mm} &
\includegraphics[width=\widthscale\textwidth]{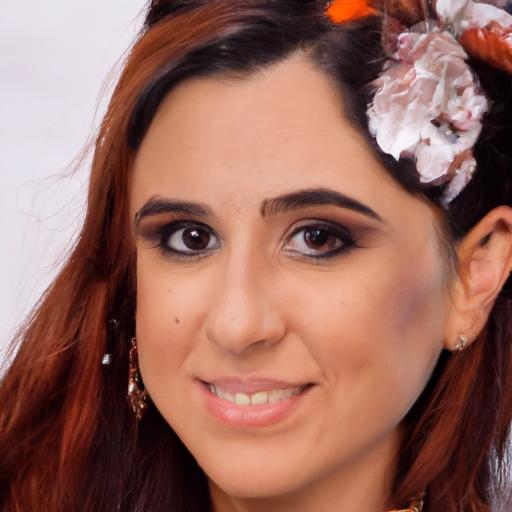} \hspace{-4mm} 
\\
HQ - \dataset~\imgnote \hspace{-4mm} &
RestoreFormer++~\cite{wang2023restoreformer++} \hspace{-4mm} &
VQFR~\cite{gu2022vqfr} \hspace{-4mm} &
CodeFormer~\cite{zhou2022codeformer} \hspace{-4mm} &
DAEFR~\cite{tsai2024daefr} \hspace{-4mm} &
PGDiff~\cite{yang2023pgdiff} \hspace{-4mm}
\\
\includegraphics[width=\widthscale\textwidth]{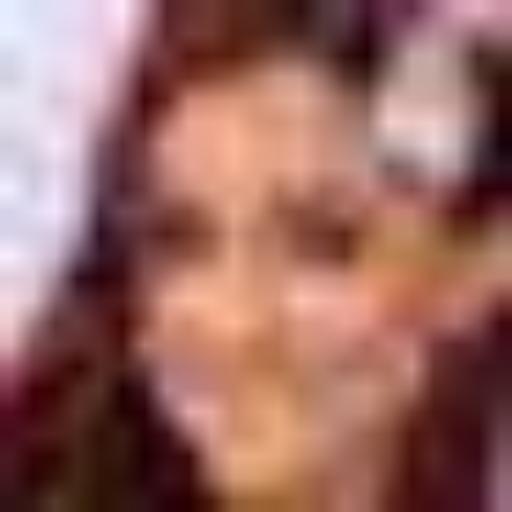} \hspace{-4mm} &
\includegraphics[width=\widthscale\textwidth]{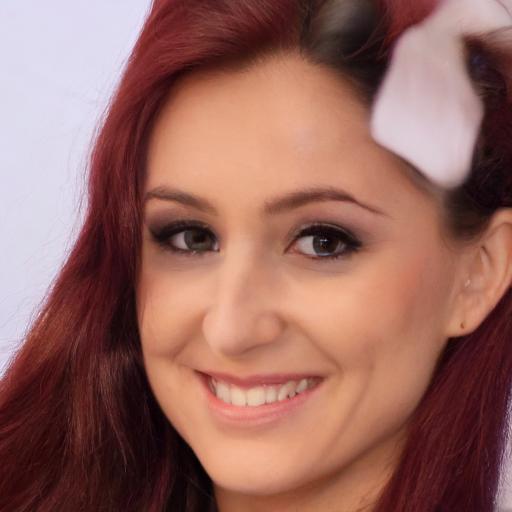} \hspace{-4mm} &
\includegraphics[width=\widthscale\textwidth]{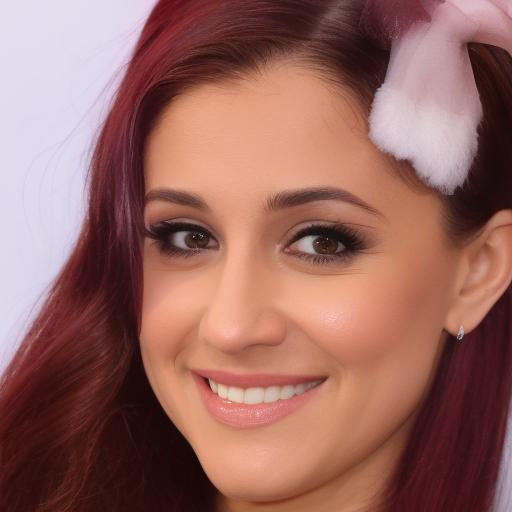} \hspace{-4mm} &
\includegraphics[width=\widthscale\textwidth]{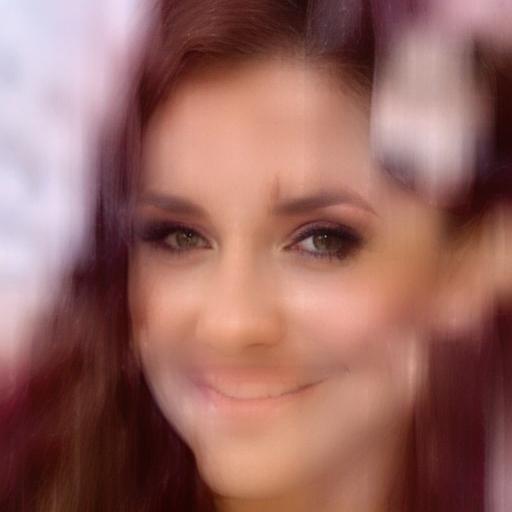} \hspace{-4mm} &
\includegraphics[width=\widthscale\textwidth]{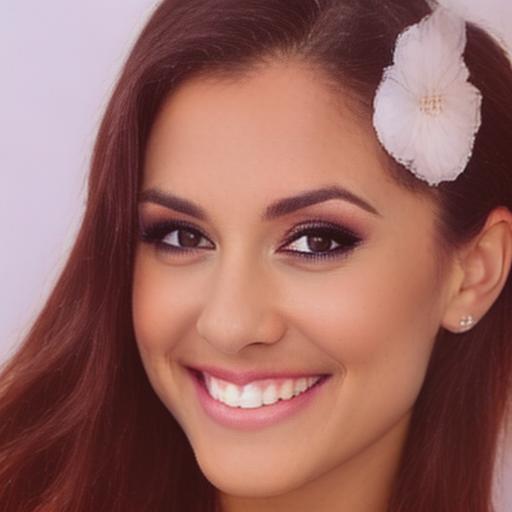} \hspace{-4mm} &
\includegraphics[width=\widthscale\textwidth]{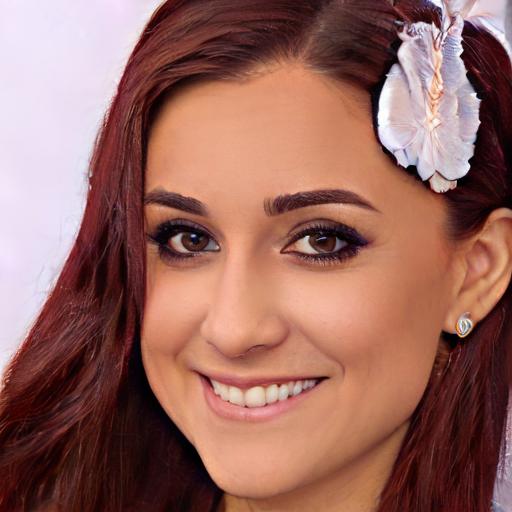} \hspace{-4mm} 
\\ 
LQ - \dataset~\imgnote \hspace{-4mm} &
DifFace~\cite{yue2024difface} \hspace{-4mm} &
DiffBIR~\cite{lin2024diffbir} \hspace{-4mm} &
OSEDiff~\cite{wu2024osediff}  \hspace{-4mm} &
OSEDiff*~\cite{wu2024osediff}  \hspace{-4mm} &
OSDFace (ours) \hspace{-4mm}
\\
\end{tabular}
\end{adjustbox}

\renewcommand{\imgid}{00000509}
\renewcommand{\imgnote}{0509}

\hspace{\leftspace}
\begin{adjustbox}{valign=t}
\begin{tabular}{cccccc}
\includegraphics[width=\widthscale\textwidth]{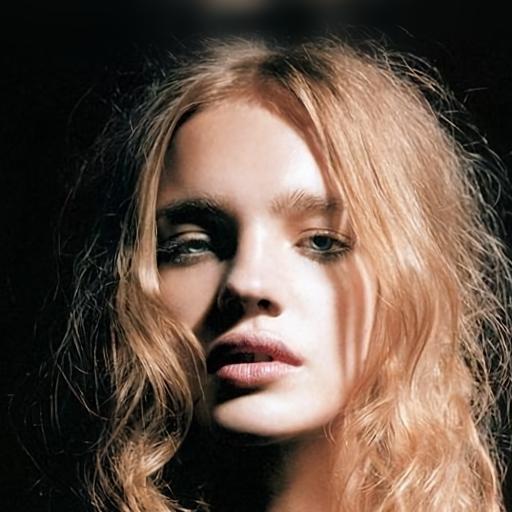} \hspace{-4mm} &
\includegraphics[width=\widthscale\textwidth]{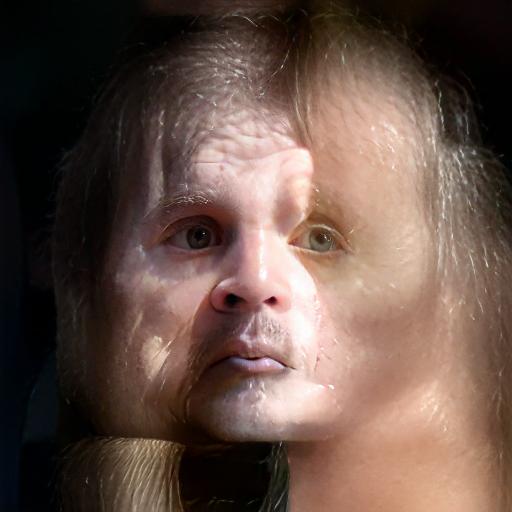} \hspace{-4mm} &
\includegraphics[width=\widthscale\textwidth]{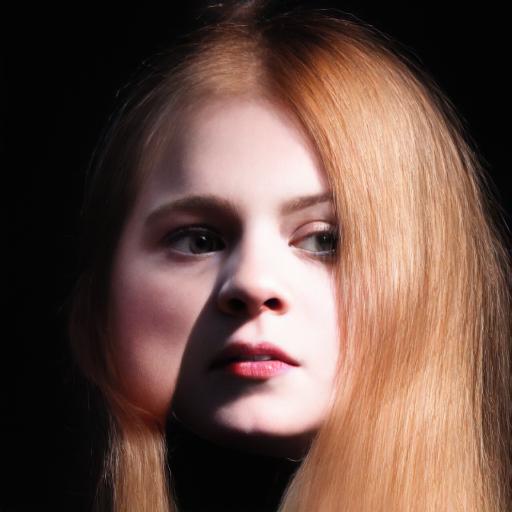} \hspace{-4mm} &
\includegraphics[width=\widthscale\textwidth]{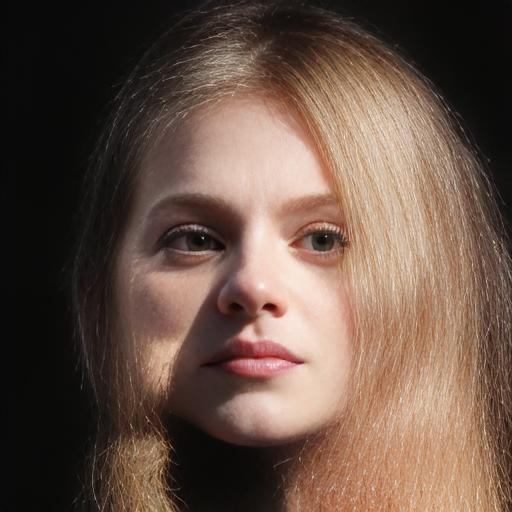} \hspace{-4mm} &
\includegraphics[width=\widthscale\textwidth]{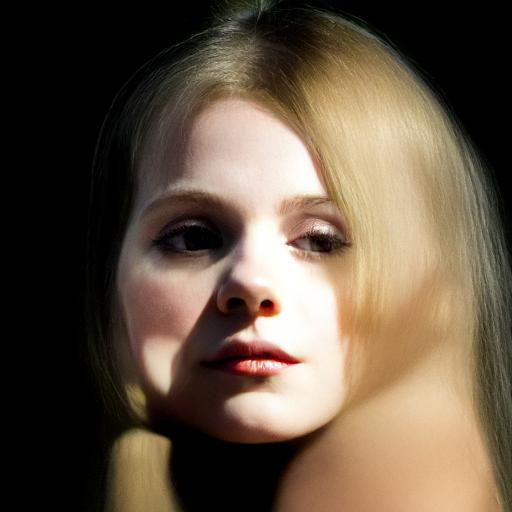} \hspace{-4mm} &
\includegraphics[width=\widthscale\textwidth]{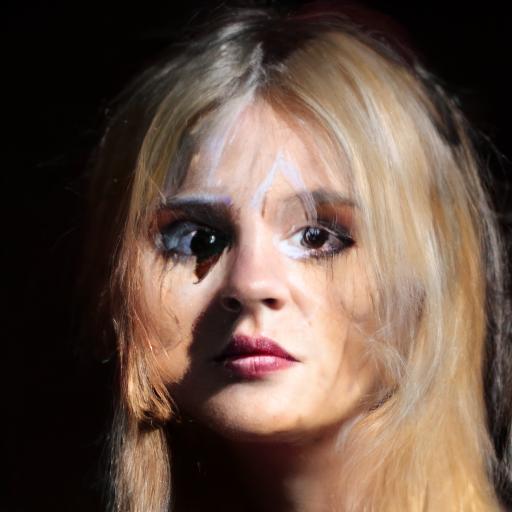} \hspace{-4mm} 
\\
HQ - \dataset~\imgnote \hspace{-4mm} &
RestoreFormer++~\cite{wang2023restoreformer++} \hspace{-4mm} &
VQFR~\cite{gu2022vqfr} \hspace{-4mm} &
CodeFormer~\cite{zhou2022codeformer} \hspace{-4mm} &
DAEFR~\cite{tsai2024daefr} \hspace{-4mm} &
PGDiff~\cite{yang2023pgdiff} \hspace{-4mm}
\\
\includegraphics[width=\widthscale\textwidth]{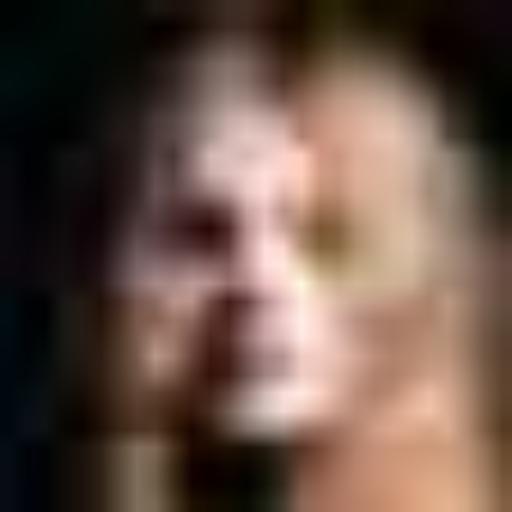} \hspace{-4mm} &
\includegraphics[width=\widthscale\textwidth]{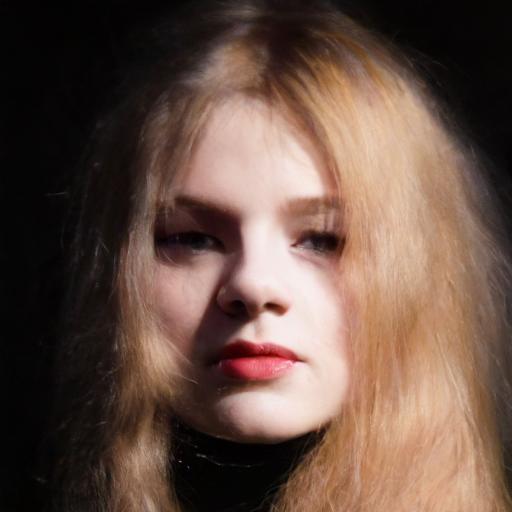} \hspace{-4mm} &
\includegraphics[width=\widthscale\textwidth]{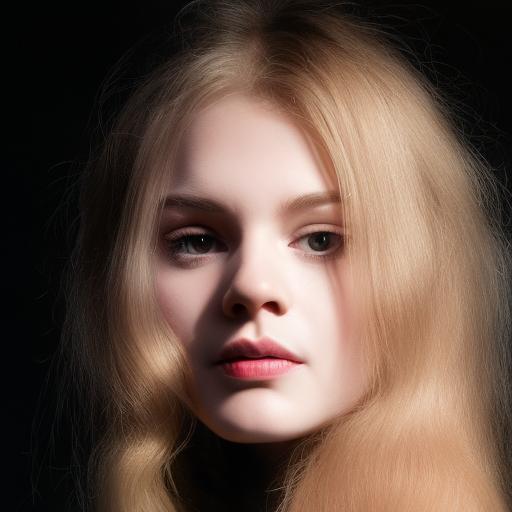} \hspace{-4mm} &
\includegraphics[width=\widthscale\textwidth]{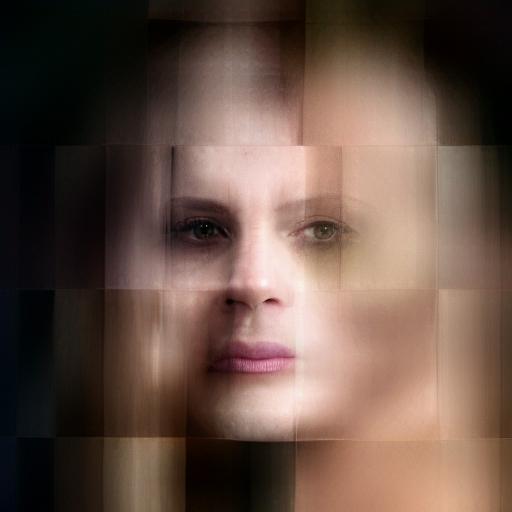} \hspace{-4mm} &
\includegraphics[width=\widthscale\textwidth]{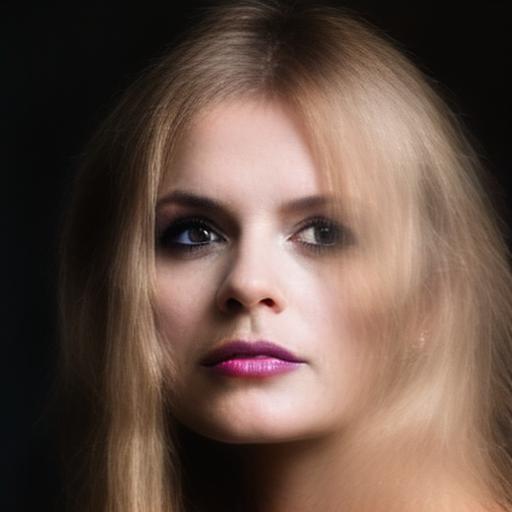} \hspace{-4mm} &
\includegraphics[width=\widthscale\textwidth]{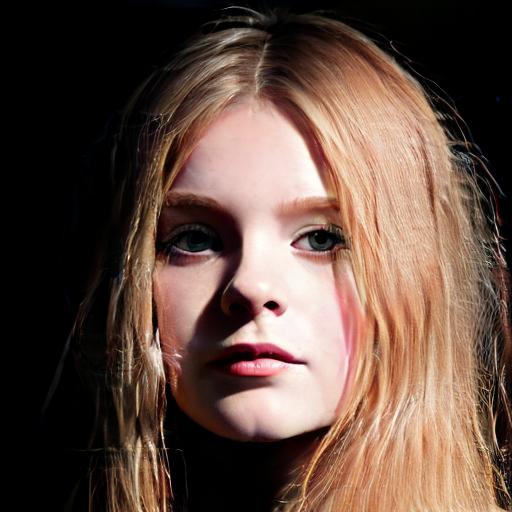} \hspace{-4mm} 
\\ 
LQ - \dataset~\imgnote \hspace{-4mm} &
DifFace~\cite{yue2024difface} \hspace{-4mm} &
DiffBIR~\cite{lin2024diffbir} \hspace{-4mm} &
OSEDiff~\cite{wu2024osediff}  \hspace{-4mm} &
OSEDiff*~\cite{wu2024osediff}  \hspace{-4mm} &
OSDFace (ours) \hspace{-4mm}
\\
\end{tabular}
\end{adjustbox}

\vspace{-2mm}
\caption{More visual comparison of the synthetic CelebA-Test dataset in challenging cases. Please zoom in for a better view.}
\label{fig:vis-supp-celeba-3}
\vspace{-2mm}
\end{figure*}

\noindent\textbf{Synthetic dataset.} Visualized results are presented in Fig.~\ref{fig:vis-supp-celeba-1}, Fig.~\ref{fig:vis-supp-celeba-2}, and Fig.~\ref{fig:vis-supp-celeba-3}. Compared to other methods, OSDFace produces more natural-looking restorations with greater detail. This is especially evident in the hair, whether long, short, straight, or curly. Additionally, our method effectively restores occluded regions, such as an arm covering the mouth or bangs obscuring the eyes. For profile views, OSDFace naturally recovers facial contours. In some ground truth images with blurred backgrounds, OSDFace performs well, even achieving higher quality and greater detail than the original HQ images. In scenarios with complex backgrounds, many VQ-based methods, such as VQFR~\cite{gu2022vqfr}, CodeFormer~\cite{zhou2022codeformer}, and DAEFR~\cite{tsai2024daefr}, fail to restore natural backgrounds. These methods often produce wallpaper-like outputs, exhibit color distortions, or even blend the person's clothing with the background. In contrast, OSDFace, which combines VQ Dict and diffusion model, successfully generates harmonious faces.

\begin{figure*}[t]
\scriptsize
\centering
\newcommand{\widthscale}{0.189}
\newcommand{\dataset}{Wider}
\newcommand{\imgid}{0003}
\newcommand{\imgnote}{0003}
\scalebox{0.98}{
    \hspace{-0.4cm}
    \begin{adjustbox}{valign=t}
    \begin{tabular}{ccccc}
    \includegraphics[width=\widthscale\textwidth]{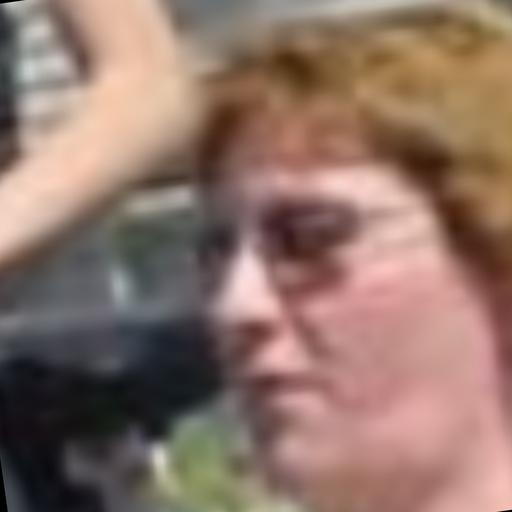} \hspace{-4mm} &
    \includegraphics[width=\widthscale\textwidth]{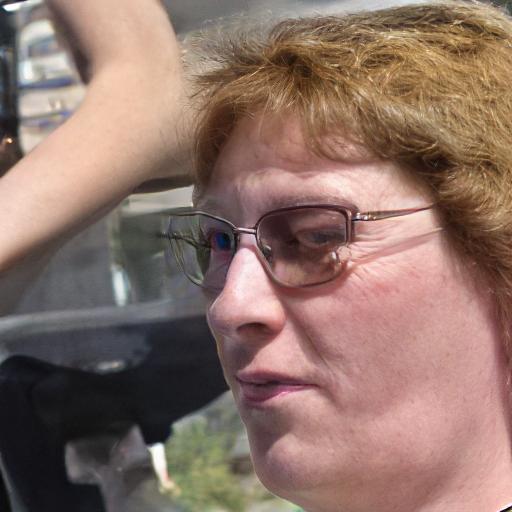} \hspace{-4mm} &
    \includegraphics[width=\widthscale\textwidth]{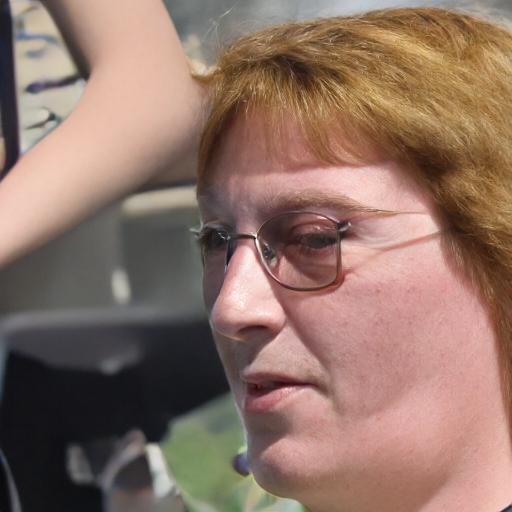} \hspace{-4mm} &
    \includegraphics[width=\widthscale\textwidth]{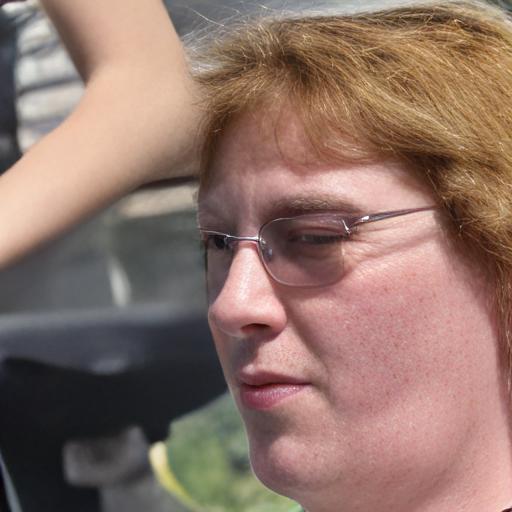} \hspace{-4mm} &
    \includegraphics[width=\widthscale\textwidth]{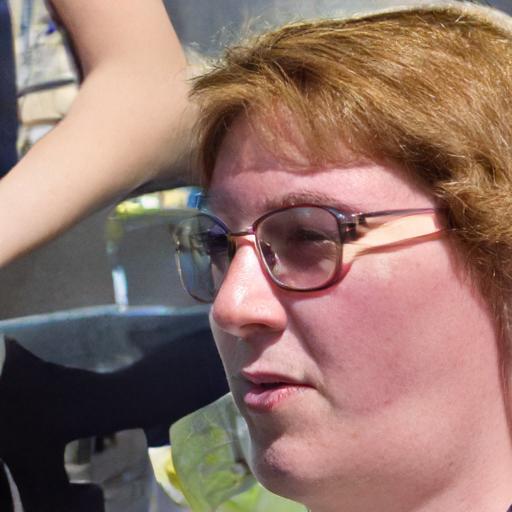} \hspace{-4mm} 
    \\
    \dataset~\imgnote \hspace{-4mm} &
    RestoreFormer++~\cite{wang2023restoreformer++} \hspace{-4mm} &
    VQFR~\cite{gu2022vqfr} \hspace{-4mm} &
    CodeFormer~\cite{zhou2022codeformer} \hspace{-4mm} &
    DAEFR~\cite{tsai2024daefr} \hspace{-4mm} 
    \\
    \includegraphics[width=\widthscale\textwidth]{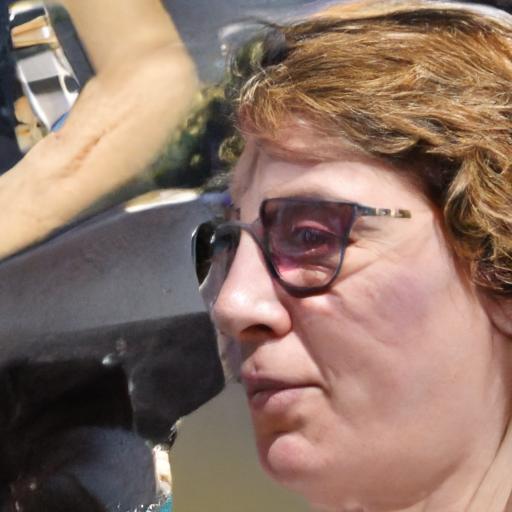} \hspace{-4mm} &
    \includegraphics[width=\widthscale\textwidth]{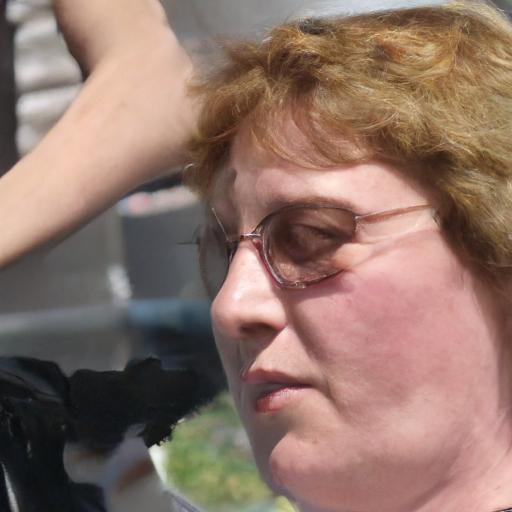} \hspace{-4mm} &
    \includegraphics[width=\widthscale\textwidth]{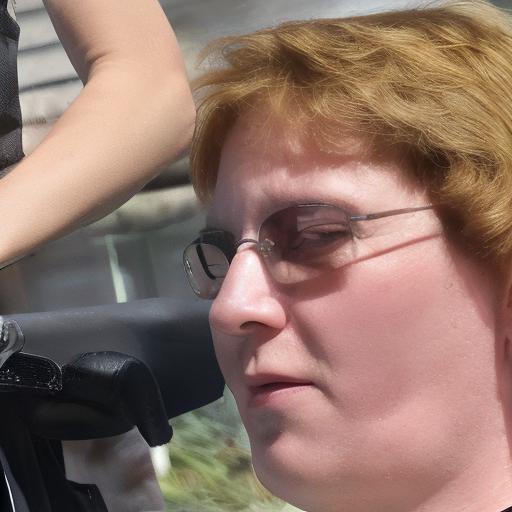} \hspace{-4mm} &
    \includegraphics[width=\widthscale\textwidth]{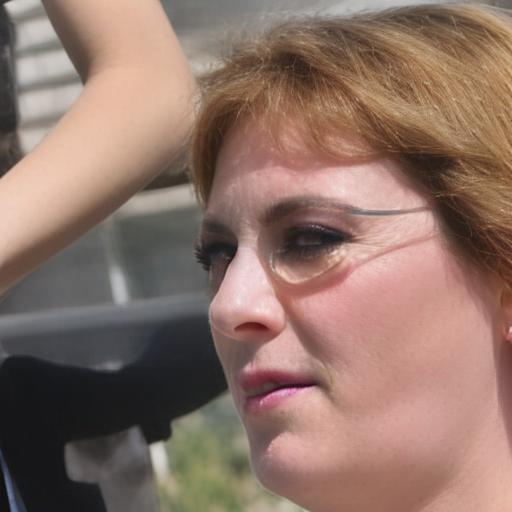} \hspace{-4mm} &
    \includegraphics[width=\widthscale\textwidth]{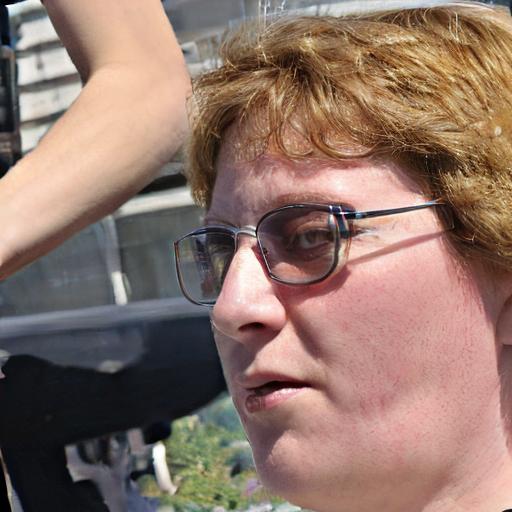} \hspace{-4mm} 
    \\ 
    PGDiff~\cite{yang2023pgdiff} \hspace{-4mm} &
    DifFace~\cite{yue2024difface} \hspace{-4mm} &
    DiffBIR~\cite{lin2024diffbir} \hspace{-4mm} &
    OSEDiff*~\cite{wu2024osediff}  \hspace{-4mm} &
    OSDFace (ours) \hspace{-4mm}
    \\
    \end{tabular}
    \end{adjustbox}
}
\renewcommand{\imgid}{0026}
\renewcommand{\imgnote}{0026}
\scalebox{0.98}{
    \hspace{-0.4cm}
    \begin{adjustbox}{valign=t}
    \begin{tabular}{ccccc}
    \includegraphics[width=\widthscale\textwidth]{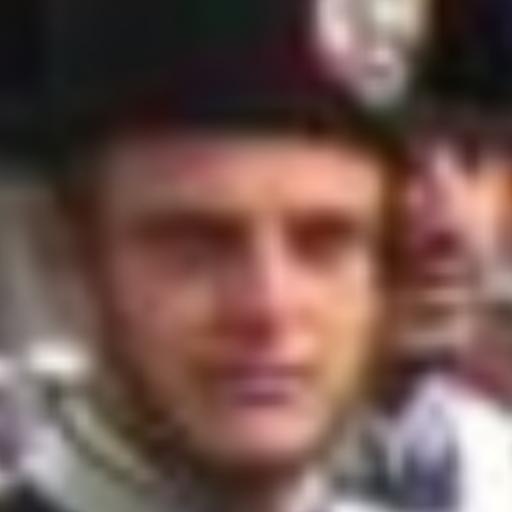} \hspace{-4mm} &
    \includegraphics[width=\widthscale\textwidth]{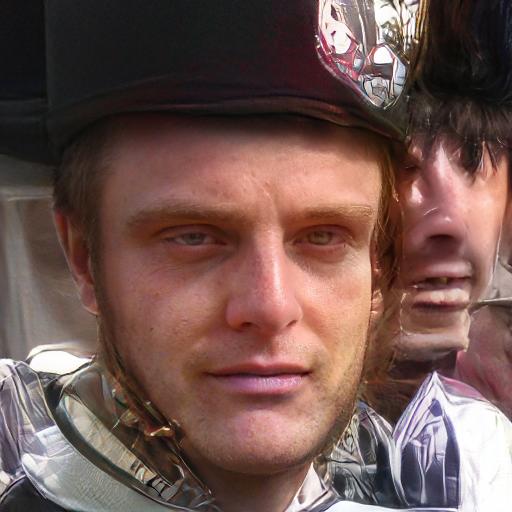} \hspace{-4mm} &
    \includegraphics[width=\widthscale\textwidth]{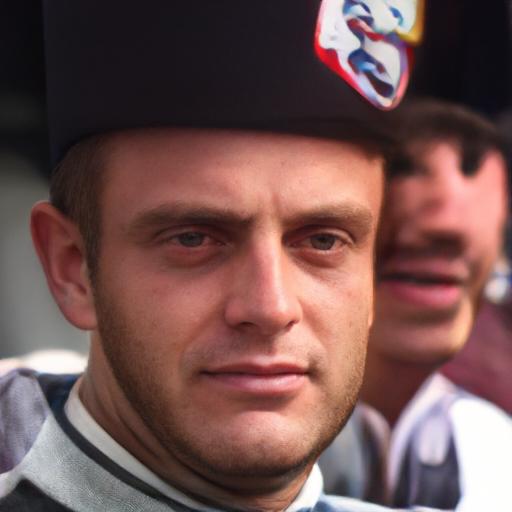} \hspace{-4mm} &
    \includegraphics[width=\widthscale\textwidth]{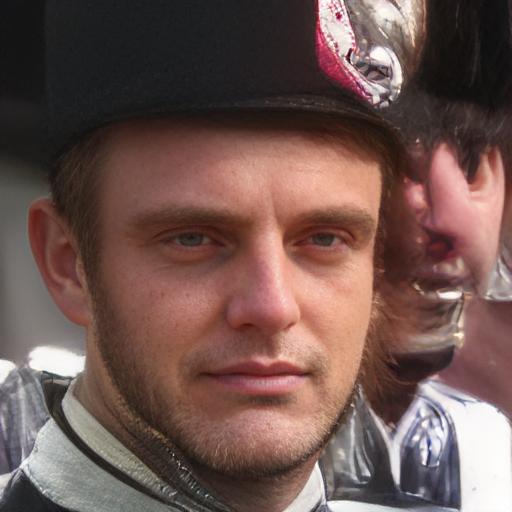} \hspace{-4mm} &
    \includegraphics[width=\widthscale\textwidth]{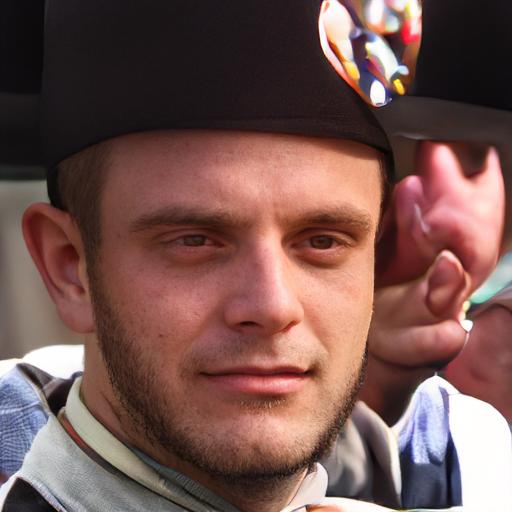} \hspace{-4mm} 
    \\
    \dataset~\imgnote \hspace{-4mm} &
    RestoreFormer++~\cite{wang2023restoreformer++} \hspace{-4mm} &
    VQFR~\cite{gu2022vqfr} \hspace{-4mm} &
    CodeFormer~\cite{zhou2022codeformer} \hspace{-4mm} &
    DAEFR~\cite{tsai2024daefr} \hspace{-4mm} 
    \\
    \includegraphics[width=\widthscale\textwidth]{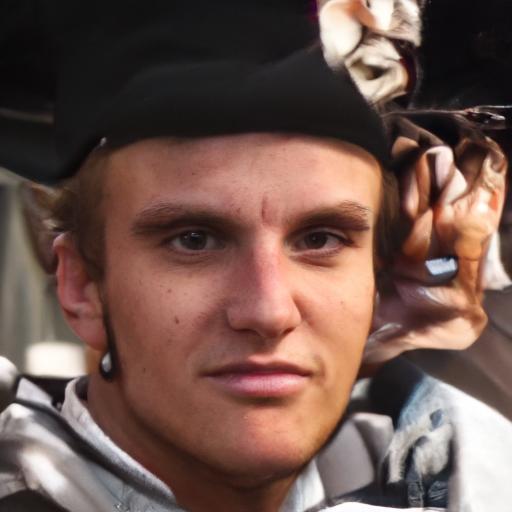} \hspace{-4mm} &
    \includegraphics[width=\widthscale\textwidth]{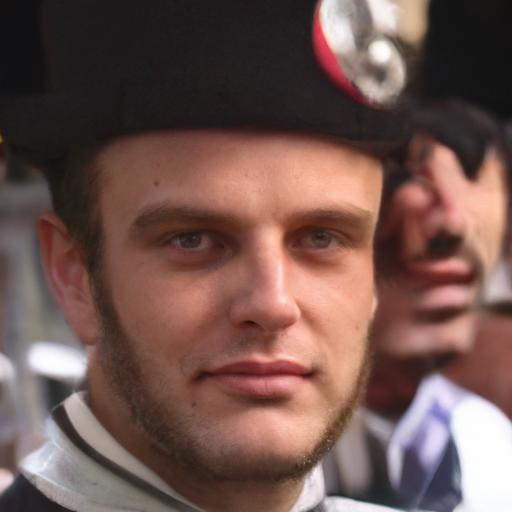} \hspace{-4mm} &
    \includegraphics[width=\widthscale\textwidth]{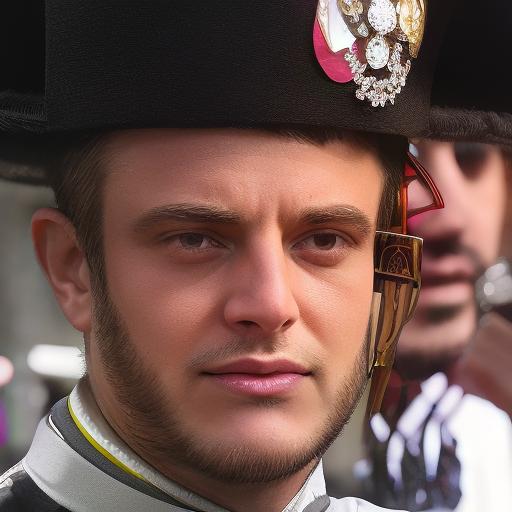} \hspace{-4mm} &
    \includegraphics[width=\widthscale\textwidth]{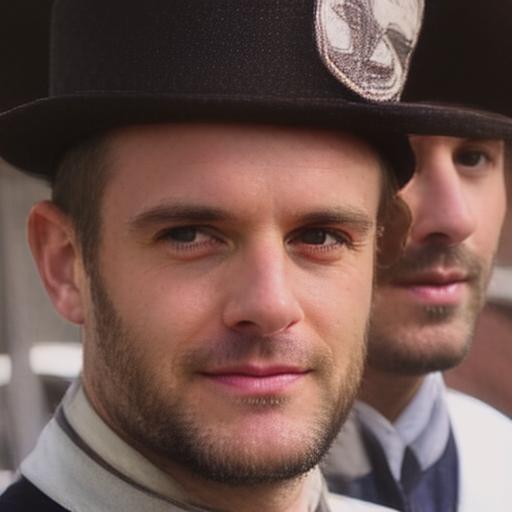} \hspace{-4mm} &
    \includegraphics[width=\widthscale\textwidth]{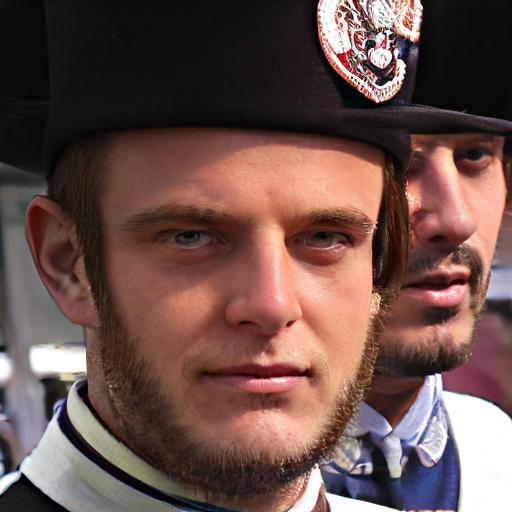} \hspace{-4mm} 
    \\ 
    PGDiff~\cite{yang2023pgdiff} \hspace{-4mm} &
    DifFace~\cite{yue2024difface} \hspace{-4mm} &
    DiffBIR~\cite{lin2024diffbir} \hspace{-4mm} &
    OSEDiff*~\cite{wu2024osediff}  \hspace{-4mm} &
    OSDFace (ours) \hspace{-4mm}
    \\
    \end{tabular}
    \end{adjustbox}
}
\renewcommand{\imgid}{0047}
\renewcommand{\imgnote}{0047}
\scalebox{0.98}{
    \hspace{-0.4cm}
    \begin{adjustbox}{valign=t}
    \begin{tabular}{ccccc}
    \includegraphics[width=\widthscale\textwidth]{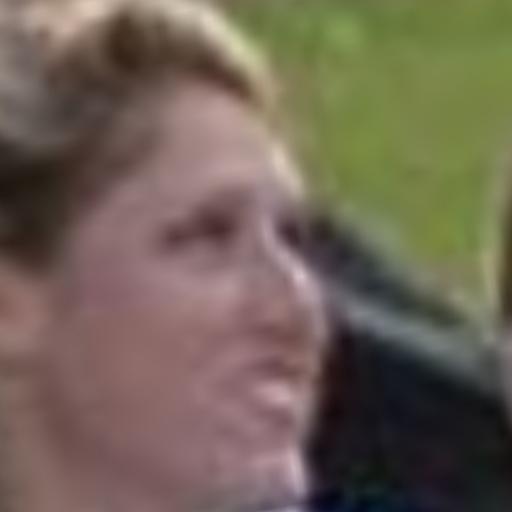} \hspace{-4mm} &
    \includegraphics[width=\widthscale\textwidth]{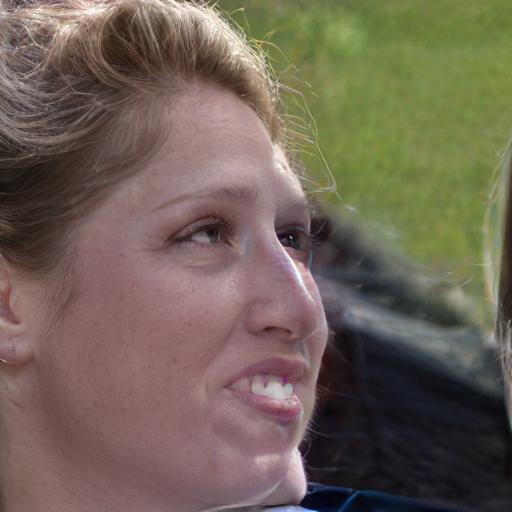} \hspace{-4mm} &
    \includegraphics[width=\widthscale\textwidth]{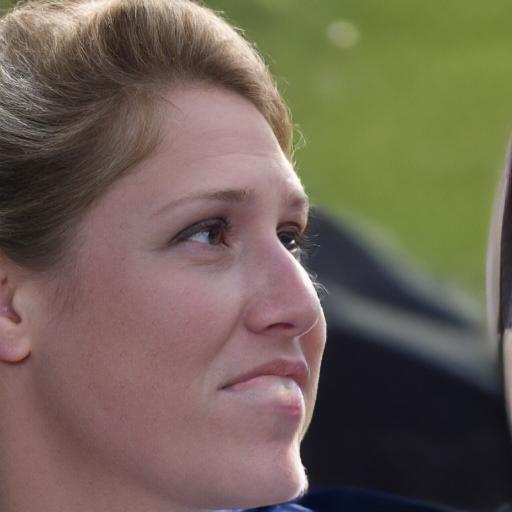} \hspace{-4mm} &
    \includegraphics[width=\widthscale\textwidth]{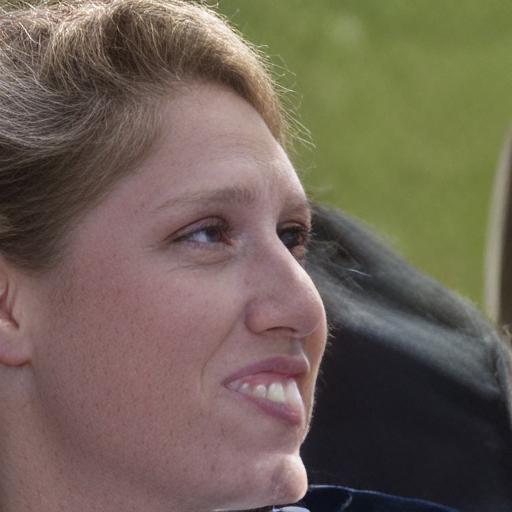} \hspace{-4mm} &
    \includegraphics[width=\widthscale\textwidth]{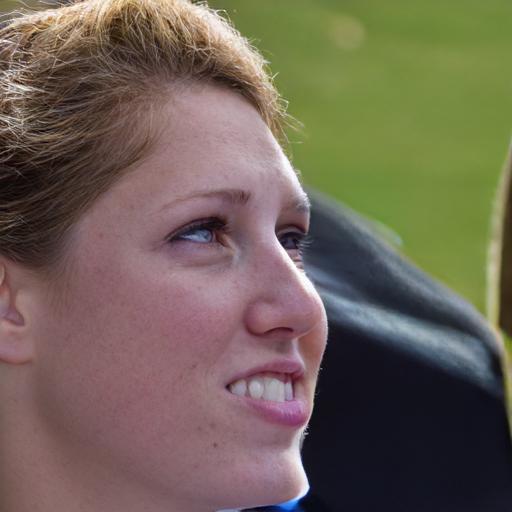} \hspace{-4mm} 
    \\
    \dataset~\imgnote \hspace{-4mm} &
    RestoreFormer++~\cite{wang2023restoreformer++} \hspace{-4mm} &
    VQFR~\cite{gu2022vqfr} \hspace{-4mm} &
    CodeFormer~\cite{zhou2022codeformer} \hspace{-4mm} &
    DAEFR~\cite{tsai2024daefr} \hspace{-4mm} 
    \\
    \includegraphics[width=\widthscale\textwidth]{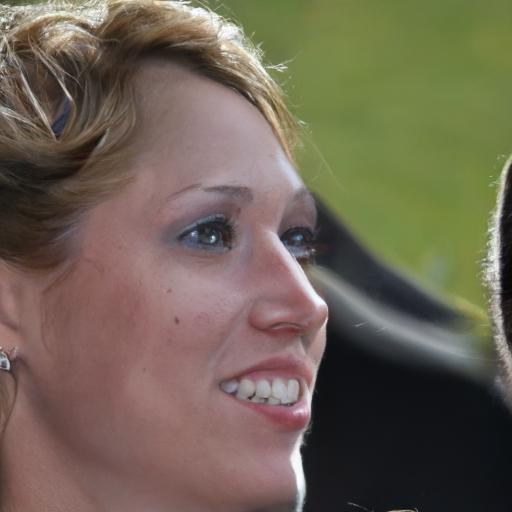} \hspace{-4mm} &
    \includegraphics[width=\widthscale\textwidth]{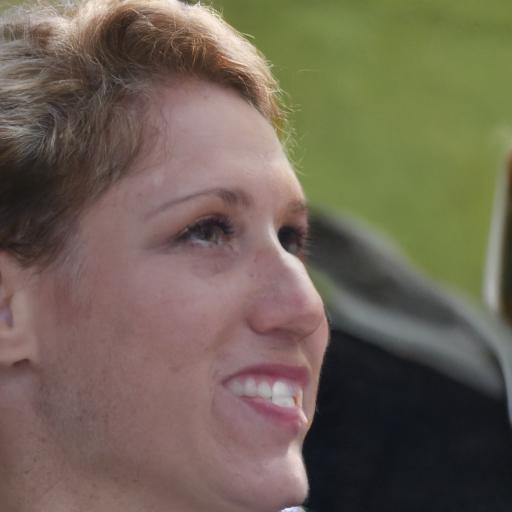} \hspace{-4mm} &
    \includegraphics[width=\widthscale\textwidth]{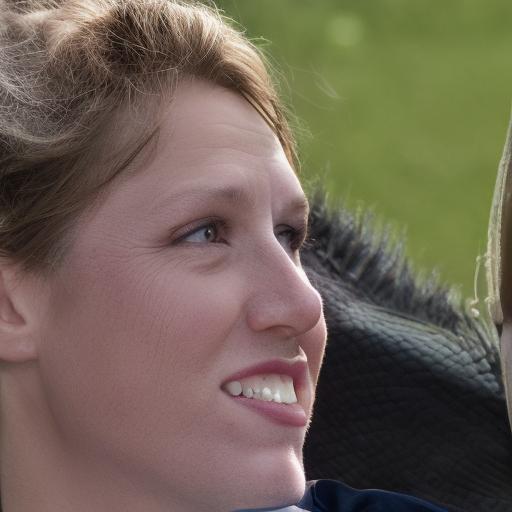} \hspace{-4mm} &
    \includegraphics[width=\widthscale\textwidth]{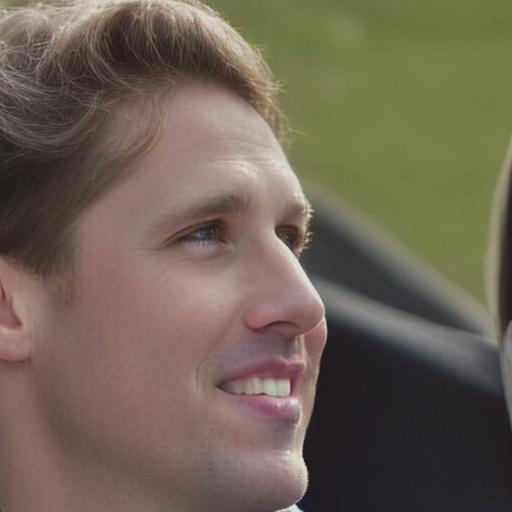} \hspace{-4mm} &
    \includegraphics[width=\widthscale\textwidth]{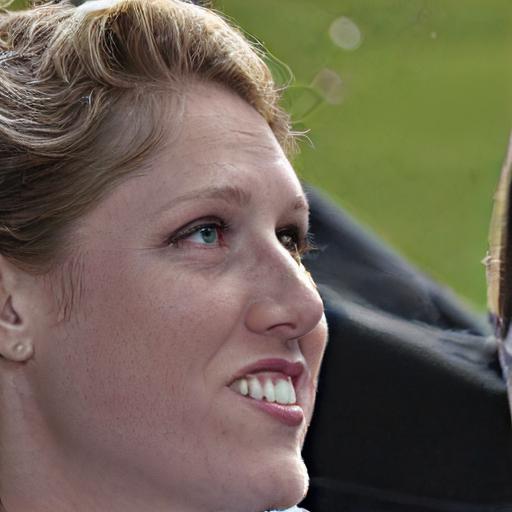} \hspace{-4mm} 
    \\ 
    PGDiff~\cite{yang2023pgdiff} \hspace{-4mm} &
    DifFace~\cite{yue2024difface} \hspace{-4mm} &
    DiffBIR~\cite{lin2024diffbir} \hspace{-4mm} &
    OSEDiff*~\cite{wu2024osediff}  \hspace{-4mm} &
    OSDFace (ours) \hspace{-4mm}
    \\
    \end{tabular}
    \end{adjustbox}
}
\vspace{-2mm}
\caption{More visual comparison of the real-world Wider-Test dataset in challenging cases. Please zoom in for a better view.}
\label{fig:vis-supp-wider}
\vspace{-2mm}
\end{figure*}

\begin{figure*}[t]
\scriptsize
\centering
\newcommand{\widthscale}{0.189}
\newcommand{\dataset}{LFW}
\newcommand{\imgid}{Anna_Faris_0001_00}
\newcommand{\imgnote}{Anna Faris}
\scalebox{0.98}{
    \hspace{-0.4cm}
    \begin{adjustbox}{valign=t}
    \begin{tabular}{ccccc}
    \includegraphics[width=\widthscale\textwidth]{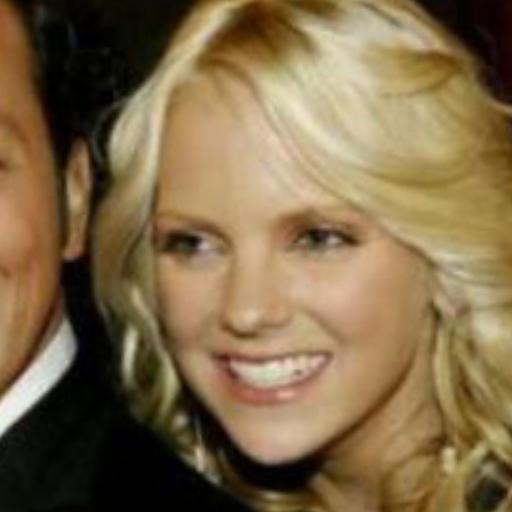} \hspace{-4mm} &
    \includegraphics[width=\widthscale\textwidth]{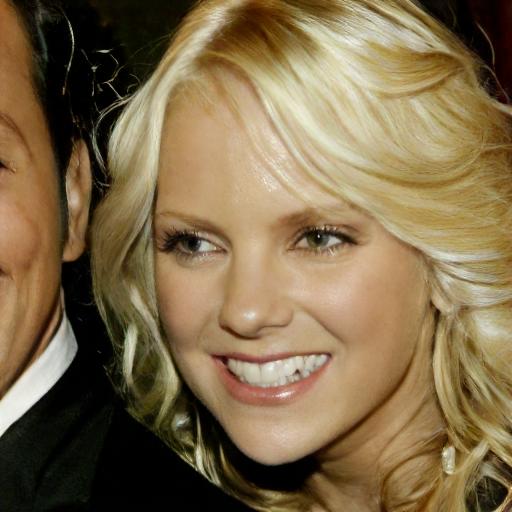} \hspace{-4mm} &
    \includegraphics[width=\widthscale\textwidth]{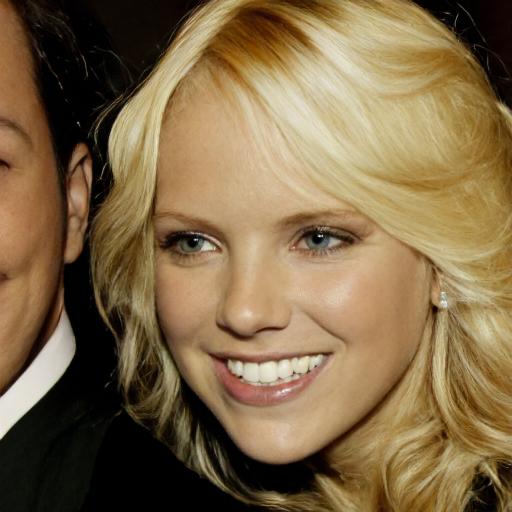} \hspace{-4mm} &
    \includegraphics[width=\widthscale\textwidth]{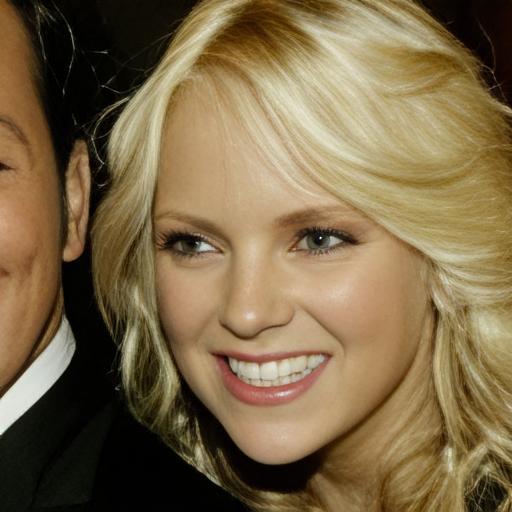} \hspace{-4mm} &
    \includegraphics[width=\widthscale\textwidth]{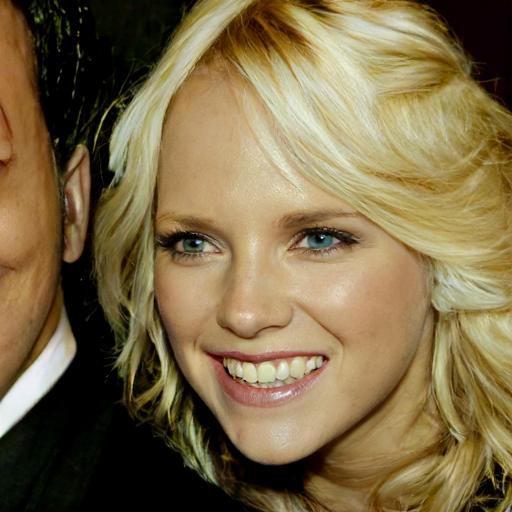} \hspace{-4mm} 
    \\
    \dataset~\imgnote \hspace{-4mm} &
    RestoreFormer++~\cite{wang2023restoreformer++} \hspace{-4mm} &
    VQFR~\cite{gu2022vqfr} \hspace{-4mm} &
    CodeFormer~\cite{zhou2022codeformer} \hspace{-4mm} &
    DAEFR~\cite{tsai2024daefr} \hspace{-4mm} 
    \\
    \includegraphics[width=\widthscale\textwidth]{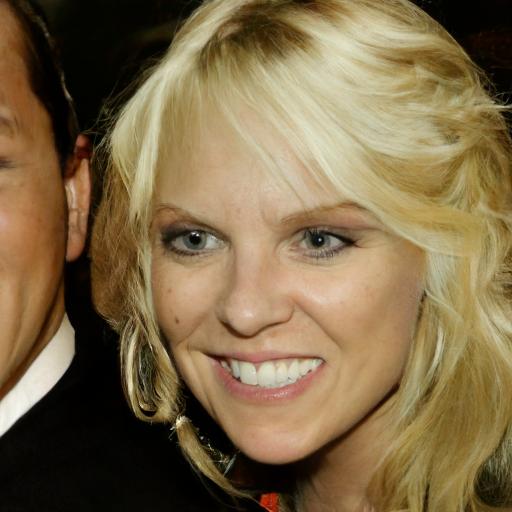} \hspace{-4mm} &
    \includegraphics[width=\widthscale\textwidth]{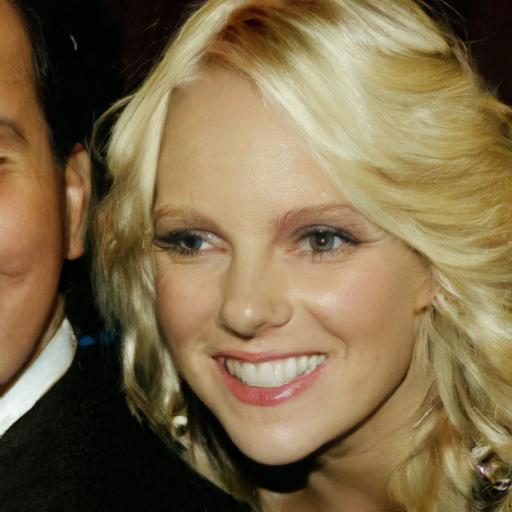} \hspace{-4mm} &
    \includegraphics[width=\widthscale\textwidth]{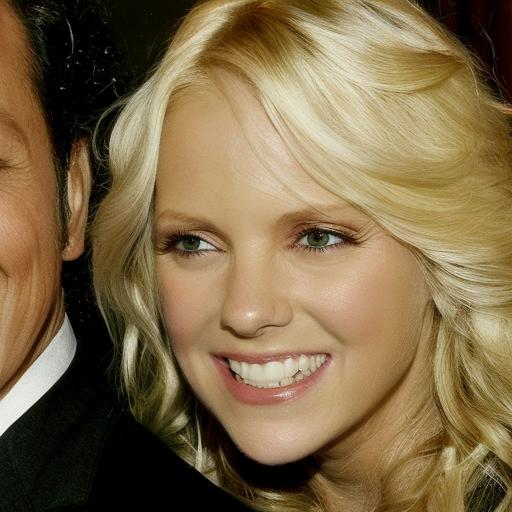} \hspace{-4mm} &
    \includegraphics[width=\widthscale\textwidth]{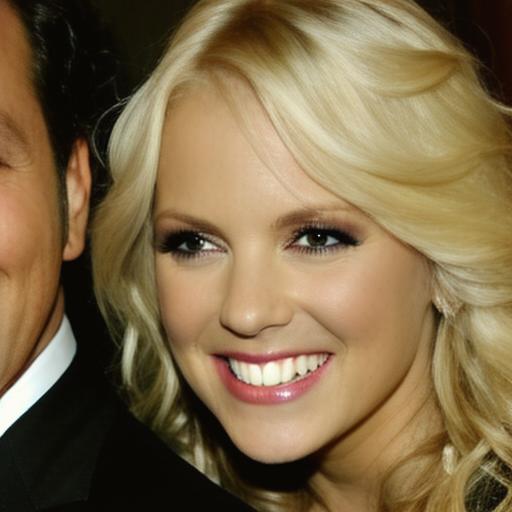} \hspace{-4mm} &
    \includegraphics[width=\widthscale\textwidth]{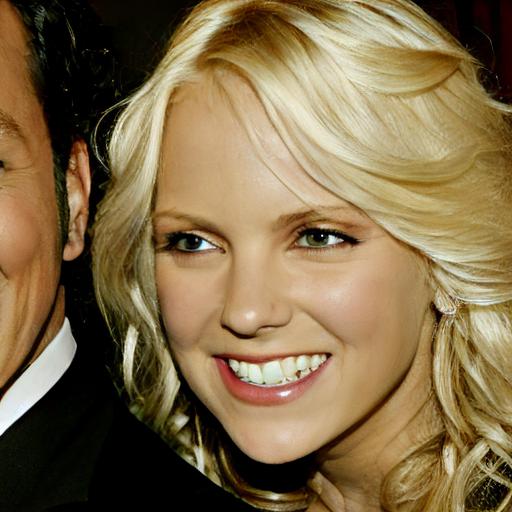} \hspace{-4mm} 
    \\ 
    PGDiff~\cite{yang2023pgdiff} \hspace{-4mm} &
    DifFace~\cite{yue2024difface} \hspace{-4mm} &
    DiffBIR~\cite{lin2024diffbir} \hspace{-4mm} &
    OSEDiff*~\cite{wu2024osediff}  \hspace{-4mm} &
    OSDFace (ours) \hspace{-4mm}
    \\
    \end{tabular}
    \end{adjustbox}
}
\renewcommand{\imgid}{Damon_Stoudamire_0001_00}
\renewcommand{\imgnote}{Damon Stoudamire}
\scalebox{0.98}{
    \hspace{-0.4cm}
    \begin{adjustbox}{valign=t}
    \begin{tabular}{ccccc}
    \includegraphics[width=\widthscale\textwidth]{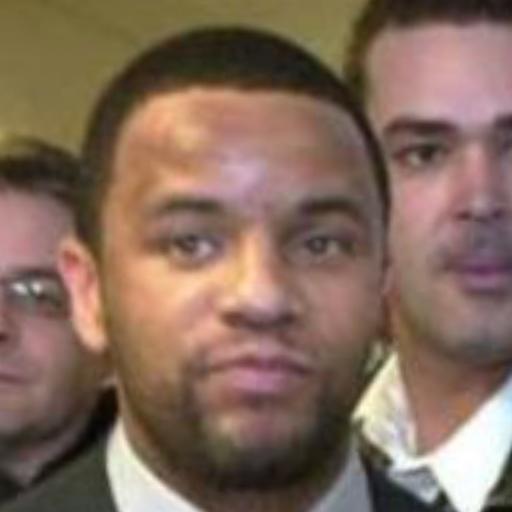} \hspace{-4mm} &
    \includegraphics[width=\widthscale\textwidth]{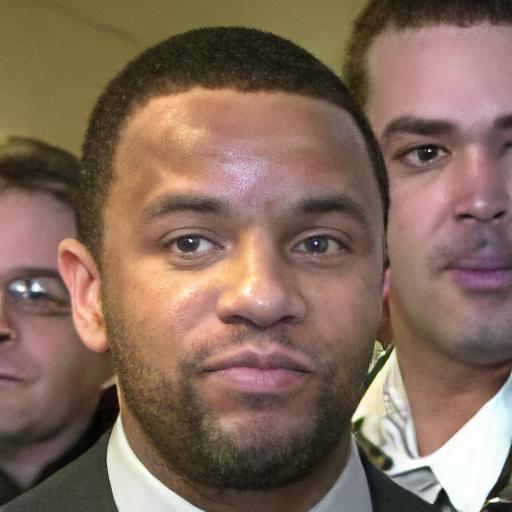} \hspace{-4mm} &
    \includegraphics[width=\widthscale\textwidth]{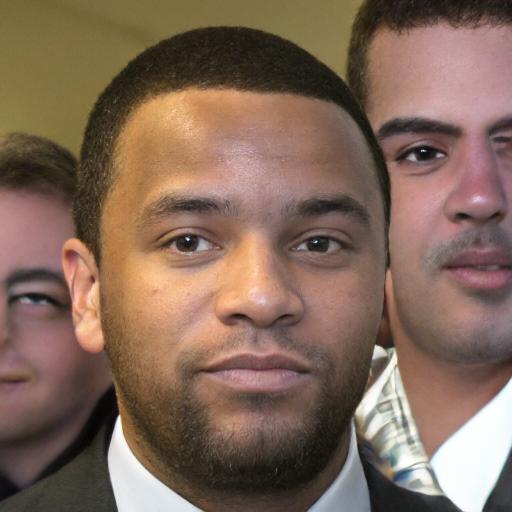} \hspace{-4mm} &
    \includegraphics[width=\widthscale\textwidth]{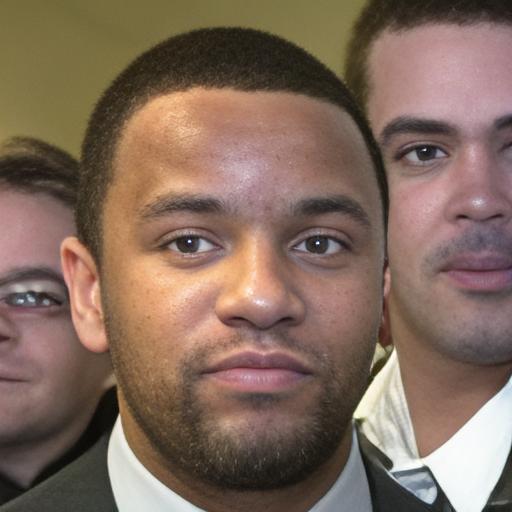} \hspace{-4mm} &
    \includegraphics[width=\widthscale\textwidth]{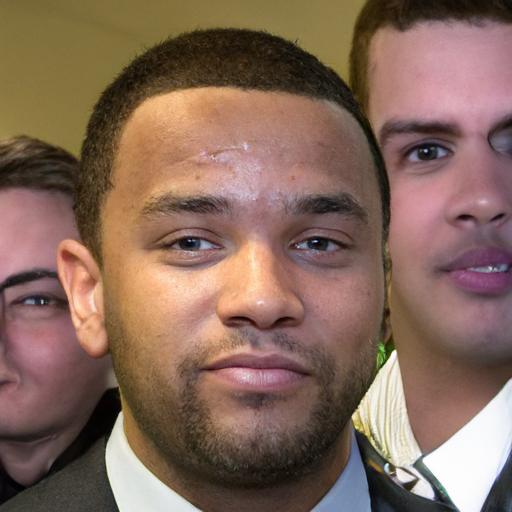} \hspace{-4mm} 
    \\
    \dataset~\imgnote \hspace{-4mm} &
    RestoreFormer++~\cite{wang2023restoreformer++} \hspace{-4mm} &
    VQFR~\cite{gu2022vqfr} \hspace{-4mm} &
    CodeFormer~\cite{zhou2022codeformer} \hspace{-4mm} &
    DAEFR~\cite{tsai2024daefr} \hspace{-4mm} 
    \\
    \includegraphics[width=\widthscale\textwidth]{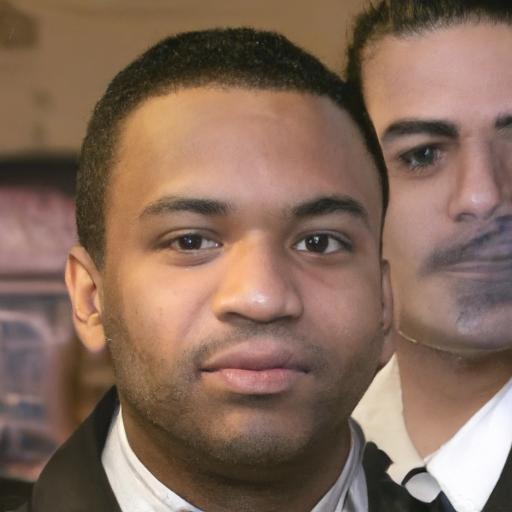} \hspace{-4mm} &
    \includegraphics[width=\widthscale\textwidth]{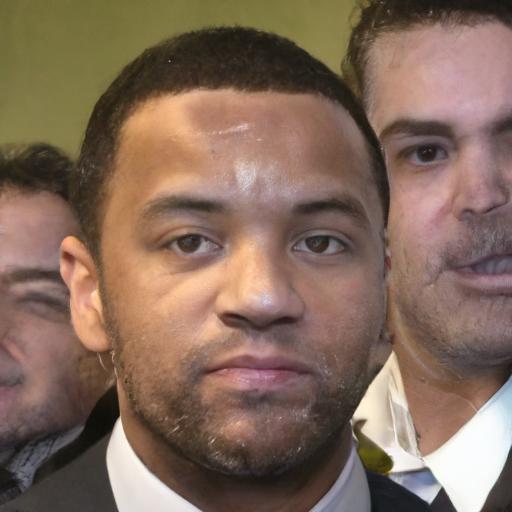} \hspace{-4mm} &
    \includegraphics[width=\widthscale\textwidth]{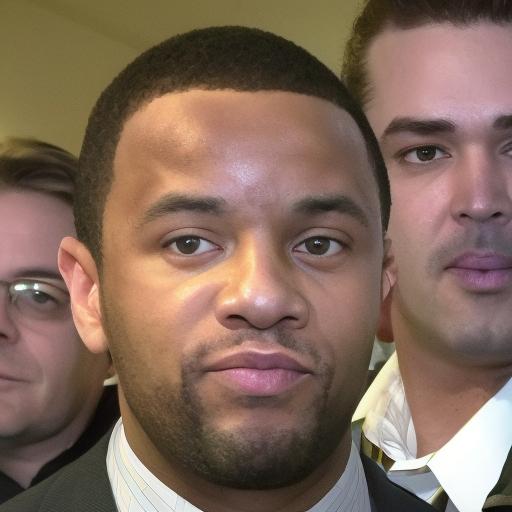} \hspace{-4mm} &
    \includegraphics[width=\widthscale\textwidth]{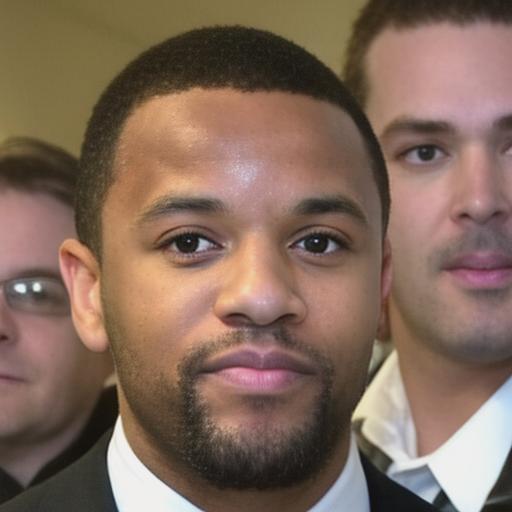} \hspace{-4mm} &
    \includegraphics[width=\widthscale\textwidth]{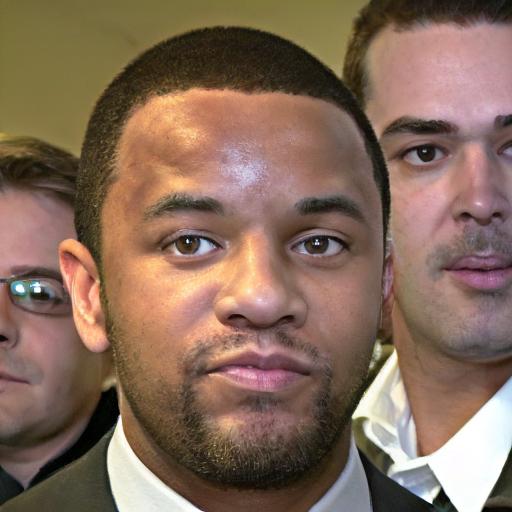} \hspace{-4mm} 
    \\ 
    PGDiff~\cite{yang2023pgdiff} \hspace{-4mm} &
    DifFace~\cite{yue2024difface} \hspace{-4mm} &
    DiffBIR~\cite{lin2024diffbir} \hspace{-4mm} &
    OSEDiff*~\cite{wu2024osediff}  \hspace{-4mm} &
    OSDFace (ours) \hspace{-4mm}
    \\
    \end{tabular}
    \end{adjustbox}
}
\renewcommand{\imgid}{Daniel_Osorno_0001_00}
\renewcommand{\imgnote}{Daniel Osorno}
\scalebox{0.98}{
    \hspace{-0.4cm}
    \begin{adjustbox}{valign=t}
    \begin{tabular}{ccccc}
    \includegraphics[width=\widthscale\textwidth]{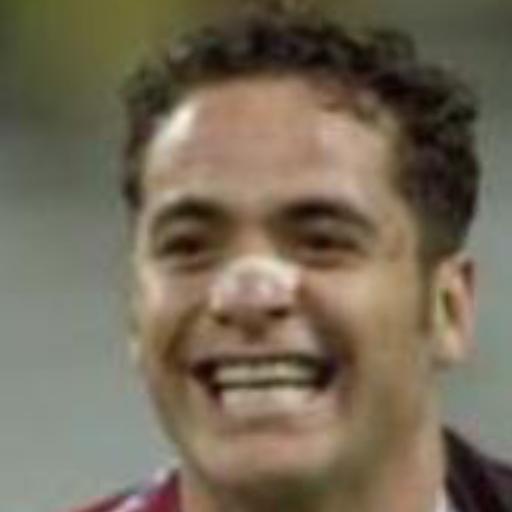} \hspace{-4mm} &
    \includegraphics[width=\widthscale\textwidth]{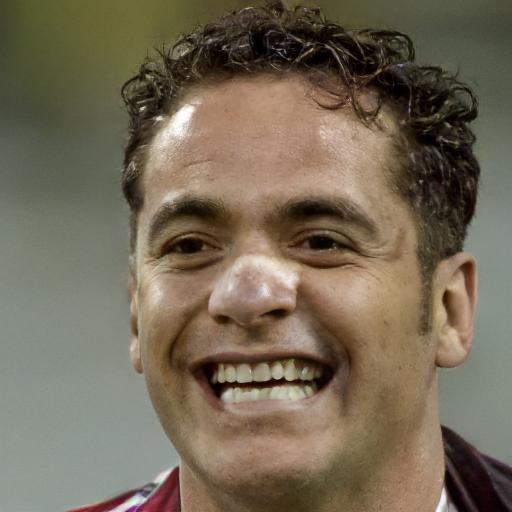} \hspace{-4mm} &
    \includegraphics[width=\widthscale\textwidth]{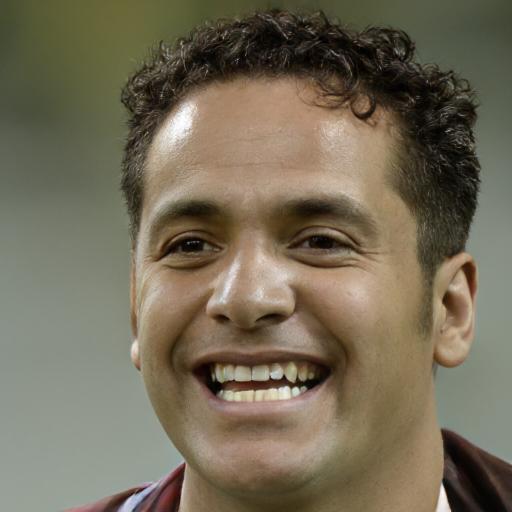} \hspace{-4mm} &
    \includegraphics[width=\widthscale\textwidth]{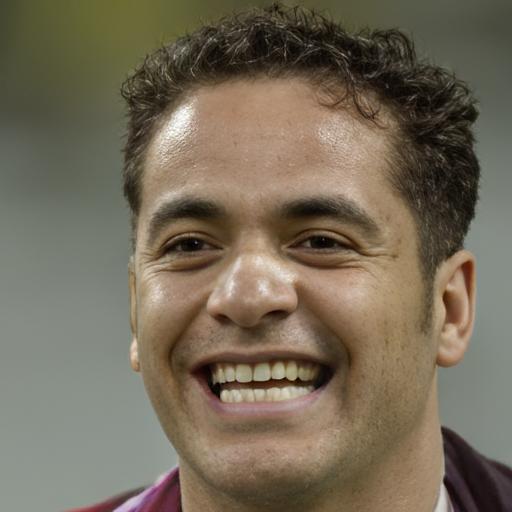} \hspace{-4mm} &
    \includegraphics[width=\widthscale\textwidth]{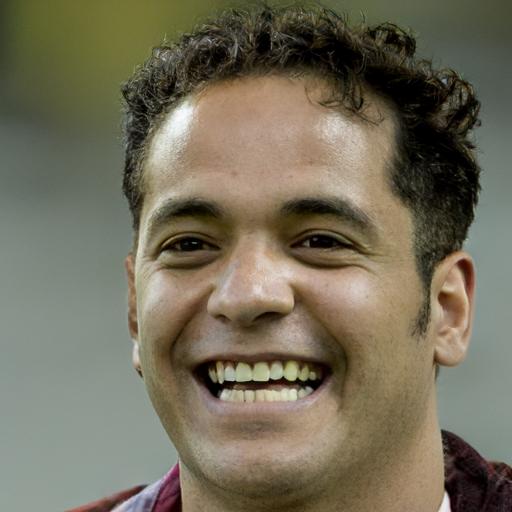} \hspace{-4mm} 
    \\
    \dataset~\imgnote \hspace{-4mm} &
    RestoreFormer++~\cite{wang2023restoreformer++} \hspace{-4mm} &
    VQFR~\cite{gu2022vqfr} \hspace{-4mm} &
    CodeFormer~\cite{zhou2022codeformer} \hspace{-4mm} &
    DAEFR~\cite{tsai2024daefr} \hspace{-4mm} 
    \\
    \includegraphics[width=\widthscale\textwidth]{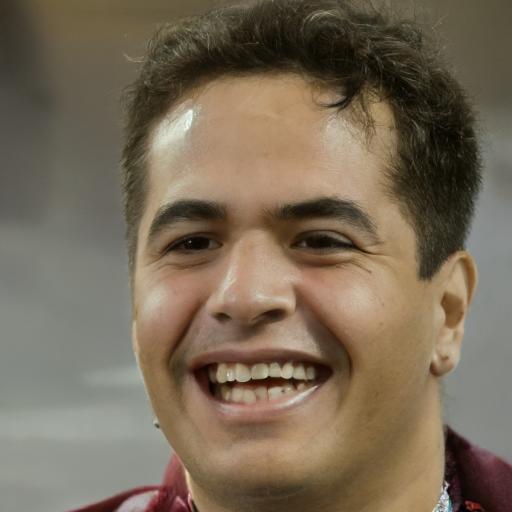} \hspace{-4mm} &
    \includegraphics[width=\widthscale\textwidth]{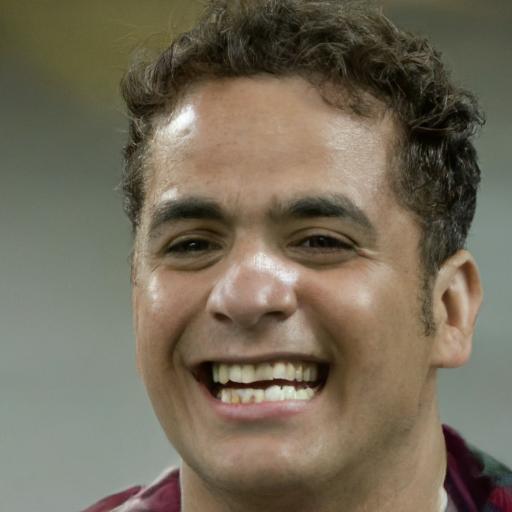} \hspace{-4mm} &
    \includegraphics[width=\widthscale\textwidth]{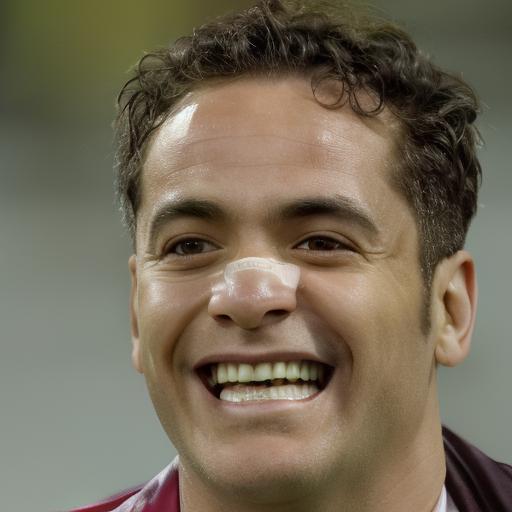} \hspace{-4mm} &
    \includegraphics[width=\widthscale\textwidth]{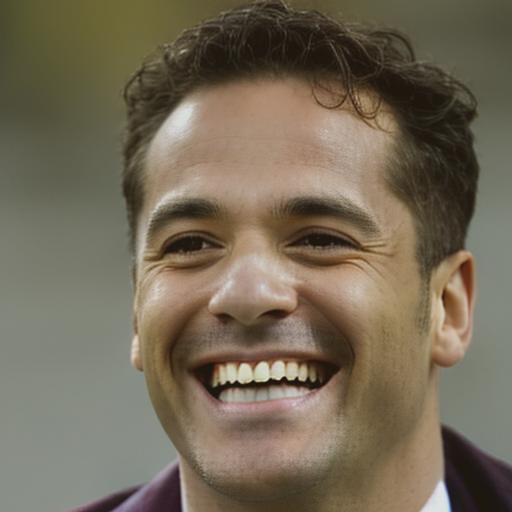} \hspace{-4mm} &
    \includegraphics[width=\widthscale\textwidth]{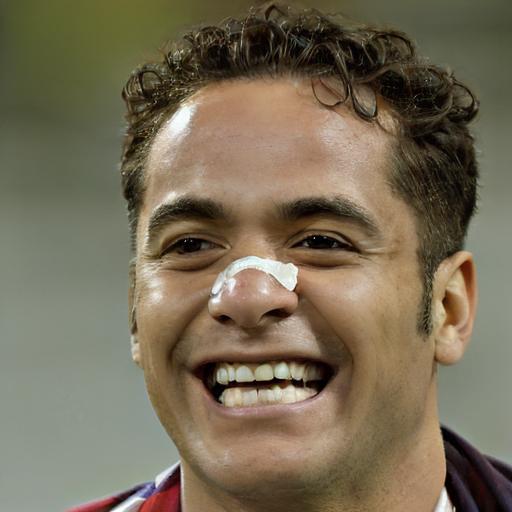} \hspace{-4mm} 
    \\ 
    PGDiff~\cite{yang2023pgdiff} \hspace{-4mm} &
    DifFace~\cite{yue2024difface} \hspace{-4mm} &
    DiffBIR~\cite{lin2024diffbir} \hspace{-4mm} &
    OSEDiff*~\cite{wu2024osediff}  \hspace{-4mm} &
    OSDFace (ours) \hspace{-4mm}
    \\
    \end{tabular}
    \end{adjustbox}
}
\vspace{-2mm}
\caption{More visual comparison of the real-world LFW-Test dataset in challenging cases. Please zoom in for a better view.}
\label{fig:vis-supp-lfw}
\vspace{-2mm}
\end{figure*}

\begin{figure*}[t]
\scriptsize
\centering
\newcommand{\widthscale}{0.189}
\newcommand{\dataset}{WebPhoto}
\newcommand{\imgid}{00058_00}
\newcommand{\imgnote}{00058}
\scalebox{0.98}{
    \hspace{-0.4cm}
    \begin{adjustbox}{valign=t}
    \begin{tabular}{ccccc}
    \includegraphics[width=\widthscale\textwidth]{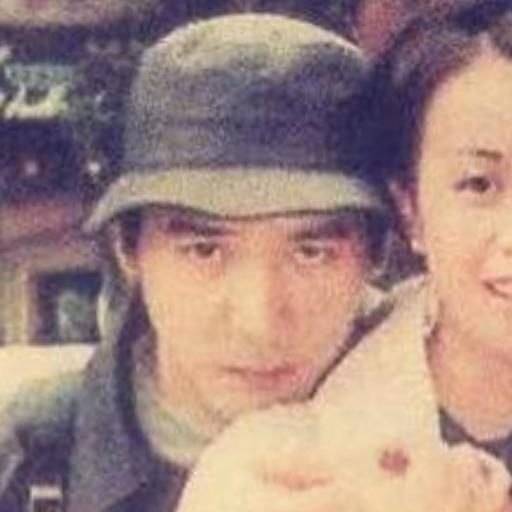} \hspace{-4mm} &
    \includegraphics[width=\widthscale\textwidth]{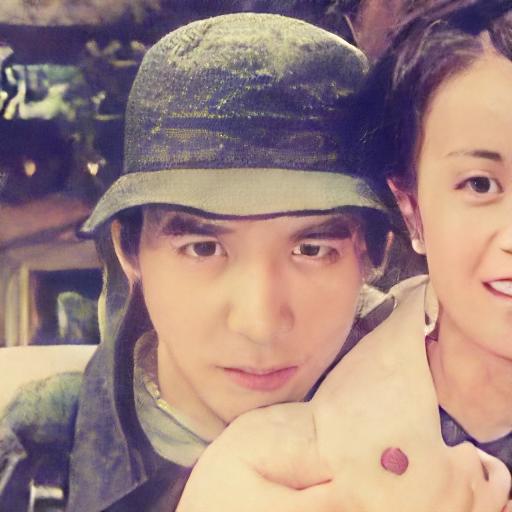} \hspace{-4mm} &
    \includegraphics[width=\widthscale\textwidth]{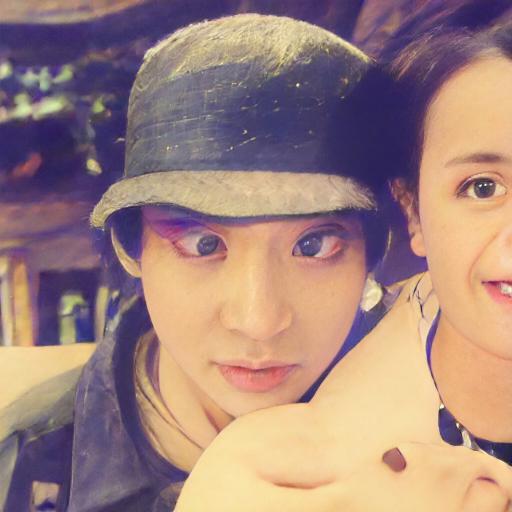} \hspace{-4mm} &
    \includegraphics[width=\widthscale\textwidth]{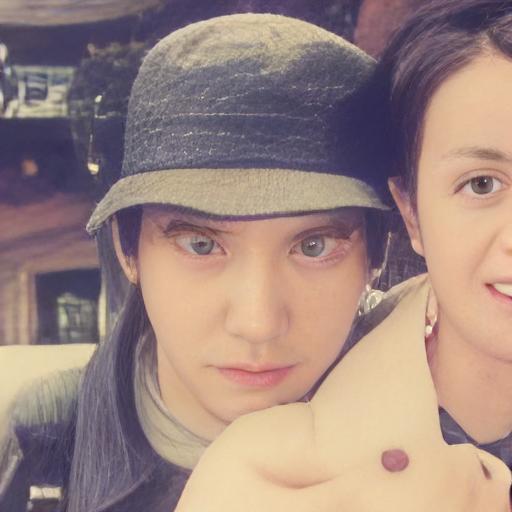} \hspace{-4mm} &
    \includegraphics[width=\widthscale\textwidth]{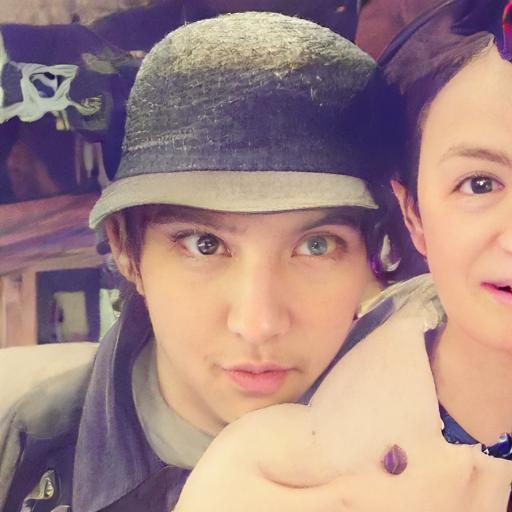} \hspace{-4mm} 
    \\
    \dataset~\imgnote \hspace{-4mm} &
    RestoreFormer++~\cite{wang2023restoreformer++} \hspace{-4mm} &
    VQFR~\cite{gu2022vqfr} \hspace{-4mm} &
    CodeFormer~\cite{zhou2022codeformer} \hspace{-4mm} &
    DAEFR~\cite{tsai2024daefr} \hspace{-4mm} 
    \\
    \includegraphics[width=\widthscale\textwidth]{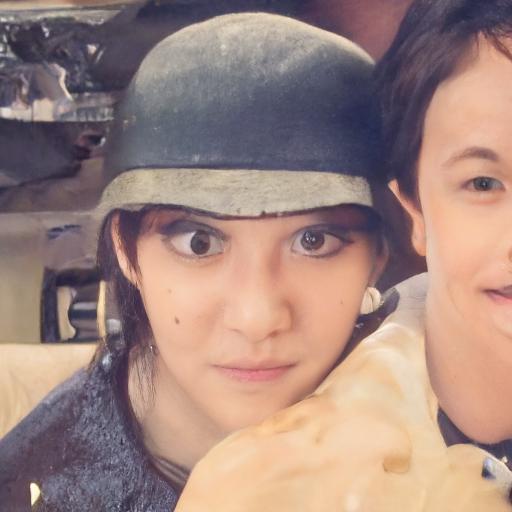} \hspace{-4mm} &
    \includegraphics[width=\widthscale\textwidth]{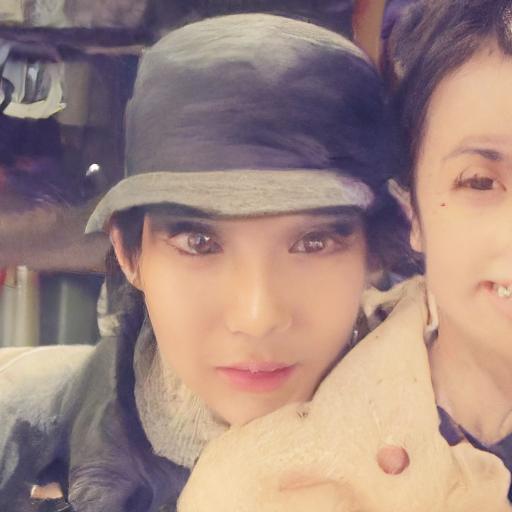} \hspace{-4mm} &
    \includegraphics[width=\widthscale\textwidth]{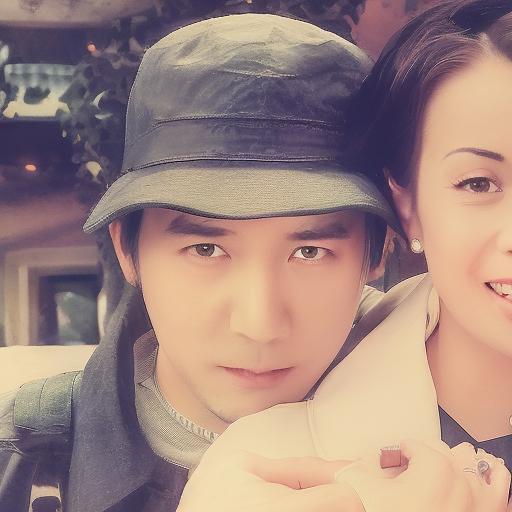} \hspace{-4mm} &
    \includegraphics[width=\widthscale\textwidth]{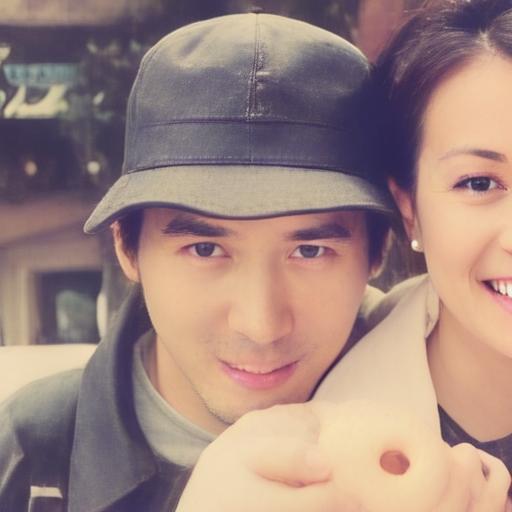} \hspace{-4mm} &
    \includegraphics[width=\widthscale\textwidth]{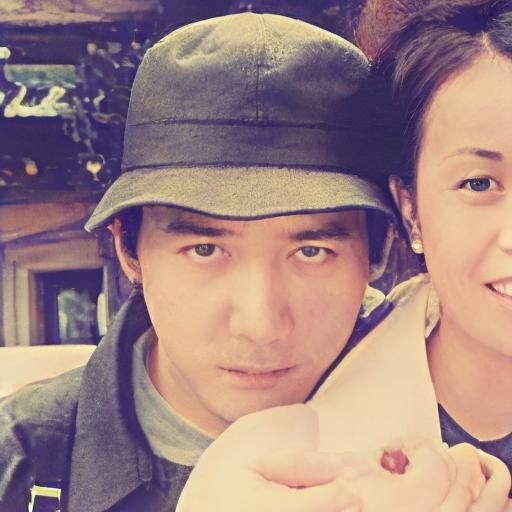} \hspace{-4mm} 
    \\ 
    PGDiff~\cite{yang2023pgdiff} \hspace{-4mm} &
    DifFace~\cite{yue2024difface} \hspace{-4mm} &
    DiffBIR~\cite{lin2024diffbir} \hspace{-4mm} &
    OSEDiff*~\cite{wu2024osediff}  \hspace{-4mm} &
    OSDFace (ours) \hspace{-4mm}
    \\
    \end{tabular}%
    \end{adjustbox}
}
\vspace{-2mm}
\caption{More visual comparison of the real-world WebPhoto-Test dataset in challenging cases. Please zoom in for a better view.}
\label{fig:vis-supp-webphoto}
\vspace{-2mm}
\end{figure*}

\noindent\textbf{Real-world dataset.} More visual comparisons on real-world datasets are shown in Fig.~\ref{fig:vis-supp-wider}, Fig.~\ref{fig:vis-supp-lfw}, Fig.~\ref{fig:vis-supp-webphoto}, and Fig.~\ref{fig:vis-supp-webphoto-2}. Our OSDFace demonstrates strong capabilities in detail generation and boundary distinction. Some images contain multiple closely positioned faces, such as image 0026 in the Wider-Test and Damon Stoudamire in the LFW-Test. Our method successfully restores each individual face. In Wider 0003, only OSDFace successfully generates complete glasses and clearly separates the arm from the face. For faces with varying skin tones, our method consistently maintains the realism of the images. Furthermore, our approach accurately restores facial accessories, including patterns on hats (Wider 0026), bandages (Daniel Osorno in LFW-Test), and earrings (Wider 0173). In old photo restoration scenarios, our OSDFace also effectively handles unknown degradations.

\begin{figure*}[t]
\scriptsize
\centering
\newcommand{\widthscale}{0.189}
\newcommand{\dataset}{Wider}
\newcommand{\imgid}{0173}
\newcommand{\imgnote}{0173}
\scalebox{0.98}{
    \hspace{-0.4cm}
    \begin{adjustbox}{valign=t}
    %
    \end{adjustbox}
}
\vspace{-2mm}
\caption{More visual comparison of the real-world datasets in challenging cases. Please zoom in for a better view.}
\label{fig:vis-supp-webphoto-2}
\vspace{-2mm}
\end{figure*}

{
    \small
    \bibliographystyle{ieeenat_fullname}
    \bibliography{main}
}

%% file: preamble.tex
\usepackage{booktabs}
\usepackage{color}

\usepackage{soul}

\usepackage{amsmath}
\usepackage{amssymb}
\usepackage{booktabs}

\usepackage[dvipsnames]{xcolor}
\usepackage{times,graphicx,tabulary,overpic,xcolor}
\usepackage{colortbl}
\usepackage{makecell}
\usepackage{float}

\definecolor{sh_gray}{rgb}{0.84,0.84,0.84}
\definecolor{sh_gray2}{rgb}{1,0.89,0.75}
\definecolor{color3}{rgb}{0.95,0.95,0.95}
\definecolor{color4}{rgb}{0.96,0.96,0.86}
\definecolor{color5}{rgb}{0.90,0.90,0.90}

\DeclareRobustCommand{\eg}{\emph{e.g.}}
\DeclareRobustCommand{\ie}{\emph{i.e.}}

\usepackage{adjustbox}
\usepackage{engord} 

\usepackage{caption}
\usepackage{subcaption}
\usepackage{float}

\usepackage{tabularx}
\usepackage{makecell}
\usepackage{multirow}
\usepackage{wrapfig}

\usepackage[accsupp]{axessibility}  